\documentclass[twoside,11pt]{article}

\usepackage{blindtext}
\usepackage{jmlr2e}

\usepackage{bibspacing}
\usepackage{appendix}
\usepackage[utf8]{inputenc} 
\usepackage[T1]{fontenc}    
\usepackage{hyperref}       
\usepackage{url}            
\usepackage{booktabs}       
\usepackage{amsfonts}       
\usepackage{nicefrac}       
\usepackage{microtype}      
\usepackage{lipsum}
\usepackage{fancyhdr}       
\usepackage{graphicx} 
\usepackage{verbatim} 
\graphicspath{{media/}}     
\usepackage{amsmath}
\usepackage{pgffor}
\usepackage{multirow}
\usepackage{float}
\usepackage{array}
\usepackage{mdsymbol}

\usepackage{MnSymbol}
\usepackage{rotating}
\usepackage{algpseudocode}
\usepackage{algorithm}
\usepackage{lscape}

\usepackage{mathtools}

\usepackage{rotating}

\newcommand{\ip}[2]{\parbox{#1}{\centering\vspace{0.001cm}#2}}
\newcommand{\cb}[1]{$\left\lbrace #1 \right\rbrace$}
\newcommand{\cm}{$\checkmark$}

\usepackage{acro}
\DeclareAcronym{ER}{
  short = ER,
  long = Erdös-Rényi
}

\DeclareAcronym{CSL}{
  short = CSL,
  long = causal structure learning
}

\DeclareAcronym{SF}{
  short = SF,
  long = scale-free
}

\DeclareAcronym{DAG}{
  short = DAG,
  long = directed acyclic graph
}

\DeclareAcronym{CGM}{
  short = CGM,
  long = causal graphical model
}

\DeclareAcronym{CPDAG}{
  short = CPDAG,
  long = completed partially directed acyclic graph
}

\DeclareAcronym{SCM}{
  short = SCM,
  long = structural causal model
}

\DeclareAcronym{CI}{
  short = CI,
  long = conditional independence
}

\DeclareAcronym{SEM}{
  short = SEM,
  long = structural equation model
}

\DeclareAcronym{DAS}{
  short = DAS,
  long = discovery at scale
}

\DeclareAcronym{ANM}{
  short = ANM,
  long = additive noise model
}

\DeclareAcronym{PNL}{
  short = PNL,
  long = post-nonlinear
}

\DeclareAcronym{GP}{
  short = GP,
  long = gaussian process
}

\DeclareAcronym{MLP}{
  short = MLP,
  long = multi-layer perceptron
}

\DeclareAcronym{NANM}{
  short = NANM,
  long = non-additive noise model
}

\DeclareAcronym{SHD}{
  short = SHD,
  long = structural hamming distance
}

\DeclareAcronym{SID}{
  short = SID,
  long = structural intervention distance
}

\DeclareAcronym{FPR}{
  short = FPR,
  long = false positive rate
}

\DeclareAcronym{TPR}{
  short = TPR,
  long = true positive rate
}

\DeclareAcronym{COD}{
  short = COD,
  long = causal order divergence
}

\DeclareAcronym{DOS}{
  short = DOS,
  long = distance to the optimal solution
}

\DeclareAcronym{EBM}{
  short = EBM,
  long = explainable boosting machine
}

\DeclareAcronym{HC}{
  short = HC,
  long = hill-climbing
}

\DeclareAcronym{TS}{
  short = TS,
  long = tabu
}

\DeclareAcronym{MCDM}{
  short = MCDM,
  long = multi-criteria decision-making
}

\usepackage{lastpage}

\ShortHeadings{Interpretable, multi-dimensional evaluation framework}{Velev and Lessmann}
\firstpageno{1}

\begin{document}

\title{Interpretable, multi-dimensional Evaluation Framework for Causal Discovery from observational i.i.d. Data}

\author{\name Georg Velev$^{1}$ \email velegeor@hu-berlin.de
       \AND
       \name Stefan Lessmann$^{1,2}$ \email stefan.lessmann@hu-berlin.de \\
       \addr $^{1}$ 
       Humboldt University Berlin\\
       Unter den Linden 6, 10999 Berlin, Germany\\
       $^{2}$ 
       Bucharest University of Economic Studies\\
       6 Piata Romana, 1st district Bucharest, 010374 Romania}

\editor{-}

\maketitle

\begin{abstract}
Nonlinear causal discovery from observational data imposes strict identifiability assumptions on the formulation of structural equations utilized in the data-generating process. However, in real-life settings, the ground-truth mechanism responsible for cause-effect transformations is unknown. Thus, it is impossible to verify its identifiability. This is the first research to assess the performance of structure learning algorithms from seven different families in non-identifiable settings with an increasing degree of nonlinearity. The evaluation of structure learning methods under assumption violations requires a rigorous and interpretable approach, which quantifies both, the structural similarity of the estimation with the ground truth and the capacity of the discovered graphs to be used for causal inference. Motivated by the lack of a unified performance assessment indicator, we propose an interpretable, multi-dimensional evaluation framework, which is specifically tailored to the field of causal discovery from i.i.d. data. In particular, we introduce a six-dimensional evaluation metric, called \ac{DOS}, which aims at providing a holistic overview of the performance of structure learning techniques. Our large-scale simulation study, which incorporates seven experimental factors, shows that besides causal order-based methods, amortized causal discovery delivers results with comparatively high proximity to the optimal solution. 

\end{abstract}


\section{Introduction}
Causal discovery from passively observed data plays a pivotal role in the modeling of interventional queries and counterfactual outcomes. The former provide insights about the change in target distributions, which is caused by changes in predictor distributions \citep{BookIntQueries}. The computation of the unobserved outcome resulting from distributional changes in one or more of the passively observed covariates amounts to estimating counterfactual distributions \citep{CFDistributions}. Forecasting the causal impact of various interventions before the actual manipulation of a real-life system has taken place is highly important for the planning and decision-making phase of policy formulation processes in many fields, e.g., politics, economics, healthcare, etc. In this regard, having access to interventional data, in which the values of some or all graph vertices are altered through interventions prior to the structure learning process, can enhance the identifiability of the ground-truth causal effects among a set of graph nodes \citep{ENCO}. Despite this fact, many research studies focus on the inference of graphical structures from observational data only, e.g., \citep{RSortability,DAGMA,NoCurl}, as interventional data is often costly or even impossible to gain access of \citep{GranDAG}.\\

In addition to the challenges resulting from the availability of interventional data, the causal structure among a set of passively observed variables in real-life settings is often assumed to be characterized by nonlinear relationships. This adds another layer of complexity to the task of \ac{CSL}. While various techniques for inferring nonlinear causal relationships from observational data have been suggested in the literature, the studies published in the field of causal discovery impose strict conditions on the data simulation process, so that the ground truth graph becomes identifiable from observational data only, e.g., \citep{NoTears_nonlinear,DAS, GranDAG, DAGMA,AVICI,NoCurl}. The identifiability assumptions w.r.t. nonlinear causal transformations make the task of structure learning particularly daunting in real-life scenarios since the strict conditions cannot be verified due to the unknown formulation of the ground truth \ac{SEM}.\\

The graphical structures discovered from data containing non-identifiable transformations should be evaluated from a multi-dimensional perspective, to provide maximum transparency, to which extent the true causal graphs could be inferred without interventional data under assumption violations. Besides the most commonly selected evaluation criterion, i.e., \ac{SHD}, which quantifies the structural similarities between the estimations and the ground truth, most studies report a slightly different set of one-dimensional metrics, e.g., \citep{PC_Consistent,GranDAG,DAS,NoTears_nonlinear}, to name a few. While benchmarking studies such as \citep{BNS_AssumptionViolations} include several performance indicators in their evaluation framework, the study does not measure the amount of falsely inferred interventional distributions by the estimated graphs. This is of particular importance in settings characterized by assumption violations. Similarly to \citep{BNS_AssumptionViolations}, \ac{SID}, which quantifies the capacity of recovered graphs for causal inference, is not reported by many studies introducing a novel causal discovery technique, e.g., \citep{NoTears_Linear,NoTears_nonlinear,DAGMA,NoCurl}, to name a few. In addition to this, the lack of a unified multi-dimensional evaluation framework in the field of causal discovery results in reporting that different approaches deliver optimal results in terms of different one-dimensional metrics, e.g., \citep{BNS_CIPCAD,BNS_IIDTimeSeries}. This makes the in-depth analysis of interaction effects between experimental factors of the data-generating framework quite challenging. In this context, the two main contributions of our research are the following:\\
\begin{itemize}
\item First, we perform a large-scale sensitivity analysis of the performance of a diverse set of \ac{CSL} models w.r.t. seven experimental factors. The causal discovery approaches, each of which we evaluate on a total of 15.360 simulated datasets, cover seven different families of techniques from the combinatorial and continuous optimization branches. Our study uncovers the performance of \ac{CSL} models on a non-identifiable, nonlinear structural equation. This is relevant because in real-world scenarios, identifiability of the true \ac{DAG} cannot be guaranteed.
Furthermore, we analyze interaction effects between selected experimental factors of our simulation framework, which provides further insights into the expected performance of causal discovery models in different scenarios.\\
\item Second, to the best of our knowledge, we are the first to formulate an interpretable, six-dimensional performance indicator, which embodies \ac{MCDM} concepts to address specific requirements of causal discovery. In addition to assessing the quality of the computed \ac{DAG} w.r.t. structural differences with the ground truth causal matrix, our metric, i.e., distance to the optimal solution (\ac{DOS}), incorporates the evaluation of interventional distributions, which can be inferred from the estimation. Therefore, our unified evaluation framework overcomes an important limitation of previously used one-dimensional metrics, most of which do not quantify the capacity of structure learning approaches to produce graphical models suitable for causal inference.\\
\end{itemize}
The structure of this study is the following: in Section \ref{theoryObservational} we introduce theoretical details about causal discovery from passively observed data. In Section \ref{sec:relatedWork}, first, we present different formulations of linear and nonlinear \ac{SCM} for data-generating purposes. Then, we provide an overview of combinatorial and continuous optimization causal discovery techniques as well as of the simulation frameworks implemented by different studies. Furthermore, we provide an overview of existing benchmarking studies in \ac{CSL}. In Section \ref{sec:interpretableEvalution}, we present the components of our interpretable, multi-dimensional evaluation framework, which we utilize to explore interactions effects between different experimental factors of the data simulation process. The latter is described in detail in Section \ref{sec:exp}. Finally, in Section \ref{sec:results}, we present the results of our large-scale sensitivity analysis.\\

\section{Theoretical Background: Causal Discovery from i.i.d. observational Data}\label{theoryObservational}
In this section, we present the task of structure learning from observational data in the context of causal inference. We also highlight some of the most widely adopted assumptions in the field of \ac{CSL}.\\

Structure learning from passively observed data aims at learning a graphical model, which is capable of describing the underlying data-generating process by producing similar distributions as those obtained from the ground truth \ac{SEM} \citep{CSLbook}. Not all graphical models can be interpreted causally. Research studies dealing with causal discovery often assume that the ground truth graph can be represented as a \ac{DAG}, in which every sequence of arcs connecting the graph nodes represents a path. The latter implies that none of the nodes in the edge sequences is repeated. In a \ac{DAG}, which is also a \ac{CGM}, every edge between a pair of nodes $X_i \rightarrow X_j$ implies that $X_i$ is a direct cause of $X_j$. In contrast to graphical models in general, \ac{CGM}s support interventional queries, which aim to infer the change in the distribution of $X_j$ caused by an intervention performed on $X_i$. The main reason for computing \ac{CGM}s
from observational data is to estimate the impact of different interventions. While interventional data is expected to significantly boost the performance of \ac{CSL} models, its limited availabity has led to the formulation of specific assumptions, which render the causal graph identifiable from observational data only.\\

In the field of \ac{CSL}, the assumption is often made that the probability distribution over the observed variables X is Markov w.r.t. to the ground truth graph, e.g., \citep{GOLEM,NoGAM,MMHC}. This indicates that the joint probability distribution of the vertices $V$ linked with directed edges $E$ in a graph $G(V,E)$ factorizes in the following way:
\setlength{\belowdisplayskip}{0.5cm} \setlength{\belowdisplayshortskip}{0.5cm}
\setlength{\abovedisplayskip}{0.5cm} \setlength{\abovedisplayshortskip}{0.5cm}
\begin{equation}
p(X)=p(X_1,...X_d)=\prod_{i=1}^{d} p(X_i|pa_i(X))
\label{eq:factorization}
\end{equation}
In equation \ref{eq:factorization}, $X_i$ is directly dependent on its parents. If $X_j \in pa(X_i)$, then $X_j$ is said to be independent of its non-descendants given the direct causes of $X_j$. Put in simple terms, the Markov property assumes direct dependence only among those variables in the true \ac{CGM}, which are connected with a directed causal link. \ac{CGM}s encodes a set of \ac{CI} relationships among the observed vertices. The factorized distribution is assumed to be causally faithful to $G$, only if it represents the exact equivalent to the complete set of \ac{CI} relationships embedded in the true causal graph \citep{PC_Stable,CSLbook}. Identifying \ac{CI}s amounts to eliminating the edge between a pair of nodes given the so-called separating subsets $S$. The latter render a pair of vertices independent of each other by blocking every path between them. In Appendix \ref{sec:Appendix1}, we exemplify the concept of \ac{CI}s based on three basic causal structures. While discovering \ac{CI}s can recover certain components of the ground truth \ac{CGM}, it is not possible to recover the complete causal \ac{DAG}. The faithfulness assumption allows one to identify the Markov equivalence class of $G$, called \ac{CPDAG}. \ac{CSL} methods, which rely on \ac{CI}s testing, e.g., PC-Stable, produce \ac{CPDAG}s, see Section \ref{subsec:CSLSummary}.\\

In addition to the assumption about the Markov property and the causal faithfulness assumption, many research studies in \ac{CSL} adopt the assumption of causal sufficiency, e.g., \citep{GranDAG,AVICI,DAGMA}. According to the latter, before modeling the causal \ac{DAG}, the assumption is made that no unmeasured variables are acting as a hidden cause of a pair of observed nodes. This is because the presence of latent confounders can significantly complicate causal discovery. Further assumptions are related to the identifiability of linear and nonlinear \ac{SEM}, which we detail in Section \ref{subsec:SCM}, when introducing different algebraic \ac{SEM} formulations.\\

\section{Related Work}
\label{sec:relatedWork}

\subsection{Structural Causal Models}
\label{subsec:SCM}
In this section, we describe the application of the \ac{SCM}, also called \ac{SEM}, for data-generating purposes in the field of causal discovery. In addition to linear \ac{SCM}s, we provide an overview of different algebraic formulations of nonlinear structural equations. We also highlight under which conditions the ground-truth graph structure is identifiable when sampling datasets with linear as well as nonlinear \ac{SCM}s.\\ 

SCMs characterize a system of causal relationships between a set of endogenous and exogenous variables. While endogenous attributes descend from other nodes in the causal graph, the cause of exogenous variables is unknown. The causal links in SCMs are expressed with a set of equations, which determine the causal dependence of the attributes on their direct predecessors in mathematical terms. The general form of structural assignments in SCMs is given by:
\begin{equation}
X_i=f_i(pa_i(X),N_i),  \quad i\in\{1,...,d\},
\end{equation}
where $d$ is the total number of endogenous variables in the observational dataset $X\in\mathbb{R}^{n\times d}$ with $n$ number of samples, $pa_i$ is the set of parent nodes with a direct causal impact on the variable $X_i$, $N_i$ is the exogenous noise term, and $f_i$ is a transformation mechanism of specific choice. The functional form of $f_i$ defines how the information contained in the parent nodes is altered to produce child vertices in the ground truth \ac{DAG}. The latter is commonly simulated using \ac{ER} and \ac{SF} graph models. While \ac{ER} graphs have a homogenous node degree distribution, i.e., most vertices in \ac{ER} \ac{DAG}s have a comparatively similar number of connections, \ac{SF} graphs are characterized by high heterogeneity. The sequential growth of \ac{SF} graphs combined with the so-called preferential attachment process makes them more similar to real-world networks, e.g. social networks, than \ac{ER} graphs. For more details on \ac{ER} and \ac{SF} graph models see Appendix \ref{sec:Appendix2}.\\

\ac{SCM}s, which are used for the sampling of synthetic observational data based on the causal structure generated by \ac{ER} and \ac{SF} models, are often assumed to follow the \ac{ANM}. This is because \ac{ANM}s have been shown in the literature to be identifiable from observational data under certain conditions \citep{NPM,CSLadditive,DiffCSL,GOLEM}. In ANMs, the noise term is an additive component in the structural equation. In the linear regime, ANMs with a single parameter per edge can be expressed in the following way \citep{GOLEM,NoTears_Linear}:
\begin{equation}
X_i=X \cdot W_i \plus N_i,  \quad i\in\{1,...,d\},
\label{eq:linearANM}
\end{equation}
where $W_i$ is the vector of weights from the weighted adjacency matrix, which are associated with the parent nodes of variable $X_i$. Non-zero elements in $W$ correspond to cause-effect pairs in the ground-truth graph. Equation \ref{eq:linearANM} indicates that in a linear causal \ac{DAG} the child nodes simply represent the weighted sum of their parent vertices. \citet{NonGaussianANM} report that linear non-Gaussian \ac{SCM}s render the ground-truth causal graph identifiable from the observed dataset. By contrast, sampling the disturbance components from a Gaussian distribution makes the graph identifiable only up to the Markov equivalence class called \ac{CPDAG}, when the faithfulness assumption holds. The only condition, under which the true \ac{DAG} can be recovered from a distribution induced by linear Gaussian \ac{ANM}s, is when the error terms have equal variances \citep{GaussianANMNonEqualVar}.\\

In comparison to linear \ac{CSL}, the causal discovery of nonlinear patterns represents a more challenging task. \citet{NPM} highlight that the assumption of linear causal relationships in real-world datasets is rather unrealistic, as many causal links in practice could be classified as approximately nonlinear. Therefore, numerous studies have examined the performance of \ac{CSL} models on nonlinear \ac{SCM}s \citep{GranDAG,RL,ENCO,DIBS,DAGMA,DiffCSL,AVICI,NoCurl,DAG_GNN,NoTears_nonlinear, GAE}. The different formulations of nonlinear structural equations presented in the literature can be broadly divided into three main groups: nonlinear \ac{ANM}, \ac{PNL} causal model, and nonlinear \ac{NANM}. \ac{ANM}s sampled from a \ac{GP} are among the most common type of \ac{SCM}s with nonlinear patterns included in the simulation framework of \ac{CSL} studies:\\
\begin{equation}
X_i=f_i(pa_i(X)) \plus N_i,  \quad i\in\{1,...,d\},
\label{eq:nonlinearANMGP}
\end{equation}
\begin{equation}
X_i=\sum_{j\in pa_i(X)}f_{i,j}(X_j) \plus N_i,  \quad i\in\{1,...,d\},
\label{eq:additivenonlinearGP}
\end{equation}
where $f$ is the $\ac{GP}$ frequently chosen to have a radial basis kernel, also called a squared-exponential kernel, with length-scale one, and $N_i \sim N(0,\sigma^2)$ represents the mutually independent noise terms. The latter are either standard gaussian with $\sigma^2=1$, or have a uniformly sampled variance. While in  Equation \ref{eq:nonlinearANMGP} the values of each child node are generated by sampling from a multivariate \ac{GP} using all parent nodes, in Equation \ref{eq:additivenonlinearGP} the univariate \ac{GP}-based transformation is applied to every individual parent vertex, and the sum of all resulting values is computed. Research studies implementing the nonlinear additive model in Equation \ref{eq:additivenonlinearGP} include an additive noise term in the \ac{SCM}, e.g., \citep{GranDAG,NoTears_nonlinear}. Therefore, this model can be regarded as a special case of the nonlinear \ac{ANM} in Equation \ref{eq:nonlinearANMGP}. In comparison to the linear formulation of \ac{ANM}s in Equation \ref{eq:linearANM}, when generating causal links from \ac{GP}s, the parent nodes are not weighted with the parameters from the graph adjacency $W$. Similarly, nonlinear simulations using a \ac{MLP} do not take into account the edge weights in the sampling process. \ac{MLP}-based data generation follows the notation presented in Equation \ref{eq:nonlinearANMGP}, where $f_i$ is the forward pass of a neural network. The nonlinear transformation is accomplished by the activation function in the hidden layers. \citet{ENCO} sample \ac{ANM}s from a $Leaky$ $ReLU$-based \ac{MLP} consisting of a single hidden layer with 10 units. In comparison to that, \citet{DAGMA} and \citet{NoTears_nonlinear} increase the capacity of the units in the hidden layer to 100 neurons, the values of which are squashed with $Sigmoid$. \\

Contrary to the formulation of the structural assignments in Equations \ref{eq:nonlinearANMGP} and \ref{eq:additivenonlinearGP}, some research studies incorporate the parameters from the graph adjacency $W$ in the implementation of nonlinear \ac{ANM}s. For instance, \citet{NoCurl},\citet{DAG_GNN} and \citet{GAE} suggest applying $Sine$ and $Cosine$ transformations either before or after the parent nodes are linearly weighted with parameters from $W$. \citet{DAGMA} draw samples from a Bernoulli distribution with a probability computed using a logistic transformation of the linearly weighted causes. Furthermore, \citet{RL} uniformly sample weights for the first-order terms in \ac{ANM}s as well as for second-order features computed with quadratic functions. In addition to that, \citet{NoTears_nonlinear} generate synthetic data with the so-called index models, which combine several nonlinear transformations, each of which is associated with a distinctive set of weights:\\
\begin{equation}
X_i=\sum_{m=1}^{3} f_{i,m}(pa_i(X) \cdot W_{i,m}) \plus N_i,  \quad i\in\{1,...,d\},
\label{eq:indexmodels}
\end{equation}
where $f_{i,1}=Tanh$, $f_{i,2}=Sine$ and $f_{i,3}=Cosine$. Thus, index models apply not a single weight, but three coefficients per causal edge. Since the error term in Equation \ref{eq:indexmodels} is additive, index models implemented in this way can be considered as a special case of nonlinear \ac{ANM}s. In the context of the identifiability of nonlinear \ac{ANM}s, \citet{CSLadditive} prove that provided the transformation mechanism is three-times differentiable and nonlinear in every component, the true \ac{DAG} could be recovered from the observed dataset.\\

Compared to nonlinear \ac{ANM}s, both \ac{PNL}s and \ac{NANM}s apply a nonlinear transformation not only to the parent nodes but also to the latent disturbance terms. \citet{PNL_Original} express \ac{PNL}s as a nested nonlinear structural equation:\\
\begin{equation}
X_i=f_{i,2}(f_{i,1}(pa_i(X)) \plus N_i),  \quad i\in\{1,...,d\},
\label{eq:PNL_original}
\end{equation}
\ac{PNL}s apply a second nonlinearity $f_{i,2}$ on top of the well-known nonlinear \ac{ANM}s. As a result, the inner noise term is also nonlinearly transformed, which can be attributed to measurement distortions in practice. \citet{PNL_Rankbased} suggest a multivariate formulation of \ac{PNL}s, where $f_{i,1}$ is a function of linearly weighted first-order terms and quadratic components of $pa_{i}(X)$, and $f_{i,2}$ is a power function raised to the exponent $\frac{1}{3}$. \citet{PNL_Second_Multivariate} select \ac{GP} and $Sigmoid$-based transformations for $f_{i,1}$ and $f_{i,2}$, respectively. Concerning the identifiability of \ac{PNL}s, \citet{PNL_Indentify} report that among the two nonlinearities, $f_{i,1}$ is assumed not to be an invertible function.  Furthermore, the true \ac{DAG} can be recovered from the \ac{PNL} model, if both $f_{i,1}$ and $f_{i,2}$ are three times differentiable, and if the density of the disturbances is not Gaussian, log-mixed-linear-and-exponential, or a generalized mixture of two exponentials.\\ 

In contrast to \ac{PNL} causal models, \ac{NANM}s do not assume the nonlinearly transformed noise terms to be an additive component in the \ac{SCM}. Based on the formulations presented in the literature, \ac{NANM}s can be expressed in the following way \citep{ENCO, ENCO_SEM,SAM,CAM}:\\
\begin{equation}
X_i=f_{i}([pa_i(X)),N_i]),  \quad i\in\{1,...,d\},
\label{eq:NANM}
\end{equation}
where the disturbances $N_i$ are stacked to the set of parent attributes before the nonlinear transformation is applied. Suggested choices for $f_i$ are the forward pass of a $Tanh$-based \ac{MLP} with a single hidden layer having 20 units and an \ac{GP} function. In contrast to the identifiability assumptions of linear and nonlinear \ac{SCM}s presented in this section, the research studies dealing with \ac{NANM}s do not impose any restriction on the distribution of the noise terms.\\ 

\subsection{Causal Discovery Modelling Techniques}
\label{subsec:CSLSummary}
In this section, first, we present a taxonomy of existing families of causal discovery methods including the latest advancements in the field of \ac{CSL} from passively observed data. We provide a comparative overview of the methodology implemented by structure learning models. Whenever a structure learning approach is not considered in the empirical analysis, we elaborate on the specific reasons for its exclusion.\\

\begin{figure}[h]
\centering
\includegraphics[scale=0.75]{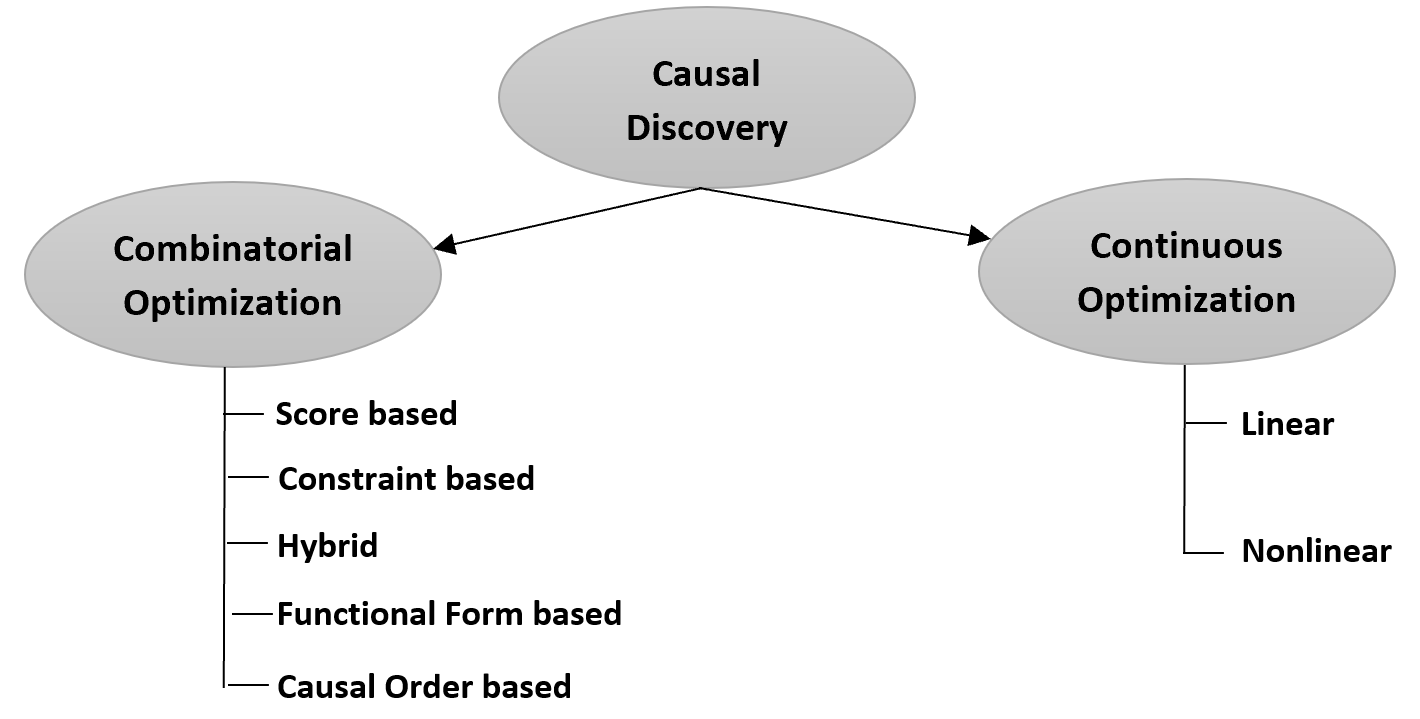}
\caption{Classification of \ac{CSL} Methods}
\label{fig:ClassificationCSL}
\end{figure}
\indent\\
\ac{CSL} models either formulate the search process as a combinatorial optimization problem or estimate the causal structure with a continuous optimization program, as visualized in Figure \ref{fig:ClassificationCSL}. While the former apply iterative testing methods to estimate a discrete-valued graph adjacency, the latter employ gradient descent-based models for the optimization of differentiable objective functions to approximate the real-valued coefficients of the cause-effect pairs in $W$. \\

\textbf{Combinatorial Optimization}\\

The family of combinatorial optimization methods can be split into five types of causal discovery techniques, which are visualized in Figure \ref{fig:ClassificationCSL}. Constraint-based \ac{CSL} approaches apply a set of \ac{CI} tests to infer the underlying causal graph. The PC algorithm presented by \citet{PC_original} is often chosen as a bayesian network learning benchmark of more recently published \ac{CSL} methods, e.g., \citep{RL, DAGMA, AVICI}, even though the original version of the algorithm produces output that is sensitive to the order of the input variables. Thus, rearranging the variables in the passively observed dataset could result in estimating different causal structures. For this reason, we exclude the order-dependent PC algorithm from our empirical research. \citet{PC_Stable} introduce the order-independent PC-Stable method, which estimates the causal structure in three steps. First, the complete undirected graph is pruned by iteratively discovering \ac{CI} relationships. Then, the vertices in the separating sets $S$ are used to identify v-structures, which consist of a directed triple of nodes with a collision vertex in the middle. The last step, which aims at orienting as many as possible of the remaining arcs, produces a \ac{CPDAG} and not a \ac{DAG}, as by using \ac{CI} rules not all triples can be oriented. \\

In comparison to constraint-based \ac{CSL} methods, score-based models are concerned with the optimization of a specific non-differentiable score metric, which depicts how well the estimated \ac{DAG}s fit the observed dataset from a causal perspective. The \ac{HC} search represents a classical score-based greedy search algorithm, which explores the neighborhood of the \ac{DAG} in the current iteration by adding, deleting, or reversing edges, and selects the graph that maximizes the pre-defined score for further modifications. The search is terminated, once no improvement in the score metric can be achieved by any of the neighbor \ac{DAG}s, and the final output often represents a local maxima. In comparison to \ac{HC}, the tabu search algorithm keeps track of recently visited \ac{DAG}s, in order to explore new promising regions without getting stuck in a local maxima. Hybrid causal discovery models make use of techniques from constraint-based and score-based methods to discover the underlying causal \ac{DAG} in two phases. First, the skeleton of the graph is estimated with variable selection algorithms based on identified \ac{CI}s, and the following orientation phase selects the \ac{DAG} with the best score metric. For instance, MMHC presented by \citet{MMHC} starts with the Max-Min Parents and Children algorithm, which implements a forward and a backward search for the set of potential connections of every node. During the forward search, links are added to the list of potential parents and children, if no \ac{CI}s relations are discovered, which would otherwise indicate the lack of the corresponding arcs. The backward search prunes the connections to adjacent vertices by further looking for \ac{CI}s among the potential causes and effects, which were estimated in the forward search. Afterward, the \ac{HC} search determines the causal direction only of previously discovered links, if removal of the edges does not improve the score of the estimated \ac{DAG}s. In comparison to MMHC, the causal discovery method FEDHC presented in \citep{FEDHC} does not make use of the backward search during the skeleton identification phase. During the forward search of FEDHC, the statistical significance of the potential predictors for each vertex is estimated from the results of linear regression analysis, and undirected arcs are added between every pair of variables, only if the association strength is preserved in both directions. Therefore, if, e.g., the predictor $X_i$ has a statistically significant relationship with the target $X_j$, but for the reverse regression model the association strength is not significant, then no connection between $X_i$ and $X_j$ is added to the skeleton. While the hybrid bayesian network learning algorithms published by \citet{MMHC} and \citet{FEDHC} both apply the \ac{HC} search for the orientation of the undirected skeleton, in our empirical research we consider the tabu search following the skeleton identification phase, to avoid the limitations of \ac{HC} related to getting stuck in a local maxima.\\

Functional form-based causal discovery methods, which are specifically designed for linear \ac{SEM} with non-Gaussian noise terms, estimate the causal \ac{DAG} in a fundamentally different way compared to the above described \ac{CSL} models \citep{ICA_LINGAM,DIRECT_LINGAM}. ICA-LiNGAM performs a decomposition of the observed dataset with independent component analysis, which aims at separating linearly mixed variables into mutually independent components. The decomposition results in a mixing matrix, the inverse of which is used to derive an approximate lower triangular matrix through a set of rescaling and matrix permutation operations. The causal order of the \ac{DAG} nodes, which is obtained from the lower triangular matrix, implies the direction of the arcs among the variables. The edges are pruned by examining whether a statistically significant relationship can be established between the variables. \citet{DIRECT_LINGAM} highlight the shortcoming of ICA-LiNGAM, that the convergence of the model for a predefined number of steps is not guaranteed due to the random initialization of the decomposition algorithm. Therefore, the authors introduce Direct-LiNGAM, which computes the causal order by estimating pairwise least square regression models, to assess the independence of the predictors from the residuals in relative terms. Thus, vertices at preceding positions in the causal list, are associated with higher independence from their error terms compared to nodes at succeeding spots in the causal order. In this way, \citet{DIRECT_LINGAM} also avoid applying the scale-variant permutation algorithms used in ICA-LiNGAM, which could produce erroneous causal order in the case of modifications made to the original scale of the data. Since Direct-LiNGAM aims at overcoming the limitations of its predecessor, we exclude ICA-LiNGAM from our empirical research.\\   

Causal order \ac{CSL} models, which estimate the structure of passively observed data in two steps, represent a recently emerged trend in the field of combinatorial causal discovery. In the first step, the topological order of the \ac{DAG} is computed, and in the second step, the edges implied by the causal order are pruned. Therefore, causal order-based \ac{CSL} techniques bear some similarities with functional form-based models due to the estimation of the topological order of the causal \ac{DAG}. However, the algorithms computing the expected causal order are fundamentally different. In addition, causal order-based combinatorial models do not assume a non-Gaussian distribution of the noise terms. \citet{RSortability} introduce $R^2$-SortnRegress, which is specifically designed to address the problem of varsortability\footnote{We provide more details about the problem of varsortability, which was initially reported by \citet{Be_Ware} when introducing the continuous optimization model NoTears}, i.e., the highly dependent performance of some continuous optimization techniques on the scale of the observational dataset. The scale-invariant $R^{2}$-sortability criterion determines the order of the vertices in the estimated graph based on the ascending proportion of the explainable variable's variance. $R^{2}$-SortnRegress computes the topological order  using a series of Linear Regression models, which makes the model suitable for the causal discovery of linear \ac{ANM}s. Furthermore, in the second step of the algorithm, the weights of the causal links are estimated as the product of the coefficients obtained   from regressing each vertex on its predecessors in the $R^2$-sorted order with Linear and Lasso Regression models.\\

In contrast to $R^2$-SortnRegress, SCORE estimates the causal order by sequentially discovering graph leaf nodes based on the Jacobian's variance \citep{SCORE}. The Jacobian approximates the probability density score function, i.e., the gradient of the log-likelihood, of each passively observed vertex using the remaining nodes. The reason for using the variance of the partial derivatives is that the leaf nodes of a graph have constant elements on the diagonal entries of the Jacobian. Therefore, the nodes with the lowest variance of the Jacobian are iteratively added to the causal order list, and removed from the passively observed dataset. In this way, the following computation of the topological order is based on the reduced subgraph. The score’s Jacobian for each vertex is computed using the Stein identity estimator. Spurious edges along the computed causal order are eliminated using the same pruning step as in CAM \citep{CAM}. NoGAM extends SCORE’s causal order search using observational attributes without assuming Gaussian noise distribution \citep{NoGAM}. In addition to computing the output from SCORE for each vertex, NoGAM estimates each node's residuals with a series of Kernel Ridge Regressions using the remaining vertices. The residuals are then used as inputs to estimate the same score function, in order to find the leaf vertex, which has the lowest difference between SCORE’s output and the residuals-based output. NoGAM’s first step makes use of the residual estimates, since, given any generic noise distribution, the residual of a leaf node represents a sufficiently good predictor for that node. This is not the case for non-leaf vertices. DAS improves the scalability of SCORE and NoGAM by reducing the number of potential edges along the causal order to be eliminated in the final CAM-pruning step since the latter takes most of the computational time during the causal discovery process \citep{DAS}. Once the causal order is estimated using SCORE, an additional hypothesis testing step for potential parent nodes is performed using SCORE’s partial derivatives. In particular, the edges implied by the causal order are kept only if the corresponding off-diagonal entries in the Jacobian are significantly different from zero. The hypothesis testing step utilizes Welch’s t-test with a reference node, the average entries of which are the closest to zero. This step results into lower number of spurious edges to be removed in the final CAM-pruning phase. Overall, SCORE, NoGAM, and DAS show a lot of algorithmic similarities. Since DAS' hypothesis testing step prior to CAM-pruning makes the algorithm scalable, we exclude SCORE and NoGAM from our empirical analysis.\\  

\textbf{Continuous Optimization}\\

\ac{CSL} models, which formulate the task of causal discovery as a continuous optimization problem, can further be classified into two groups, as it can be seen in Figure \ref{fig:ClassificationCSL}. Table \ref{tab:contoptoverview} provides an overview of key aspects of gradient-based causal discovery. The objective function directly impacts the causal discovery process as its mathematical formulation quantifies the score produced by the estimated \ac{DAG}s in every training iteration. For instance, the main difference between squared error-based and log-likelihood-based objective functions is that the latter enables the explicit maximization of the data likelihood during the structure learning process. The main novelty in the branch of \ac{CSL} using continuous optimization techniques is related to the continuous relaxation of the acyclicity constraint, which enables the application of gradient-based models to the task of causal discovery. While different studies formulate the acyclicity regularizer in a slightly different way, all three definitions of the acyclicity function are indirectly related to each other, as we detail in this section when describing different continuous structure learning approaches. The third central component of gradient-based \ac{CSL} is related to the modeling capacity for nonlinear causal transformations, as Table \ref{tab:contoptoverview} shows. While the beginning of continuous structure learning is marked by linear models, techniques that build on NoTears, presented by \citet{NoTears_Linear}, commonly incorporate nonlinear mechanisms for inferring causal relationships among the observed variables.
\begin{table}[H]
\centering
\resizebox{\textwidth}{!}{
\begin{tabular}{|c|c|c|c|c|c|c|c|}
			\hline
			\multirow{1}{*}{\ip{1.5cm}{\textbf{CSL Model\\(Reference))}}}& \multicolumn{3}{c|}{\textbf{Objective Function}} & \multicolumn{3}{c|}{\textbf{Acyclicity Constraint/Penalty}} & {\ip{1.7cm}{\textbf{\\Nonlinear\\ causal\\Edges}}}   \\
			\cline{2-7}
			& \ip{2cm}{\textbf{Squared Error based}} &\ip{2.5cm}{\textbf{Log-Likelihood based}} & \ip{1.9cm}{\textbf{Variational}}  & \ip{1cm}{\textbf{Trace based}}  & \ip{3cm}{\textbf{Log-Determinant based}}  & \ip{2cm}{\textbf{Eigenvalue based}}&{} \\
			\hline
			 NoTears   & \ip{1cm}{\cm}   & \ip{1cm}{-}& \ip{1cm}{-}  & \ip{1cm}{\cm}  & \ip{1cm}{-}  & \ip{1cm}{-}  & \ip{1cm}{-}  \\
              \citep{NoTears_Linear} & &  &  & & & & \\
			\hline
                NoTears-\ac{MLP}&\ip{1cm}{\cm}  & \ip{1cm}{-}& \ip{1cm}{-} & \ip{1cm}{\cm} & \ip{1cm}{-} & \ip{1cm}{-}& \ip{1cm}{\cm} \\
                \citep{NoTears_nonlinear} & &  &  & & & & \\
			\hline
			GOLEM & \ip{1cm}{-} & \ip{1cm}{\cm} & \ip{1cm}{-}& \ip{1cm}{\cm} & \ip{1cm}{\cm} & \ip{1cm}{-}& \ip{1cm}{-} \\
                \citep{GOLEM}& &  &  & & & & \\
			\hline
			Gran-DAG & \ip{1cm}{-} &\ip{1cm}{\cm}& \ip{1cm}{-}  &\ip{1cm}{\cm}  &\ip{1cm}{-}  & \ip{1cm}{-}& \ip{1cm}{\cm} \\
                \citep{GranDAG} & &  &  & & &  &\\
			\hline
                DAG-GNN & \ip{1cm}{-} &\ip{1cm}{-}& \ip{1cm}{\cm}  &\ip{1cm}{\cm}  &\ip{1cm}{-}  & \ip{1cm}{-}& \ip{1cm}{\cm} \\
                \citep{DAG_GNN} & &  &  & & & & \\
			\hline
                GAE & \ip{1cm}{\cm} &\ip{1cm}{-}  &\ip{1cm}{-}& \ip{1cm}{\cm}  &\ip{1cm}{-}  & \ip{1cm}{-}& \ip{1cm}{\cm} \\
                \citep{GAE} & &  &  & & &  &\\
			\hline
                NoCurl& \ip{1cm}{\cm} & \ip{1cm}{-} & \ip{1cm}{-}& \ip{1cm}{\cm} & \ip{1cm}{-} & \ip{1cm}{-}& \ip{1cm}{\cm} \\
                \citep{NoCurl} & &  &  & & & & \\
			\hline
			AVICI& \ip{1cm}{-} & \ip{1cm}{-} & \ip{1cm}{\cm} & \ip{1cm}{-}& \ip{1cm}{-} & \ip{1cm}{\cm} & \ip{1cm}{\cm} \\
                \citep{AVICI} & &  &  & & &  &\\
			\hline
			DAGMA& \ip{1cm}{\cm} &\ip{1cm}{-}  & \ip{1cm}{-} & \ip{1cm}{-}& \ip{1cm}{\cm} & \ip{1cm}{-} & \ip{1cm}{\cm}\\
    \citep{DAGMA} & &  &  & & & & \\
			\hline
			RL-BIC& \ip{1cm}{-} & \ip{1cm}{\cm} & \ip{1cm}{-} & \ip{1cm}{\cm} & \ip{1cm}{-} & \ip{1cm}{-}& \ip{1cm}{\cm} \\
   \citep{RL} & &  &  & & &  &\\
			\hline
\end{tabular}}
\caption{Overview of continuous optimization causal discovery techniques}
\label{tab:contoptoverview}
\end{table}
 NoTears is the first linear gradient-based structure learning approach, which applies continuous relaxation to the acyclicity constraint for the discovery of linear \ac{ANM}s \citep{NoTears_Linear}. The constrained optimization problem is expressed mathematically in the following way:\\
\begin{equation}
\begin{split}
\min_{W\in \mathbb{R}^{d,d}}\quad F(X,W)=0.5 \cdot \ell(W, X) \plus \lambda\|W\|_{1},\\
\textrm{ s.t. } h(W)=\textrm{tr}\left(e^{W\circ W}\right) \minus d=0\qquad\qquad\qquad\\
\end{split}
\label{eq:ContOptNoTears}
\end{equation}
where $\ell(W,X)= \frac{1}{n}\|X \minus X\cdot W\|^{2}_{F}$ is the least squares error, and $h(W)$ is the acyclicity constraint. The optimization problem is solved with the augmented Lagrangian method, which adds a quadratic penalty term of the acyclicity constraint to $F(X,W)$ in addition to the term associated with the Lagrangian multiplier. The score function applies $l_1$-regularization to the estimated weighted adjacency to produce sparse \ac{DAG}s. Thus, the incorporation of $l_1$-regularization in the objective function accounts for the structure complexity of the estimated graphs. Furthermore, the minimization of the trace of the matrix exponential encourages the elimination of cycles in the estimated graphs:\\
\begin{equation}
\textrm{tr}\left(e^A\right)=\textrm{tr}(\pmb{I} ) \plus  \frac{\sum^{\infty}_{k=1}}{!k}\textrm{tr} \left(A^k \right),\\
\label{eq:Trace}
\end{equation}
where $\pmb{I}$ is the identity matrix, and $tr(\pmb{I} )=d$. Furthermore, $tr(A^k)$ is associated with the count of all $k$-long closed walks in $A$. Therefore, substituting $A$ with the Hadamard product $W \circ W$ in the trace function, indicates that most \textit{weighted} cycles with $k$ number of vertices between the repeating nodes would be removed from the estimated adjacency matrix, once $tr(e^{W \circ W})$ gets close to zero. However, the final solution obtained from NoTears may have several non-zero entries with very low absolute values, which render the estimated graph cyclic. This is due to finite samples used in the structure recovery process as well as due to the non-convex nature of the optimization problem. For this reason, \citet{NoTears_Linear} prune false discoveries by zeroing out the coefficients in the computed graph, which exceed in absolute value the threshold 0.3. \citet{Be_Ware} highlight that the performance of NoTears is highly dependent on the scale of the observational dataset. The drop in the performance of NoTears with standardized data is explained by the metric varsortability. The latter measures the proportion of edges connecting pairs of nodes, where the parent vertices have strictly lower marginal variance than the child variables. 
Since re-scaling the original features prevents NoTears from exploiting the agreement between the topological sorting of the true causal \ac{DAG} and the order of increasing marginal variance among the variables, the overall performance of the causal discovery method decreases significantly. The latter is positively correlated with a decrease in the varsortability. Despite the issues related to varsortability, various causal discovery frameworks build on NoTears, as it marks the beginning of the continuous optimization branch in the field of structure learning from observational data.\\

In comparison to NoTears, the linear approach GOLEM formulates the task of causal discovery as an unconstrained optimization problem with soft \ac{DAG} and sparsity constraints \citep{GOLEM}:\\
\begin{equation}
\begin{split}
\min_{W\in \mathbb{R}^{d,d}}\quad F(X,W)=\frac{d}{2}\cdot \textrm{log}\ell(W, X) \minus \textrm{log}|\textrm{det}(\textbf{I}\minus W)| \plus \psi(W,\lambda_{1},\lambda_{2}),\\
\end{split}
\label{eq:ContOptGolem}
\end{equation}
where $\psi(W,\lambda_{1},\lambda_{2})=\lambda_{1}\|W\|_{1} \plus \lambda_{2}h(W)$ is the penalty function with the soft constraint terms. The mathematical formulation of \ref{eq:ContOptGolem} makes GOLEM's structure learning problem easier to solve than the optimization problem with hard constraints of NoTears. Furthermore, the log-likelihood alternative of NoTears' least squares error is used in GOLEM's objective function to maximize the data likelihood directly during the structure learning problem. The log-determinant term in GOLEM's score function, which is not present in the objective of NoTears in \ref{eq:ContOptNoTears}, invokes a shared structure between the estimated graph weights of potential cause-effect pairs. In the presence of an edge $X_{i} \rightarrow X_{j}$ in the ground truth graph, the least squares score function without the hard \ac{DAG} constraint would estimate nonzero coefficients in both directions since each variable is a good predictor for the other vertex. By contrast, if the ground truth adjacency matrix is indeed acyclic, then using the log-determinant term would lead to maximizing the likelihood of having an edge in one of both directions for a given cause-effect pair. When the estimated matrix is cycle-free, then the log-determinant term would be zero, but not vice versa \citep{DAGMA}. Thus, GOLEM's log-likelihood score function favours graphical structures, which on the whole generate distributions similar to the underlying data-generating process. The computed graphs could be both cyclical or \ac{DAG}s depending on the ground truth graphical structure.\\

In contrast to NoTears and GOLEM, Gran-DAG supports the modelling of nonlinear causal relationships \citep{GranDAG}. Gran-DAG can be regarded as an extension of both NoTears and GOLEM, as Gran-DAG utilizes a log-likelihood objective function with hard trace-based \ac{DAG} constraint for causal discovery. In contrast to GOLEM, Gran-DAG's score function does not contain the log-determinant term. Gran-DAG estimates the conditional distribution for each node in the graph given the remaining attributes using a Leaky ReLU-based \ac{MLP}. The continuous entries in the weighted adjacency matrix produced by Gran-DAG are obtained from the \ac{MLP}-path products of the trainable weights. Thus, if the trainable variables associated with input $X_i$ and output $X_j$ produce a path product of zero, the node $X_j$ is said to be independent of input $X_i$. Contrary to both GOLEM and NoTears, Gran-DAG also applies preliminary neighbourhood search only if the passively observed dataset contains 50 nodes or more, in order to combat overfitting. \\

In comparison to Gran-DAG, the nonlinear version of NoTears formulates the acyclicity function to be independent of the depth of the \ac{MLP}, as the trace regularizer is applied only to the trainable weights in the first hidden layer of a Sigmoid-based neural network \citep{NoTears_nonlinear}. NoTears-\ac{MLP} makes use of two sets of trainable weights in the first hidden layer, $\theta\plus$ and $\theta\minus$. Both sets of trainable matrices with positive and negative bounds participate in the forward pass of the \ac{MLP} for the modelling of the passively observed attributes, as well as in the computation of the acyclicity constraint. Similarly to Gran-DAG, NoTears-\ac{MLP} relies on the idea that the graph coefficients, which are computed using an \ac{MLP}'s trainable weights, depict the functional relationship between each graph vertex, i.e. the target variables of the \ac{MLP}, and the remaining input variables, i.e., the predictors. Contrary to Gran-DAG and GOLEM, NoTears-\ac{MLP} does not maximize directly the data likelihood as the squared error loss is used for the differentiable objective function. However, NoTears-\ac{MLP} provides a flexible generic gradient-based framework using \ac{MLP}s for the structure recovery of parametric and nonparametric, including nonlinear \ac{SEM}. Besides the Sigmoid-based \ac{MLP} implementation, \citet{NoTears_nonlinear} provide a nonlinear version using orthogonal basis expansions. We do not include this version in our empirical analysis, as \citet{NoTears_nonlinear} report that the NoTears \ac{MLP} with Sigmoid delivers superior performance in a wide range of test settings.\\

In comparison to the linear and nonlinear version of NoTears, GOLEM and Gran-DAG, DAGMA introduces a novel mathematical formulation of the acyclicity constraint, which relies on the continuous relaxation of the nilpotency property of \ac{DAG}s, i.e., an adjacency matrix represents a \ac{DAG} only if all of its eigenvalues are equal to zero \citep{DAGMA}:\\
\begin{equation}
\begin{split}
 h(W)=\minus \textrm{log det}(s\textit{\textbf{I}}\minus W \cdot W ) \plus d\textrm{log}s=0,\\
\end{split}
\label{eq:AcyclicityConstraintDAGMA}
\end{equation}
While the algebraic formulation of DAGMA's $h(W)$ in \ref{eq:AcyclicityConstraintDAGMA} does not have much in common with NoTears' acyclicity constraint in \ref{eq:ContOptNoTears}, it is worth noting the direct dependency between the trace matrix exponential and the eigenvalues $\alpha$ of an adjacency matrix, i.e., $\textrm{tr}(W^k)=\sum^{d}_{i=1}\alpha^{k}_{i}$. For a cycle-free graphical structure, the eigenvalues of which are all equal to zero, \citet{DAGMA} show that $\textrm{det}(s\textit{\textbf{I}}\minus W\cdot W)=s^d$. For this reason, the domain of DAGMA's $h(W)$ is defined to the set of M-Matrices, which in general have the form $s\cdot \textit{\textbf{I}}\minus A$, where $A$ is a non-negative adjacency matrix, and $s$ is a scalar taking values higher than the spectral radius of $A$. This determines feasible options for $s$ used during the structure learning process with DAGMA. The constraint optimization problem within M-Matrices is guaranteed to return a \ac{DAG}, since the set of M-Matrices contains all feasible solutions, i.e., all \ac{DAG}s including the cycle-free matrices, which produce distributions similar to the underlying data generation model. DAGMA bears similarities to both GOLEM and NoTears-\ac{MLP}, as the log-determinant-based acyclicity regularizer is applied to the trainable weights of a Sigmoid-based \ac{MLP}. While the log-determinant term in GOLEM's score function would not necessarily produce \ac{DAG}s, relating the domain of DAGMA's $h(W)$ to the set of M-Matrices makes it a proper acyclicity constraint, i.e., if DAGMA's $h(W)=0$, then the estimation is expected to be cycle-free. In contrast to NoTears-\ac{MLP}, DAGMA makes use of a single set of trainable weights to produce the estimated graph. While the magnitude of the acyclicity regularizer is expected to gradually decrease during the causal discovery process with NoTears-\ac{MLP}, the development of DAGMA's $h(W)$ is rather expected to resemble a bell-shaped curve. The latter results from applying the acyclicity regularizer to very small values at the beginning of the training process, i.e., the trainable weights of a single matrix. The more the causal discovery progresses, the higher certain values in the estimated graph would become. This would be the result of discovered functional relationships between specific input and output vertices. Towards the end of DAGMA’s structure recovery, the acyclicity function would return to values very close to zero, since the objective of DAGMA is constrained to produce acyclic graphs. For more details on the expected development of DAGMA's $h(W)$ during the structure learning process, we refer the reader to Appendix \ref{sec:Appendix5}. In comparison to the trace-based acyclicity constraint, DAGMA's acyclicity function requires less run time to be optimized, has larger gradients, and takes nonzero values in the presence of cycles, which remain undetected by the linear and nonlinear version of NoTears.\\

In contrast to all previously introduced \ac{CSL} methods in the continuous optimization branch, NoCurl estimates the adjacency matrix in two steps \citep{NoCurl}. The computations performed in the first step bear similarities to GOLEM and NoTears, as the task of causal discovery is formulated as an unconstrained optimization problem using the least squares loss from NoTears. Furthermore, NoCurl's score function includes the polynomial \ac{DAG}-ness penalty used in the approach DAG-GNN, which is introduced by \citet{DAG_GNN}:\\ 
\begin{equation}
\begin{split}
 h(W)=\textrm{tr}[(\textit{\textbf{I}}\plus \frac{W \cdot W}{d} )^d] \minus d=0,\\
\end{split}
\label{eq:AcyclicityConstraintNoCurl}
\end{equation}
The commonly used matrix exponential is replaced with the polynomial alternative due to numerical stability, which is not guaranteed when the matrix exponential of W contains large eigenvalues. The largest eigenvalue represents to a certain extent a notion of the degrees of the vertices in a graph, i.e., the connectivity of the nodes. Thus, numerical stability could become an issue, especially at the beginning of the causal discovery process, when the estimated graphs are still densely connected. While the \ac{DAG}-ness formulation adopted by NoCurl is expected to avoid numerical issues, the first step of NoCurl does not guarantee to produce acyclic graphs. This is due to incorporating the polynomial regularizer as a penalty term in NoCurl’s unconstrained formulation of the structure recovery problem. Therefore, in the second step, Hodge decomposition is performed on the initial solution, to extract three orthogonal components, each of which provides different topological insights of the estimated graph. The first component of the Hodge decomposition, i.e., the gradient flow, which in general depicts the flow of information through the nodes in a network, is curl-free, i.e., acyclic. For this reason, NoCurl’s second step utilizes the gradient flow to refine the initially computed adjacency matrix. In contrast to other nonlinear causal discovery methods, e.g, Gran-DAG, NoTears-\ac{MLP} and DAGMA, NoCurl does not make use of an \ac{MLP} to estimate the graph, which is among the reasons why NoCurl is expected to scale much better to larger ground truth graphs than \ac{MLP}-based techniques. We refrain from including other neural-based causal discovery methods, which bear algorithmic similarities to NoTears-\ac{MLP}, Gran-DAG, and NoCurl. Examples of such \ac{CSL} techniques are DAG-GNN and GAE, which make use of the trace matrix exponential and polynomial acyclicity constraints, respectively \citep{DAG_GNN,GAE}. Furthermore, we exclude reinforcement learning-based \ac{CSL} approaches from our empirical analysis due to issues related to scalability \citep{RL}.\\

In comparison to causal discovery techniques, which require a training phase to extract the graph structure, \citet{AVICI} amortize the causal discovery for a given observational dataset by leveraging the predictive power of a pre-trained transformer-based neural network, i.e., AVICI. The parameters of the latter were trained to approximate cycle-free graphs given realizations sampled from linear and nonlinear domain distributions. The nonlinear transformations are generated with \ac{GP}s with the squared-exponential kernel. During training time, \citet{AVICI} optimize a forward Kullback-Leibler divergence-based objective function:\\
\begin{equation}
\begin{split}
 \min_{\phi}\quad \mathbb{E}_{p(X)}X_{KL}\left( p\left(G\vert X\right) \| q\left(G;f_{\phi}(X)\right) \right) =
  \minus \mathbb{E}_{p(G)}\mathbb{E}_{p(X\vert G)} [\textrm{log}q(G;f_{\phi}(X))]\plus \textrm{z},\\
 \textrm{s.t. } h(W)=\rho (W)=\frac{b^{T}Wc}{b^{T}c}=0,\qquad\qquad\quad\quad\quad\quad\\
\end{split}
\label{eq:AVICIOptimization}
\end{equation}
where $z$ is a constant, $\rho (W)$ is the spectral radius of $W$, $b$ and $c$ are randomly initialized vectors. During the structure learning process, $p(G)$ and $p(X\vert G)$ are obtained from a domain-specific simulator, which makes the objective function tractable. The graph posterior $p(G\vert X)$ is approximated using Bernoulli-based variational distribution for $q(G\vert f_{\phi} (X))$, where $f_{\phi}(X)$ is associated with the edge probabilities produced by the transformer-based model. The acyclicity constraint in AVICI is based on the minimization of the spectral radius, as the latter is equal to the largest eigenvalue in absolute magnitude. Power iteration, which is a well-known algorithm for finding the dominant eigenvalue of a matrix, is performed for a certain number of steps using AVICI's $h(W)$ in \ref{eq:AVICIOptimization}. While the mathematical formulation of AVICI's $h(W)$ does not resemble NoTears' and DAGMA's acyclicity functions in \ref{eq:ContOptNoTears} and \ref{eq:AcyclicityConstraintDAGMA}, respectively, conceptually each of these acyclicity regularizers is implicitly or explicitly related to minimizing the eigenvalues of the estimated matrices, to produce acyclic graphs. Since AVICI makes use of amortization during inference time, it is expected to be among the most scalable \ac{CSL} techniques presented in this section.\\

\subsection{Simulation Framework and Evaluation Criteria}
\label{subsec:SFECR}

In this section, we provide a tabular overview of the simulation framework used for the performance assessment of the \ac{CSL} techniques, which we include in our empirical analysis. Based on the overview, we identify inconsistencies and research gaps in the field of \ac{CSL}, which we address in Sections \ref{sec:interpretableEvalution} and \ref{sec:exp}.\\

Table \ref{tab:overview} provides details about 14 causal discovery algorithms included in this benchmark study. The columns related to the benchmark model types show inconsistencies in the set of combinatorial causal discovery methods used as a baseline when the performance of a novel technique is examined. For instance, only approximately $\frac{1}{3}$ of the studies provide a hybrid and LiNGAM-based combinatorial baseline. Additionally, most continuous optimization algorithms are not compared against the recently emerged causal order-based methods. 
\begin{sidewaystable}
		\centering
		\resizebox{\textwidth}{!}{
			\begin{tabular}{|c|c|c|c|c|c|c|c|c|c|c|c|c|c|c|c|}
			\hline
			\multirow{3}{*}{\vspace{-1.3cm}\ip{1cm}{\textbf{Opt. Type}}}
			& \multirow{3}{*}{\vspace{-1.3cm}\ip{3cm}{\textbf{CSL Model (Reference)}}} &\multicolumn{3}{c|}{\textbf{Comb. Opt.Benchmark}}  & \multicolumn{5}{c|}{\textbf{Simulation Framework}} & \multicolumn{6}{c|}{\ip{2cm}{\textbf{Evaluation Criteria}}}\\
			\cline{3-16} 
   
   &   &\multirow{2}{*}{\vspace{-.3cm}\textbf{Hybrid}} &\multirow{2}{*}{\ip{1.8cm} {\textbf{Functional Form based}}}  & \multirow{2}{*}{\ip{1.5cm}{\textbf{Causal Order based}}} &
   
   \multirow{2}{*}{\ip{2.1cm}{\textbf{Nonidentifi-able nonlinear SCM}}} & \multicolumn{2}{c|}{\textbf{Dataset}} & \multicolumn{2}{c|}{\textbf{Graph}}   & \multirow{2}{*}{\vspace{-.3cm}\textbf{SHD}} & \multirow{2}{*}{\vspace{-.3cm}\textbf{FPR}} & \multirow{2}{*}{\vspace{-.3cm}\textbf{TPR}} &\multirow{2}{*}{\ip{1cm}{\textbf{F1-Score}}} & \multirow{2}{*}{\vspace{-.3cm}\textbf{COD}}  & \multirow{2}{*}{\vspace{-.3cm}\textbf{SID}} \\
   
			 \cline{7-10}
    
			&  &  &  &  &  & \ip{2cm}{\textbf{Small Sample size $\in \textbf{[100,500]}$}} & \ip{2cm}
   {\textbf{Large Sample Size $\in \textbf{[1000,2000]}$}} &\textbf{ERk} & \textbf{SFk} &  &  &  & & & \\
			\hline
			\multirow{7}{*}{\vspace{-2.3cm} \ip{1cm}{Comb. Opt.}} & \ip{3cm}{Direct-LiNGAM \citep{DIRECT_LINGAM}}& - & \cm  & - & - & \cm & \cm & - & - & - & - & - & -& -& - \\
			\cline{2-16}
			& \ip{2cm}{PC Stable \newline \citep{PC_Stable}} & - & - & - & - & 50 & \cm & - &  -& \cm & \cm & \cm &-& -& - \\
			\cline{2-16}
			& \ip{2cm}{MMHC \citep{MMHC}} & - & - & - & - & \cm & 20000 & - & - & \cm & - & - & - & - & - \\
			\cline{2-16}
			& \ip{2cm}{PCHC \citep{PCHC}} & \cm & - & - & - & \cm & $5\cdot10^6$ & - & - & \cm &-  & - & - & -& -  \\
			\cline{2-16}
			& \ip{2cm}{FEDHC \citep{FEDHC}} & \cm & - & - & - &\cm  & $5\cdot10^6$ & - & - & \cm &-  & - & - & -& - \\
			\cline{2-16}
			& \ip{2cm}{SortnRegress \citep{RSortability}} & - & \cm & - & - & - & \cm & 2 & 2 & \cm &-  & - & - & \cm & \cm \\
            \cline{2-16}
            &\ip{2cm}{DAS \citep{DAS}}  & - & - & \cm & - &\cm  & \cm & \cb{1,4} & - & \cm & - & \cm & - & - & \cm\\
			\hline
			\multirow{7}{*}{\vspace{-2.3cm}\ip{1.3cm}{Cont. Opt.}} & \ip{3cm}{NoTears Linear \citep{NoTears_Linear}} & - & \cm & - & - & 20 & \cm & \ip{.8cm}{$\left\lbrace 1,2, \right.$ $\left.4\right\rbrace$} & 4 & \cm & \cm & \cm & - & - & -\\
			\cline{2-16}
			& \ip{2cm}{GOLEM \citep{GOLEM}} & - & \cm &-  & - & \cm & 10000 & \ip{.8cm}{$\left\lbrace 1,2, \right.$ $\left.4\right\rbrace$} & 4 &\cm  & -  &\cm  &  -& - & \cm\\
			\cline{2-16}
			& \ip{2cm}{Gran-DAG \citep{GranDAG}} & \cm & - & - & - & \cm & 10000 & \cb{1,4} & \cb{1,4} & \cm & - &-  &- & - & \cm\\
			\cline{2-16}
			& \ip{2cm}{NoCurl \citep{NoCurl}} & \cm & - &-  & - & - & 5000 & \ip{.8cm}{$\left\lbrace 3,4, \right.$ $\left.6\right\rbrace$} &\cb{3,4}  & \cm & - & - &- &-&-\\
			\cline{2-16}
			& \ip{2cm}{AVICI \citep{AVICI}} & - & \cm & - & - & -\footnote{The datasets used during AVICI's training process contain 200 samples, whereas at inference time the approach extracts the causal structure from datasets containing 1000 observations.} & \cm  &2  & 2 & \cm & - & - & \cm & - & \cm\\
			\cline{2-16}
			& \ip{2cm}{DAGMA \citep{DAGMA}} & - & - & - & - & - & 5000 & \ip{.8cm}{$\left\lbrace 2,4, \right.$ $\left.6\right\rbrace$} & 4 &\cm  & \cm &\cm  & -& -& -\\
			\cline{2-16}
			& \ip{3cm}{NoTears-\ac{MLP} \citep{NoTears_nonlinear}} & \cm & - & - & - & \cm & \cm & \ip{.8cm}{$\left\lbrace 1,2, \right.$ $\left.4\right\rbrace$} & \ip{.8cm}{$\left\lbrace 1,2, \right.$ $\left.4\right\rbrace$} & \cm & \cm & \cm & - & -& -\\
			\hline
		\end{tabular}
			
		}
		\caption{Overview of benchmark models, simulation frameworks and evaluation criteria of studies presenting causal discovery techniques, which we include in our large-scale sensitivity analysis.}
\label{tab:overview}
\end{sidewaystable}

The differences in the benchmarking setup coupled with the variation in the simulation frameworks hinder the comparability of the reported performance of \ac{CSL} techniques from different families. Table \ref{tab:overview} shows inconsistencies in, e.g.,  the sample size regimes, i.e., using both large and small data sample sizes, in the connectivity of the simulated \ac{ER} and \ac{SF} graphs, etc. The differences in the synthetic data-generating process across studies highlight the necessity for a large-scale evaluation of \ac{CSL} techniques from different families on, among others, both graph types with comparable connectivity.\\

Concerning the ground-truth \ac{SCM} utilized in the simulation process, Table \ref{tab:overview} shows that none of the 14 studies examines the performance of \ac{CSL} techniques on a nonlinear causal pattern, which can be regarded as nonidentifiable based on the assumptions described in Section \ref{subsec:SCM}. This highlights an important gap in the current state of research in \ac{CSL} since in practice the identifiability of real-world nonlinear patterns cannot be guaranteed. This is because the ground truth formulation of the underlying \ac{SCM} is not known before the modeling of the causal structure. Therefore, it is essential to extend the analysis of causal discovery results on identifiable nonlinear relationships generated with \ac{ANM}s following, e.g., \ac{GP}, Sigmoid-\ac{MLP}, Cosine and Sine based patterns as presented by \citet{GranDAG,AVICI,DAGMA,NoTears_nonlinear,NoCurl}, by incorporating nonidentifiable components in the data generating process.\\

The evaluation criteria selected for the assessment of causal discovery results directly impact the comparability of different approaches. Table \ref{tab:overview} shows that only a few of the \ac{CSL} studies report \ac{FPR}, \ac{TPR}, F1 score, \ac{COD}, and \ac{SID} in addition to \ac{SHD}. None of the \ac{CSL} models in Table \ref{tab:overview} except $R^2$-SortnRegress\footnote{$R^2$-SortnRegress defines $R^2$-sortability criterion, which measures the alignment between the true causal order and the computed $R^2$-scores. While the model is not explicitly evaluated using COD, $R^2$-scores determine the estimated causal order.} are evaluated based on \ac{COD}, although almost every method, except PC-Stable, assumes the ground truth graph is a \ac{DAG}. Details about the mathematical formulation of all six metrics can be found in Appendix \ref{sec:Appendix3}. While \ac{COD} is defined for cycle-free graphs, the computation of the causal order-based metric for adjacency matrices with cycles would require an additional pruning step, which eliminates all links to repeating nodes in the estimated edge sequences. DIRECT-LiNGAM and DAS prune causal links implied by the estimated topological order of the graph, but the structure recovery results are not assessed w.r.t. the recovered causal order. Another significant research gap in the assessment of causal discovery methods results from the fact that only approximately $\frac{1}{3}$ of all models in Table \ref{tab:overview} evaluate the estimated graphs based on \ac{SID}. While \ac{SHD} is suitable for the assessment of the recovered structure in terms of the number of falsely discovered, missing, or reversed edges, \ac{SID} is particularly useful for causal inference. The reason for this is that \ac{SID} measures the number of falsely inferred interventional distributions, when using the discovered graphs for the formation of the cause adjustments sets \citep{SID}. Therefore, \ac{SID} indicates to what extent the recovered graphical structures can be regarded as \ac{CGM}s. Most research studies in the field of \ac{CSL} use the terms structure learning and causal discovery interchangeably, despite the fact that not all graphical models have causal interpretations. This underlines the importance of assessing structure learning methods with causal inference indicators such as \ac{SID}, especially under violation of \ac{SEM} identifiability assumptions.\\

The formulation of the six assessment criteria, which can be found in Appendix \ref{sec:Appendix3}, indicates that each metric evaluates the quality of the estimated causal \ac{DAG} from a different perspective. None of the research studies in Table \ref{tab:overview} attempt to provide a unified framework, which determines the quality of the causal discovery results by considering all six evaluation criteria.\\

\subsection{Related Benchmark Studies}
\label{subsec:RelatedBenchmarks}
In this section, we provide an overview of studies, which focus on benchmarking the performance of existing causal discovery techniques. We also highlight key differences in our study based on the identified research gaps.\\

Table \ref{tab:relatedBenchmarks}, which can be found in Appendix \ref{sec:AppendixBNOverview}, presents details about nine benchmarking studies dealing with causal discovery from i.i.d. observational data. The main contributions of the studies are semi-synthetic data generating frameworks, which facilitate the simulation of realistic causal patterns, as well as the performance exploration in misspecified scenarios, e.g., confounded data generating model, autoregressive model, data generating mechanism, which is not faithful to the ground-truth graph model, etc. \citep{BNS_CausalAssembly,BNS_CIPCAD,BNS_AssumptionViolations,BNS_CovidCSL}.\\

However, none of these studies explores the quality of the estimated graphical structures on non-identifiable nonlinear \ac{SCM}. It is worth noting, that the benchmarking studies, which include the highest number of continuous optimization techniques, do not evaluate the number of wrongly inferred interventional distributions in the estimated graphs \citep{BNS_CIPCAD,BNS_IIDTimeSeries}. Similarly to \ac{SID}, \ac{COD} is rarely used for performance evaluation, although the causal order provides the backbone of the estimated \ac{DAG}. The lack of a unified evaluation framework results in reporting that different causal discovery techniques perform best in terms of different evaluation criteria, e.g., \citep{BNS_IIDTimeSeries,BNS_SurveyFCM}.
This hinders the identification of the state-of-art causal discovery technique(s). In addition, \ac{CSL} studies do not provide an analysis of interaction effects between experimental factors included in the simulation frameworks, even though interaction effects provide in-depth insights into the expected performance in settings that are characterized by multiple factors. For more details on structure learning benchmarking studies, we refer the reader to Appendix \ref{sec:AppendixBNOverview}.\\

Motivated by the identified gaps in research, our simulation framework presented in Section \ref{sec:exp} incorporates non-identifiable nonlinear components, as real-life settings may not comply with the strict identifiability assumptions of \ac{CSL} from observational data. Moreover, our empirical research in Section \ref{sec:results} includes the analysis of interaction effects based on the results obtained from a multi-dimensional performance indicator. The latter includes, among others, both \ac{SID} and \ac{COD}, as we detail in Section \ref{sec:interpretableEvalution}.\\

\section{Interpretable Multi-Criteria Evaluation Framework for Causal Discovery}
\label{sec:interpretableEvalution}
In this section, we present details about the two components of our evaluation framework. First, we provide the motivation for including all six causal discovery assessment criteria introduced in Section \ref{subsec:SFECR} in the computation of our multi-dimensional performance indicator. The second component consists of the inherently interpretable \ac{EBM}, which we utilize for the sensitivity analysis of structure learning results w.r.t. the different experimental factors of our simulation framework. The details of the synthetic data generating process are presented in Section \ref{sec:exp}.\\

\begin{figure}[h]
\centering
\includegraphics[scale=0.59]{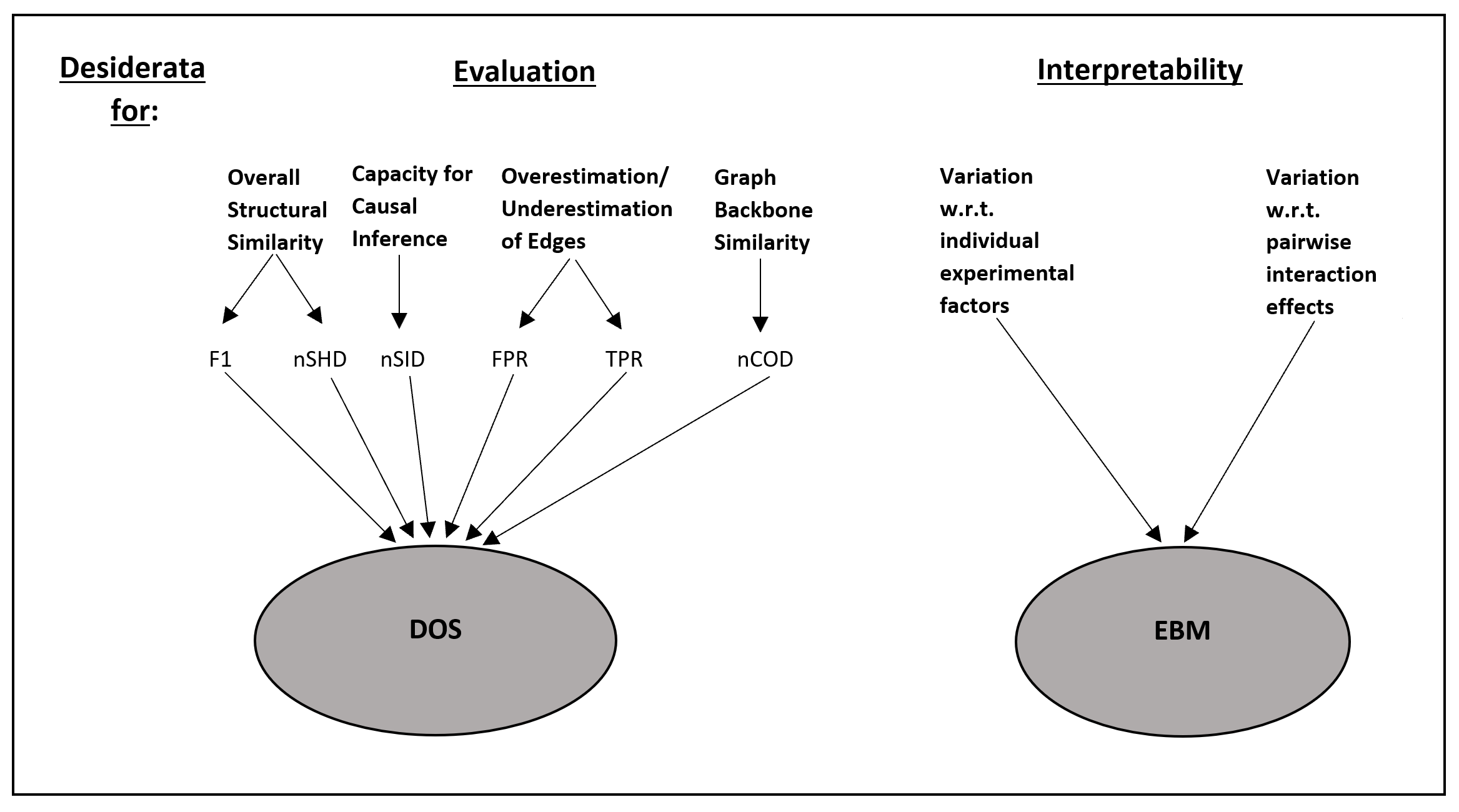}
\caption{Connection between the two components of our performance assessment framework, i.e., \ac{DOS} and \ac{EBM}, visualized based on requirements w.r.t. comprehensive evaluation as well as interpretability.}
\label{fig:DOSEBMConnection}
\end{figure}

Our framework estimates the quality of the discovered graphs from six different perspectives. The latter fulfills the requirements for a holistic performance evaluation in causal discovery, as visualized in Figure \ref{fig:DOSEBMConnection}. The assessment of the overall structural similarity between the true and the inferred graph is accomplished by the metrics \ac{SHD} and F1 Score. The latter also accounts for the discovery of sparse \ac{DAG}s, as the F1 Score is often used for the evaluation of imbalanced classification models. The estimation of the capacity for causal inference, which is a central aspect of the performance assessment of structure learning approaches, is addressed by \ac{SID}. While this metric considers graph aspects associated with wrongly inferred causal effects, \ac{SID} would produce zero if the ground truth adjacency matrix represents a subgraph of the estimation, i.e., if the estimated matrix is much more densely connected than the true graph. This makes it necessary to incorporate also a measure, which accounts for the overestimation of causal links. In this regard, the over- and underestimation of cause-effect pairs are facilitated by \ac{FPR} and \ac{TPR}, respectively. In addition, the quality of the graph backbone represents another local, and yet very important characteristic of the estimated \ac{DAG}s, as the causal order defines the direction of the inferred graph connections. The mistakes in topological ordering among the observed variables are quantified with \ac{COD}.\\

The conventional classification metrics \ac{TPR}, \ac{FPR}, and F1 Score are in the range of $[0,1]$. By contrast, this is not the case for \ac{SHD}, \ac{COD} and \ac{SID}:\\
\begin{equation}
\begin{aligned}
&0 \leq \ac{SHD} \leq E(G) \plus E(\Tilde{G}), \\
&0 \leq \ac{COD} \leq E(G),\\
&0 \leq \ac{SID} \leq d\cdot (d \minus 1),
\end{aligned}
\label{eq:maxbounds}
\end{equation}
where $E(G)$ and $E(\Tilde{G})$ are the number of edges in the ground truth graph and the estimated \ac{DAG}, respectively. Therefore, the magnitude of the upper bounds of \ac{SHD}, \ac{COD}, and \ac{SID} would vary for different number of vertices and different connectivity levels of the ground truth graphs. The difference in the upper bounds significantly hinders their interpretability, especially in comparison to TPR, FPR, and F1 Score. For this reason, in Algorithm \ref{alg:DOSalgorithm}, we present the six-dimensional metric \ac{DOS}, which is inspired by TOPSIS, a well-known \ac{MCDM} method used for ranking of alternatives \citep{MCDM}.
\begin{algorithm}
    \caption{\ac{DOS} for \ac{CSL} model \textit{m}}
    \textbf{Inputs:} \textit{T} test runs with results $\in$ \{\ \textit{FPR}, \textit{TPR}, \textit{F1}, \textit{SHD}, \textit{COD}, \textit{SID} \}\ ,  \textit{O} metric objectives $\in$ \{\ \textit{min}, \textit{max} \}\ \\
    \begin{algorithmic}[1] 
            \State $s_{\plus},s_{\minus} \gets \textrm{scenarios}(O)$
            \State $DOS_{m} \gets \textrm{empty list}$
            \For{$t$ in $T$}
                \State $t_n \gets \textrm{norm\_\ metrics} (t)$
                \State $dist^{\plus}_{t}, dist^{\minus}_{t} \gets \textrm{compute\_\ dist} (  t_n,s_{\plus},s_{\minus})$
                \State $DOS_t \gets \frac{dist^{\minus}_{t}}{dist^{\minus}_{t} \plus dist^{\plus}_{t}}$
                \State $DOS_m.\textrm{insert}(DOS_t)$
            \EndFor
    \end{algorithmic}
     \textbf{\newline Output: } avg $(DOS_m)$ 
\label{alg:DOSalgorithm}
\end{algorithm}
\setlength{\textfloatsep}{1.0cm}
TOPSIS aggregates multiple criteria for the estimation of the geometric distance of each alternative to the ideal case scenario. In comparison to TOPSIS, \ac{DOS} can be computed for each test run sample $t$, i.e., the set of one-dimensional metrics produced by causal discovery techniques on a specific i.i.d. dataset. As shown in the pseudo-code of Algorithm \ref{alg:DOSalgorithm}, all metrics are normalized to be in the range of $[0,1]$, which enables the identification of generic worst and best-case scenarios per evaluation criterion. \ac{DOS} gives all six performance indicators equal importance. Therefore, the \ac{DOS}-based evaluation framework would not be sensitive to the weights, which TOPSIS assigns to each criterion. The final ranking of a set of structure learning approaches can be computed based on $\textrm{avg}(DOS_m)$ achieved by each causal discovery model $m$. For more details on the computation of \ac{DOS}, we refer the reader to Appendix \ref{sec:Appendix7}. The fact that the range of \ac{DOS} is $[0,1]$, makes our six-dimensional causal discovery metric  easier to interpret compared to performance indicators with an open upper bound. The closer \ac{DOS} gets to 1, the smaller the distance of the estimated \ac{CGM} is to the optimal solution, and thus, the better the computed graph can describe the underlying data-generating process.\\

While \ac{DOS} provides a comprehensive way of evaluating the performance of structure learning models, the interpretability of the results from a total of 215.040 test runs w.r.t. seven experimental factors, which are described in Section \ref{sec:exp}, remains a challenge. The reason for selecting the glass-box model \ac{EBM} for the sensitivity analysis of \ac{DOS} w.r.t. different values of the experimental factors is three-fold. First, \ac{EBM} represents an interpretable, nonlinear model, which can achieve prediction results as accurate as other well-established black-box models, e.g., Random Forest \citep{EBM}. \ac{DOS} does not follow a linear computation. Therefore, the ability to model the nonlinear dependencies between the characteristics
of the simulated datasets and the six-dimensional DOS played a role in choosing the memory-efficient tree-based model over its common linear counterparts, e.g., ANOVA and Linear Regression. The second reason is related to the high dimensionality of the results obtained from comprehensive simulations. For instance, our simulation framework contains seven experimental factors, each of which has a different number of levels. The third reason for choosing \ac{EBM} is that the tree-based model can handle mixed data types. This makes the algorithm easily applicable for the interpretation of \ac{DOS} w.r.t. both continuous and categorical experimental factors without the need for any specific data preprocessing.\\

The \ac{EBM} importance scores are utilized as an indicator for the sensitivity  of the overall performance of causal discovery models w.r.t. each experimental factor. In our research, we established that in general an \ac{EBM} importance threshold of 0.01 serves as a reasonable filter for the sensitivity analysis. In Appendix \ref{sec:Appendix8}, we provide details on the selection of the \ac{EBM} threshold. In addition to the importance scores estimated by \ac{EBM}, the glass-box method computes pairwise interaction effects between the experimental factors of our simulation framework.
In this regard, Figure \ref{fig:DOSEBMConnection} highlights that the combination of the two components of our evaluation framework, i.e, \ac{DOS} and \ac{EBM}, is motivated by requirements w.r.t. holistic evaluation as well as an interpretable overview of the performance of causal discovery techniques. On the one hand, the generation of synthetic data is necessary in the field of \ac{CSL} for evaluation purposes, as the ground-truth graph is unknown in real-life settings. On the other hand, designing simulation frameworks entails many degrees of freedom, as shown in Section \ref{subsec:SFECR}. Therefore, the estimation of interaction effects with \ac{EBM} contributes to the interpretability of our evaluation framework, as the pairwise interactions enable the automated identification of scenarios characterized by multiple factors when the performance of causal discovery techniques is expected to deteriorate.\\

\section{Experimental Design}
\label{sec:exp}

In this section, we present the synthetic data-generating process implemented in our study. Besides providing details about the configuration of the sampled graphs and i.i.d. datasets, we also present the non-identifiable, nonlinear pattern used for the causal transformation of the parent vertices into child nodes.\\

\begin{figure}[h]
\centering
\includegraphics[scale=0.65]{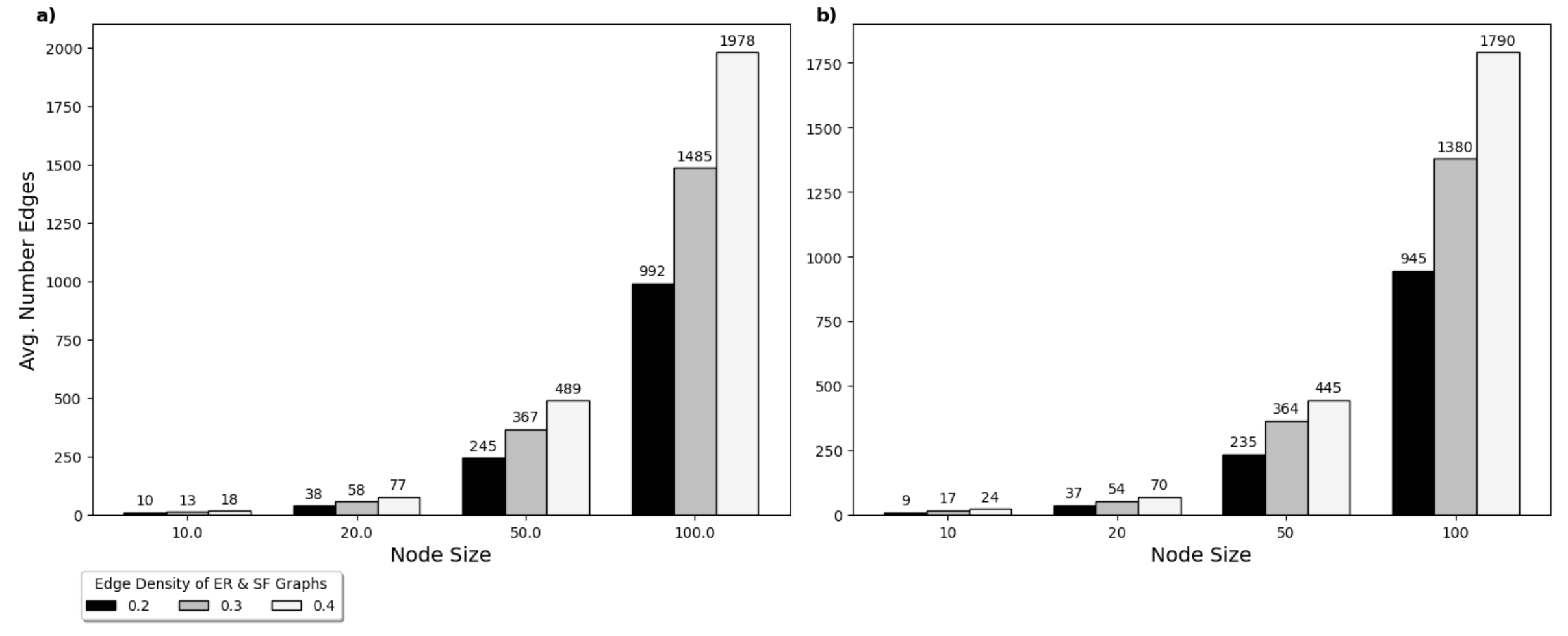}
\caption{Sampled number of edges on average for a ) \ac{ER} graphs, and b) \ac{SF} graphs}
\label{fig:ER_SF_Edges}
\end{figure}

In this benchmark study, the configuration of the simulated causal \ac{DAG}s is determined by seven experimental factors, which can be found in Table \ref{tab: config}. We vary the number of nodes and the connectivity of the sampled graphs in order to explore the performance of \ac{CSL} models in both low-dimensional and high-dimensional settings. In addition, we consider the two most common graph types in \ac{CSL}, \ac{ER}, and \ac{SF} graphs, for the generation of the cause-effect structure of datasets.
{\renewcommand{\arraystretch}{1.5}%
\begin{table}[H]
		\centering
		
		\begin{tabular}{|c|c|c|}
			\hline
			\multicolumn{2}{|c|}{\textbf{Causal Graph and Dataset Parameters}}  & \textbf{Values} \\
                \hline
			\multicolumn{2}{|c|}{Sample Size}  &  [2500, 250]\\
			\hline
			\multicolumn{2}{|c|}{Nodes}  &  [10,20,50,100]\\
			\hline
			\multicolumn{2}{|c|}{Graph Type}  & [ER, SF] \\
			\hline
			\multirow{2}{*}{Connectivity} & ER Graphs & $p$ = [0.2,0.3,0.4] \\
			\cline{2-3}
			& SF Graphs & $k$ based on sampled edges of ER Graphs \\
			\hline
			\multicolumn{2}{|c|}{Nonlinear Pattern} & \parbox{4.5cm}{[Linear 100\%\ , \\ Linear - ReLU 50\%\ , \\ Linear 30\%\ - ReLU 70\%\ , \\ Linear 10\%\ - ReLU 90\%\ ]} \\
			\hline
			\multicolumn{2}{|c|}{$W$ Upper Limit}  & [1,2,3,4] \\
			\hline
			\multicolumn{2}{|c|}{Data Scale}  & [original, standardized] \\
			\hline
\end{tabular}
\caption{Experimental Factors of our Large Scale Simulation Study}
\label{tab: config}
\end{table}}
We initiate the simulation with the \ac{ER} graph type. The edge density of \ac{ER} graphs is determined by $p$, the probability for sampling an edge between each pair of nodes. For \ac{SF} graphs, we configure the connectivity based on the average number of sampled connections in \ac{ER} graphs for a given node size and edge density. The number of outgoing connections to each vertex, which are added during the preferential attachment process, determines the connectivity of \ac{SF} graphs. We create a mapping between the edge densities of \ac{ER} \ac{DAG}s and the scalar value $k$ from Equation \ref{eq:graphcapacity} in Appendix \ref{sec:Appendix2} for \ac{SF} graphs, so that the generated \ac{DAG}s remain comparable for both graph types in terms of connectivity. This can be seen in Figure \ref{fig:ER_SF_Edges}.\\
\indent\\
With regard to the algebraic formulation of the \ac{SCM}, besides simulating identifiable linear \ac{ANM}s with standard normal error terms, we generate nonlinear causal transformations in the following way:\\
\begin{equation}
X_i=ReLU(X \cdot W_i) \plus N_i,  \quad i\in\{1,...,d\},
\label{eq:nonlinearANM}
\end{equation}
We choose $ReLU$ due to the relevance of this nonlinearity for many real-world applications including the modeling of causal relationships. For instance, \citet{DIBS} utilize a ReLU-based gaussian bayesian network to infer the causal structure among variables. In our study, instead of inferring the causal graphical structure with a ReLU-based model, we utilize the nonlinearity in the synthetic data-generating process. This is also motivated by ReLU's thresholding mechanism, which mimics the cause for some neurons in the human brain to fire signals, while others block the information flow. ReLU-based transformations can be found in different fields. For instance, in a business context, ReLU patterns mimic customer buying behavior, since purchasing a product takes place only when customers' willingness to buy exceeds a certain threshold. Analogous examples of ReLU-based transformations could be found in politics, healthcare, academia, etc.\\ 

In addition, ReLU is not three times differentiable. Thus, by simulating ReLU-based cause-effect transformations, we aim to explore to what extent the causal relationships can be inferred by existing \ac{CSL} methods in the challenging scenario, when the theoretical assumptions for identifiability of nonlinear \ac{ANM}s, as described in Section \ref{subsec:SCM}, do not hold. This is of high importance because many research studies in the field of causal discovery present \ac{CSL} methods, which are commonly tested on identifiable cause-effect structures of a specific form, e.g., \citep{NoTears_Linear,GranDAG, SCORE,DAS} etc. However, the identifiability of the transformation mechanism cannot be guaranteed in practice, as the ground truth causal patterns are not known. In this regard, our study is the first to examine the performance of causal discovery methods on ReLU-based \ac{ANM}s on a large scale. Moreover, inspired by the data generation framework presented by \citet{MANM_CS}, we simulate nonlinear cause-effect pairs for three settings of increasing probability (50\%, 70\%, and 90\%), as shown in Table \ref{tab: config}, to analyze the impact of increasing ReLU causal transformations on causal discovery methods. Once the connections between the parents and the child nodes are simulated, the transformation of the incoming edges to a child vertex is sampled from the pre-defined probability cases for linear and nonlinear patterns.  In addition, the causal transformation of the parent vertices is influenced by the parameters in the weighted graph adjacency matrix $W$. We sample $W$ uniformly from $[\minus W_{upper}, \minus W_{lower}] \cup [\plus W_{lower}, \plus W_{upper}]$, where the lower limit is set to 0.5, the default value from \citep{MANM_CS}, and the upper limit is sampled from the list of possible values $[1,2,3,4]$, as indicated in Table \ref{tab: config}. By manipulating the upper limit of $W$ values, we seek to verify the robustness of \ac{CSL} methods towards different scales of the ground truth graph adjacency parameters, as the scale of $W$ has a direct impact on the strength of the causal links. We also examine the impact of changes in the scale of the original variables on the performance of \ac{CSL} models. \citet{Be_Ware} argue that the scale of the sampled datasets contains information about the data generating process, as the marginal variance increases along the causal order for \ac{ANM}s. This could in turn lead to successfully discovering edges from lower-variance attributes to higher-variance nodes by exploiting the data scale of the variables. Therefore, a comparison between the results obtained from the original scale and the standardized version of each dataset provides insights into which \ac{CSL} methods can infer the causal links among the variables when the pattern of marginal variance is eliminated. For each possible configuration, we sample ten datasets containing 2500 data points with different seeds. To explore the performance of \ac{CSL} methods in the case of a small sample size, we also draw 250 samples randomly from each simulated dataset. In Appendix \ref{sec:Appendix4}, we exemplify the simulation process in order to clarify the connection between the experimental factors for the generation of synthetic observational datasets.\\

\section{Large-Scale Causal Discovery Results}
\label{sec:results}

In this section, we present the results of our large-scale benchmarking study. First, we analyze the performance robustness of the \ac{CSL} techniques in terms of \ac{DOS} based on the computed \ac{EBM} scores. We use the latter as a proxy for the sensitivity of the models w.r.t. the experimental factors of the data generating framework as well as w.r.t. the pairwise interaction effects between these factors. The remaining part of this section presents, among others, details about the impact of significant interaction effects on the achieved \ac{DOS} values. The section concludes with the ranking of the 14 \ac{CSL} techniques included in our empirical research.\\

\begin{figure}[h]
\centering
\includegraphics[scale=0.59]{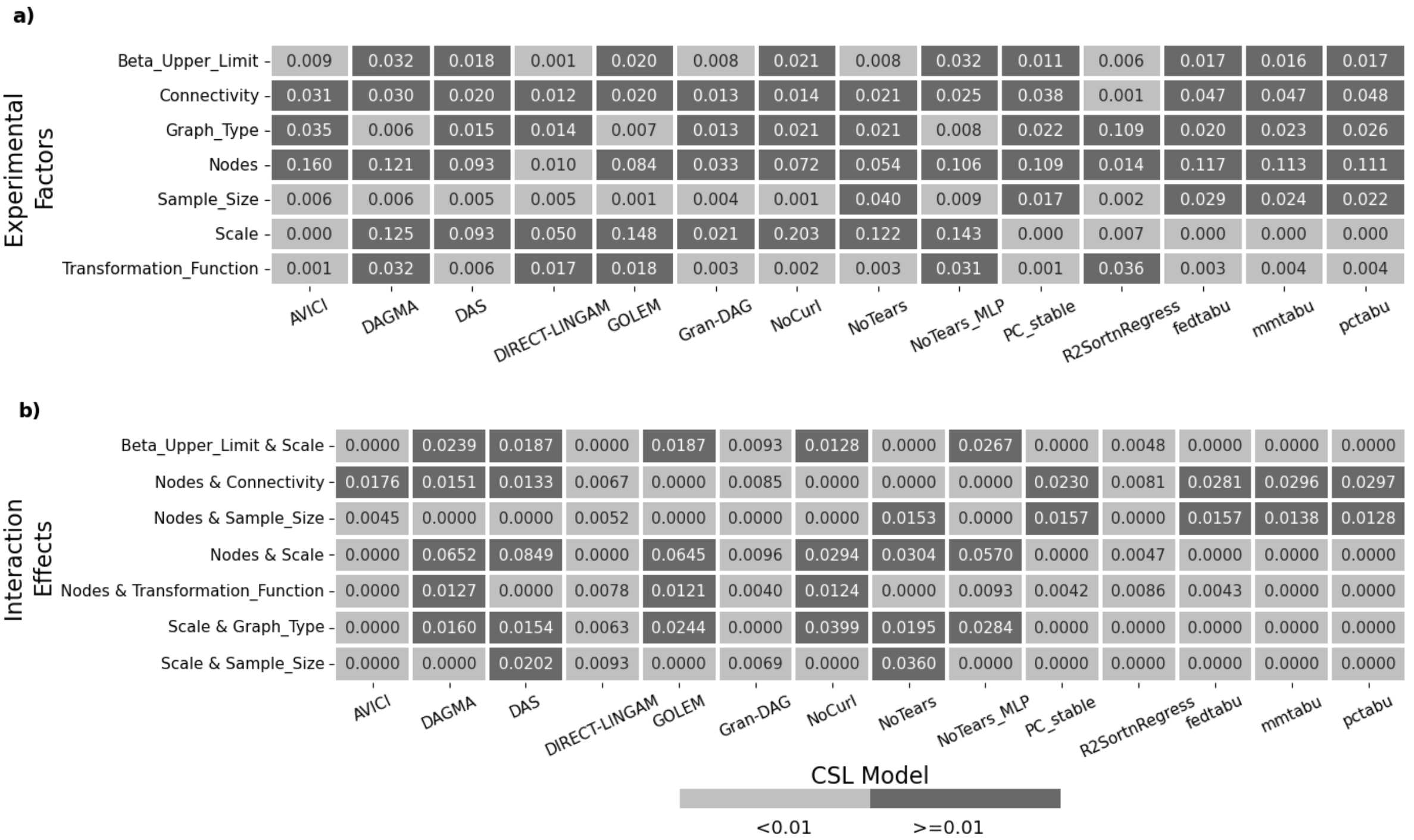}
\caption{EBM importance scores, a) for each of the seven experimental factors in our simulation framework, and b) for the interaction effects estimated by EBM between a pair of experimental factors}
\label{fig:EBM_Importance_Interaction}
\end{figure}

The heatmaps in Figures \ref{fig:EBM_Importance_Interaction}a) and \ref{fig:EBM_Importance_Interaction}b) visualize the \ac{EBM} importance scores per causal discovery model for each of the seven experimental factors and for the pairwise interaction effects.
The dark gray boxes in the heatmaps are associated with importance scores higher or equal to 0.01. This threshold indicates whether the \ac{DOS} values of the \ac{CSL} techniques show significant variation according to the \ac{EBM} scores. For more details on the selection of the \ac{EBM} threshold, we refer the reader to Appendix \ref{sec:Appendix8}. The visual analysis in the remaining part of this section includes only these structure learning approaches, the performance of which is associated with an importance score equal to or higher than 0.01. The heatmap in Figure \ref{fig:EBM_Importance_Interaction}a) shows that the recently introduced linear $R^2$-SortnRegress and the pre-trained nonlinear transformer AVICI deliver more robust performance than all other models. The \ac{DOS} values produced by these two approaches show significant variation only w.r.t. three experimental factors. In this regard, sorting the graph nodes based on increasing cause-explained variance, i.e., $R^2$ coefficients computed from a series of linear regressions, discovers graphical structures, that are sensitive to nonlinear causal transformations. In real-life scenarios, amortized variational inference would deliver more robust results than the causal order-based approach, when modeling cause-effect transformations with unknown formulation, as AVICI's \ac{DOS} values remain invariant to nonlinear structural equations, even when formulated with a non-identifiable causal mechanism. The heatmap in Figure \ref{fig:EBM_Importance_Interaction}b) shows that in addition to $R^2$-SortnRegress and AVICI, Direct-LiNGAM and Gran-\ac{DAG} show insignificant variation in terms of \ac{DOS} w.r.t. the pairwise interaction effects. The fact that Direct-LiNGAM bears methodological similarities to $R^2$-SortnRegress, implies that rather simple linear \ac{CSL} models, which estimate the causal order in the first step, deliver comparatively robust performance. Gran-DAG bears similarities to AVICI in that it represents a nonlinear approach, which makes use of stochastic sampling from a specific distribution to estimate the graph connections. We suspect the reason for Gran-\ac{DAG}'s invariant performance to pairwise interactions can be attributed to the techniques, employed by the approach to combat overfitting, e.g., early stopping of the training process, preliminary neighborhood search prior to the structure learning process of graphs with 50 or more nodes, etc. However, the attempts of \citet{GranDAG} to improve Gran-\ac{DAG}'s robustness make a difference mainly w.r.t to the hidden interaction effects. While the approach remains invariant to the latter, Gran-\ac{DAG}'s performance is sensitive to variation in the values of four out of seven experimental factors.\\  

\begin{figure}[h]
\centering
\includegraphics[scale=0.65]{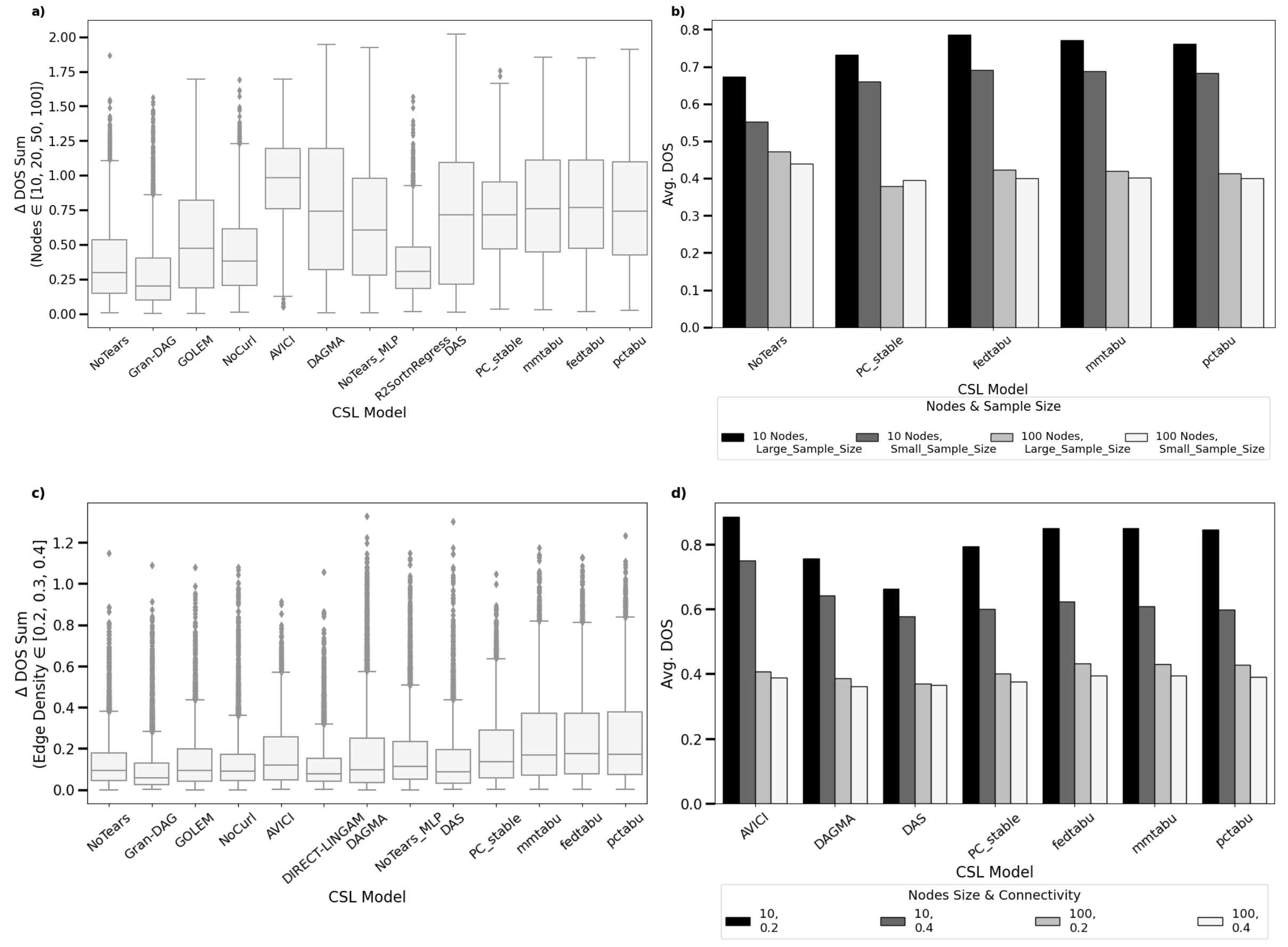}
\caption{a) Sensitivity of causal discovery techniques to varying node sizes, b) interaction effects between node size and sample size, c) sensitivity to different edge density levels, and d) interaction effects between connectivity and node size.}
\label{fig:Nodes_Connectivity}
\end{figure}

Figures \ref{fig:Nodes_Connectivity}a) and \ref{fig:Nodes_Connectivity}c) visualize the distribution of the sum of differences in \ac{DOS} values for varying vertex 
sizes and edge densities \footnote{The sum of $\Delta$ \ac{DOS} values were computed in the following way: each test run with the same dataset description except the node size is selected, the absolute difference in \ac{DOS} is computed for node sizes $\in [20, 50, 100]$ from the setting with 10 nodes, and the $\Delta$ \ac{DOS} values are summed up per dataset description. An analogous computation is performed for varying connectivity levels.}. The inter-quartile range of the boxplots indicates, that the performance of \ac{CSL} methods is in most cases more sensitive to changes in the number of graph nodes than to different edge density levels. This has important implications for causal discovery in practice, as the number of graph nodes is known before the structure learning process and can be manipulated to a certain extent, whereas this is not the case for graph connectivity. In this regard, amortized variational inference is less suitable for the causal discovery of large graphs than the linear $R^2$-based sorting method. Furthermore, the performance of Gran-\ac{DAG} and Direct-LiNGAM is less sensitive to variation in the graph connectivity compared to most methods except for $R^2$-SortnRegress. The latter is the only model, which is invariant to changes in the edge density. The conditional bar plots in Figures \ref{fig:Nodes_Connectivity}b) and \ref{fig:Nodes_Connectivity}d) show that the interaction effects of node size with sample size and edge density make a difference in terms of \ac{DOS} mainly in settings with a small number of variables. The interaction with the number of data samples has important implications for the computational cost of the structure learning process. In settings with a high number of nodes, using only a small portion of the entire dataset would significantly reduce the running time of NoTears and (hybrid) bayesian networks at almost no expense of the quality of the recovered graphical structures. The interaction with graph connectivity indicates that in the presence of high uncertainty about the expected edge density of the true graph, settings with a high number of nodes would produce more robust, and yet approximately twice less accurate results in terms of proximity to the optimal solution than settings with 10 variables. DAGMA, which represents an advanced neural-based extension of NoTears, is among the models in Figure \ref{fig:Nodes_Connectivity}d), the performance of which is strongly affected by the interaction of node size with edge density. This finding highlights the negative correlation between the increase in algorithmic complexity and the expected decrease in performance robustness of causal discovery methods, which extend rather simple linear models from the continuous optimization branch. In our empirical research, we observe a similar tendency also w.r.t. interaction effects of the data scale with other experimental factors.\\ 

\begin{figure}[h]
\centering
\includegraphics[scale=0.75]{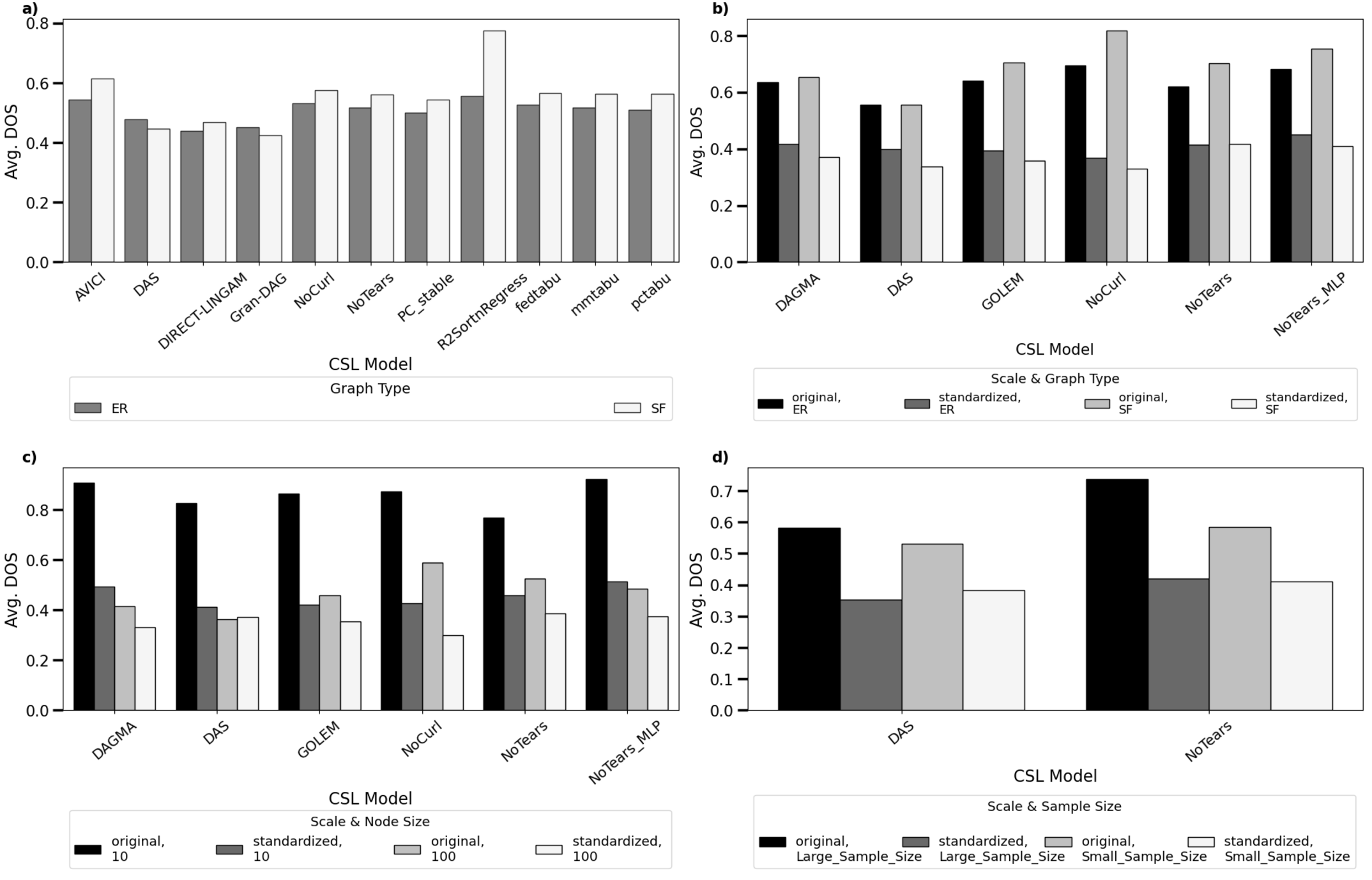}
\includegraphics[scale=0.67]{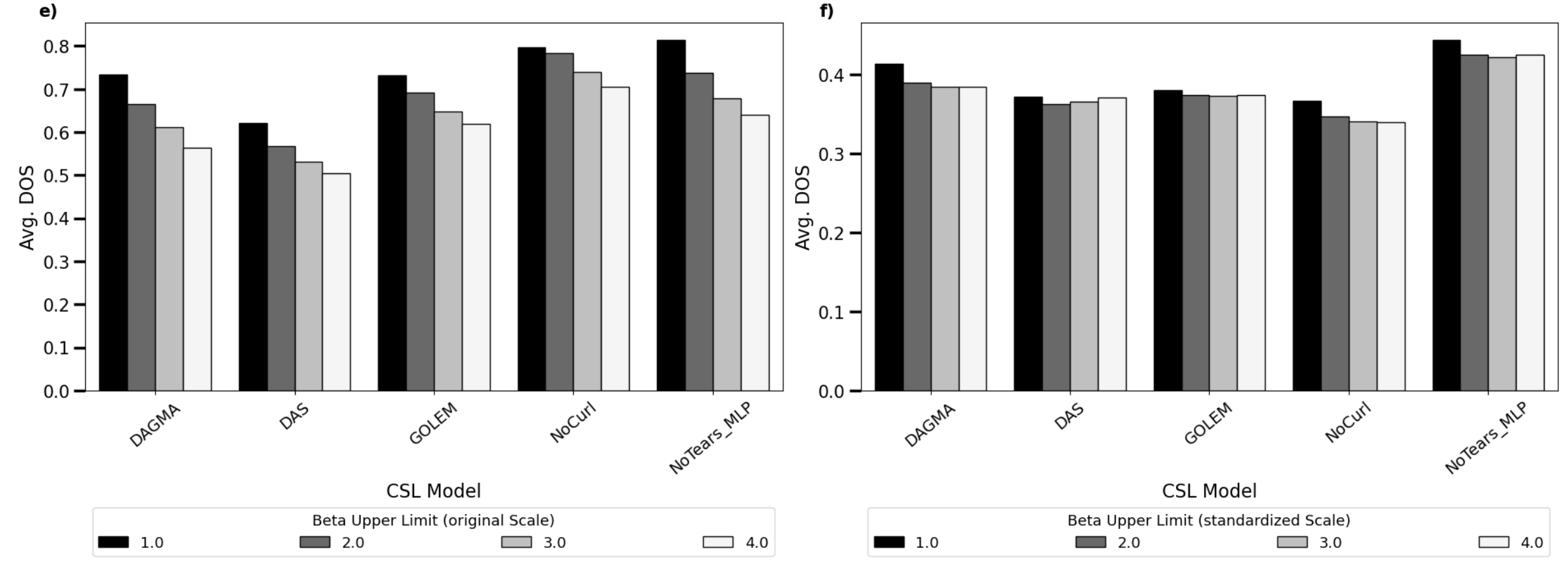}
\caption{a) Sensitivity to different graph types, b), c), d), e), and f) interaction effects between data scale and graph type, node size, sample size, and beta upper limit, respectively.}
\label{fig:Scale_Interactions}
\end{figure}

Therefore, in the following analysis, we devote our attention to significant interaction effects with the scale of the dataset. Overall, changes in the data scale have the highest number of interactions with other experimental factors, which can be seen in Figure \ref{fig:Scale_Interactions}. This underlines the severity of the problem of varsortability in causal discovery. The conditional bar plots in Figures \ref{fig:Scale_Interactions}a) and \ref{fig:Scale_Interactions}b) visualize the variation in the performance w.r.t. the graph type and w.r.t. the interaction of scale with \ac{ER} and \ac{SF} graphs. While $R^2$-SortnRegress is the most sensitive approach to the changes in the graph model, the DOS values produced by this approach remain invariant to interaction effects with scale. The fact that $R2$-SortnRegress achieves \ac{DOS} values, that are approximately 20\%\ higher on \ac{SF} graphs than \ac{ER} graphs on average implies that the approach is suitable for the extraction of the causal structure from real-life observational data, as \ac{SF} graphs bear more similarities with real-world networks than \ac{ER} graphs due to the preferential attachment process. $R^2$-SortnRegress is also expected to produce more accurate causal structures on graphs characterized by high node heterogeneity than any of the remaining models in Figure \ref{fig:Scale_Interactions}a), which further highlights the potential of the linear causal-order-based approach for real-life applications. The interaction effects of the graph type with the data scale in Figure \ref{fig:Scale_Interactions}b) show that changes in the data scale have a slightly bigger impact on graphs with heterogeneous node distribution than on homogeneous graphs. We suspect the reason for this is related to the preferential attachment process, which creates subgraphs within the whole network, i.e., hubs. The more densely connected these hubs are, the bigger the difference in the variance of individual parent and child vertices is expected to be. This, in turn, would enhance the impact of eliminating the pattern of marginal variance through re-scaling on the inferred causal structures. 
Figures \ref{fig:Scale_Interactions}c) and \ref{fig:Scale_Interactions}d) show that, overall, the interaction of data scale with node size has a bigger impact on \ac{DOS} than the interaction with sample size.\\
\newpage
\newpage
Similarly to the interaction of the number of vertices with the number of data points, the interaction effects of scale are expected to result in the highest variation of \ac{DOS} in settings, characterized by a rather low number of graph vertices and large sample size. Furthermore, the nonlinear extensions of NoTears, i.e., DAGMA, NoTears-\ac{MLP} and NoCurl, produce \ac{DOS} values, that are more sensitive to the interaction of scale with node size than NoTears itself. Consequently, the performance of novel causal discovery techniques, which extend the functionality of existing simpler algorithms, should ideally be examined on re-scaled passively observed data only, to account for the tendency of structure learning models to exploit the agreement between the data scale and the increasing marginal variance. The conditional bar plots in Figures \ref{fig:Scale_Interactions}e) and \ref{fig:Scale_Interactions}f) show that  data re-scaling can also reduce the impact of varying magnitude of the ground-truth causal links. The coefficients in the simulated weighted adjacency matrices are influenced by changes in the experimental factor related to the beta upper limit. The higher the latter is set, the stronger the causal links are between a set of parent and child nodes. We suspect that the counter-intuitive performance of the models in Figure \ref{fig:Scale_Interactions}e) to produce less accurate graphical structures with increasing causal strength among the observed variables can mostly be attributed to the post-processing techniques, applied at the end of the structure learning process to prune falsely identified connections. In Appendix \ref{sec:Appendix5}, we show that increasing the magnitude of the experimental factor beta upper limit results in an increase in the variance of the observed attributes. GOLEM, NoCurl, DAGMA and NoTears-\ac{MLP} produce a weighted adjacency matrix, the weights of which are pruned with the pre-defined threshold of 0.3. We suspect that the increased variation in the variables of the observational datasets with increasing beta upper limit results in most cases in a higher spread of the estimated coefficients in the continuous adjacency matrices. However, the post-processing step of the models in Figure \ref{fig:Scale_Interactions}e) applies a static threshold, that is not designed to adapt to statistical characteristics of the datasets. Thus, the pruning phase does not take into account that the higher the variance of the estimated beta weights is, the more likely it would be that more of the predicted betas in absolute magnitude exceed the pre-defined pruning threshold of 0.3. Since standardizing the observed dataset brings all variables to a common range, data re-scaling minimizes the negative impact of increasing causal strength on \ac{DOS}, as shown in Figure \ref{fig:Scale_Interactions}f). For more details on the counter-intuitive connection between increasing beta upper limit and decreasing \ac{DOS} on average we refer the reader to Appendix \ref{sec:Appendix5}.\\

Next, we move on to analyzing the impact of increasing ReLU non-identifiable causal transformations on the \ac{DOS} distribution.  
\begin{figure}[h]
\centering
\includegraphics[scale=0.63]{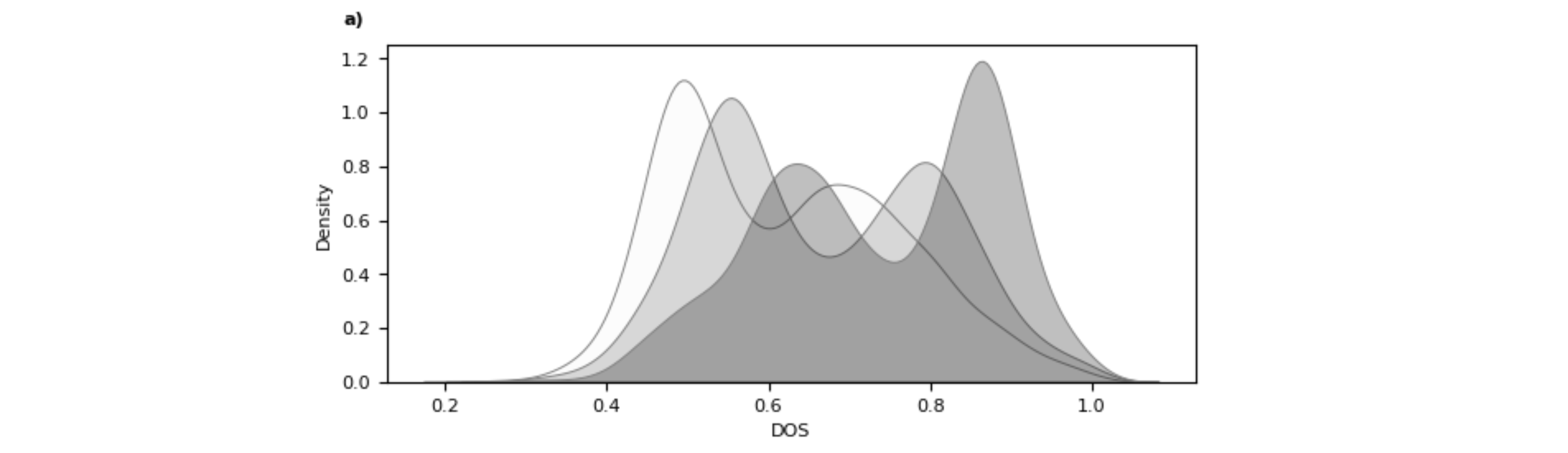}
\includegraphics[scale=0.63]{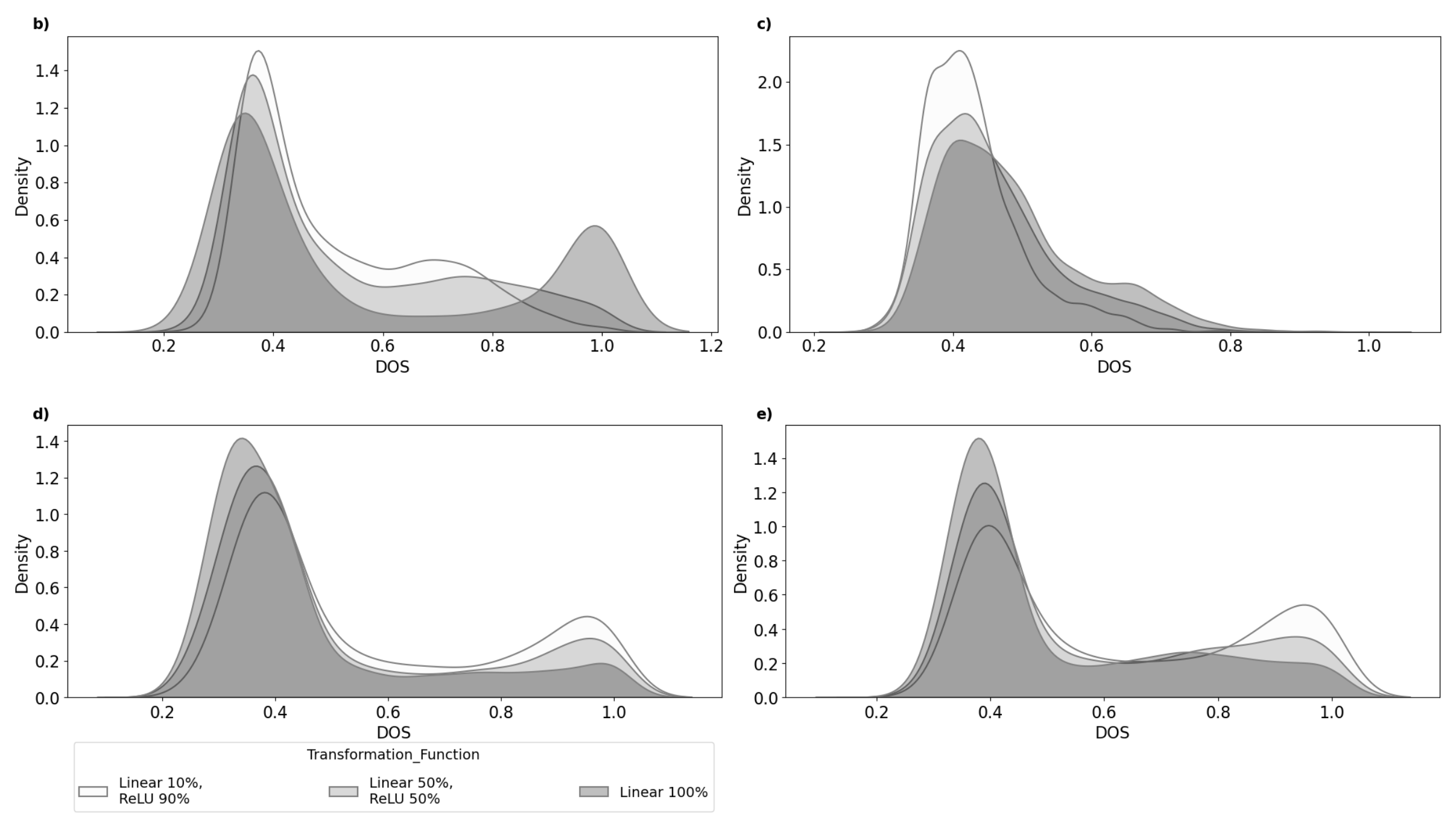}
\caption{\ac{DOS} distribution per ratio of ReLU causal transformations for a) R$^2$-SortnRegress, b) GOLEM, c) Direct-LiNGAM, d) DAGMA, and e) NoTears-\ac{MLP}. The setting of linear 30\%\ and ReLU 70\%\ is excluded from the plots to avoid visualizing kernel densities, which are close to overlapping, and to improve the readability.}
\label{fig:Transformation_Function}
\end{figure}
The kernel density estimation plots in Figure \ref{fig:Transformation_Function} show that often causal discovery techniques, which are sensitive to nonlinear \ac{SCM}, tend to produce bimodal \ac{DOS} distributions in settings, characterized by a specific percentage of ReLU structural equations. In Figure \ref{fig:Transformation_Function}b), the linear approach GOLEM produces two very-well defined peaks in the linear setting. Figures \ref{fig:Transformation_Function}d) and \ref{fig:Transformation_Function}e) show that DAGMA's and NoTears \ac{MLP}'s second cluster, which is associated with \ac{DOS} values higher than 0.8, becomes bigger with increasing ReLU causal transformations. While DAGMA and NoTears \ac{MLP} do not make any assumptions about the distribution of the passively observed data, the models are designed to handle nonlinear \ac{SEM}. The peaks of GOLEM, DAGMA, and NoTears \ac{MLP} are associated with high \ac{DOS} values mostly resulting from test runs, which make use of the original scale of the passively observed dataset. By contrast, the clusters related to \ac{DOS} values lower than approximately 0.5 are associated with graphical structures extracted from re-scaled attributes. This highlights the tendency of linear (nonlinear) causal discovery models to overfit linear (nonlinear) settings when no changes are performed to the scale of the original dataset. The hidden dependency in some settings between the data scale and the causal transformation function implies that future research should focus on designing simulation frameworks, which explicitly incorporate re-scaling operations when computing nonlinear \ac{SEM}. In contrast to GOLEM, DAGMA, and NoTears \ac{MLP}, the bimodal distributions of \ac{DOS} values produced by $R^2$-SortnRegress in Figure \ref{fig:Transformation_Function}a) are resulting from the sensitivity of the approach to the graph type.\\    

With regard to the current state-of-the-art approach in the field of causal discovery, one of the key findings of our empirical analysis so far is the potential of $R^2$-SortnRegress to produce accurate graphical structures in real-life scenarios. 
\begin{figure}[h]
\centering
\includegraphics[scale=0.53]{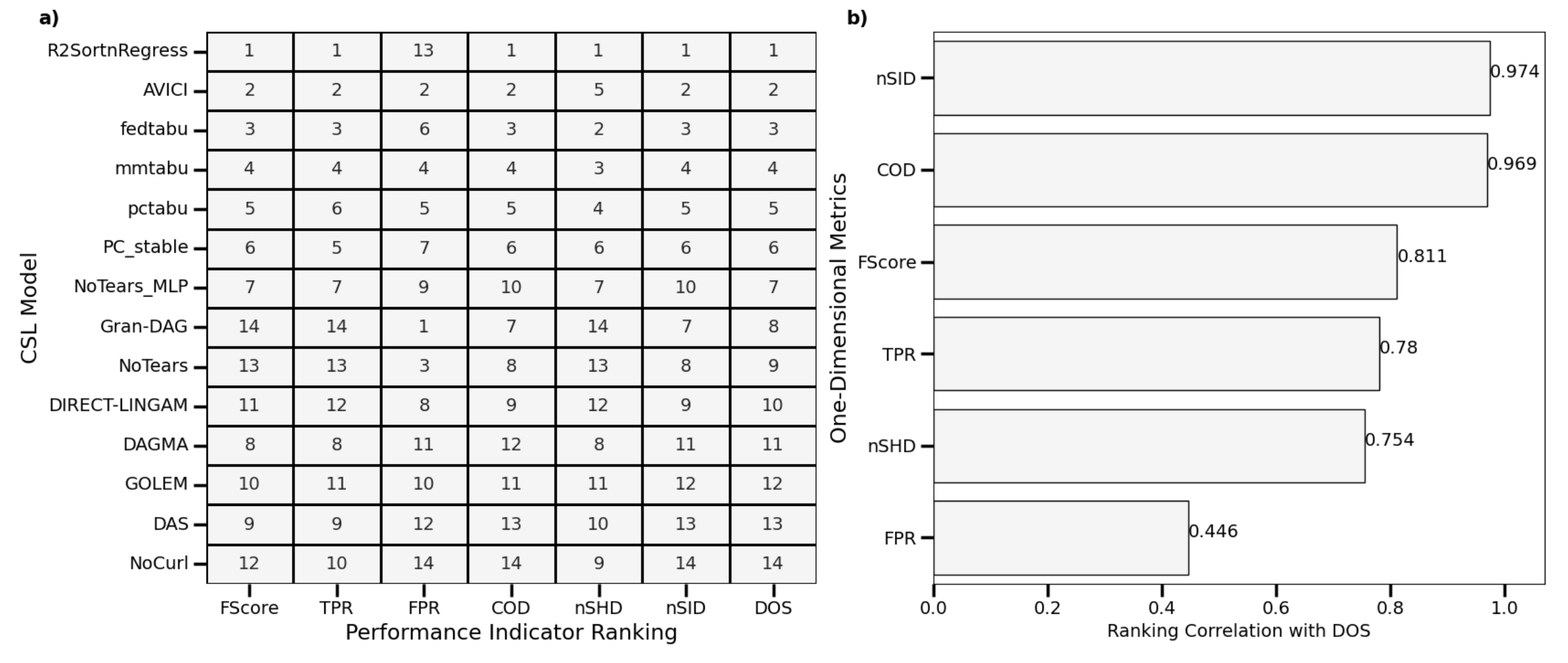}
\includegraphics[scale=0.50]{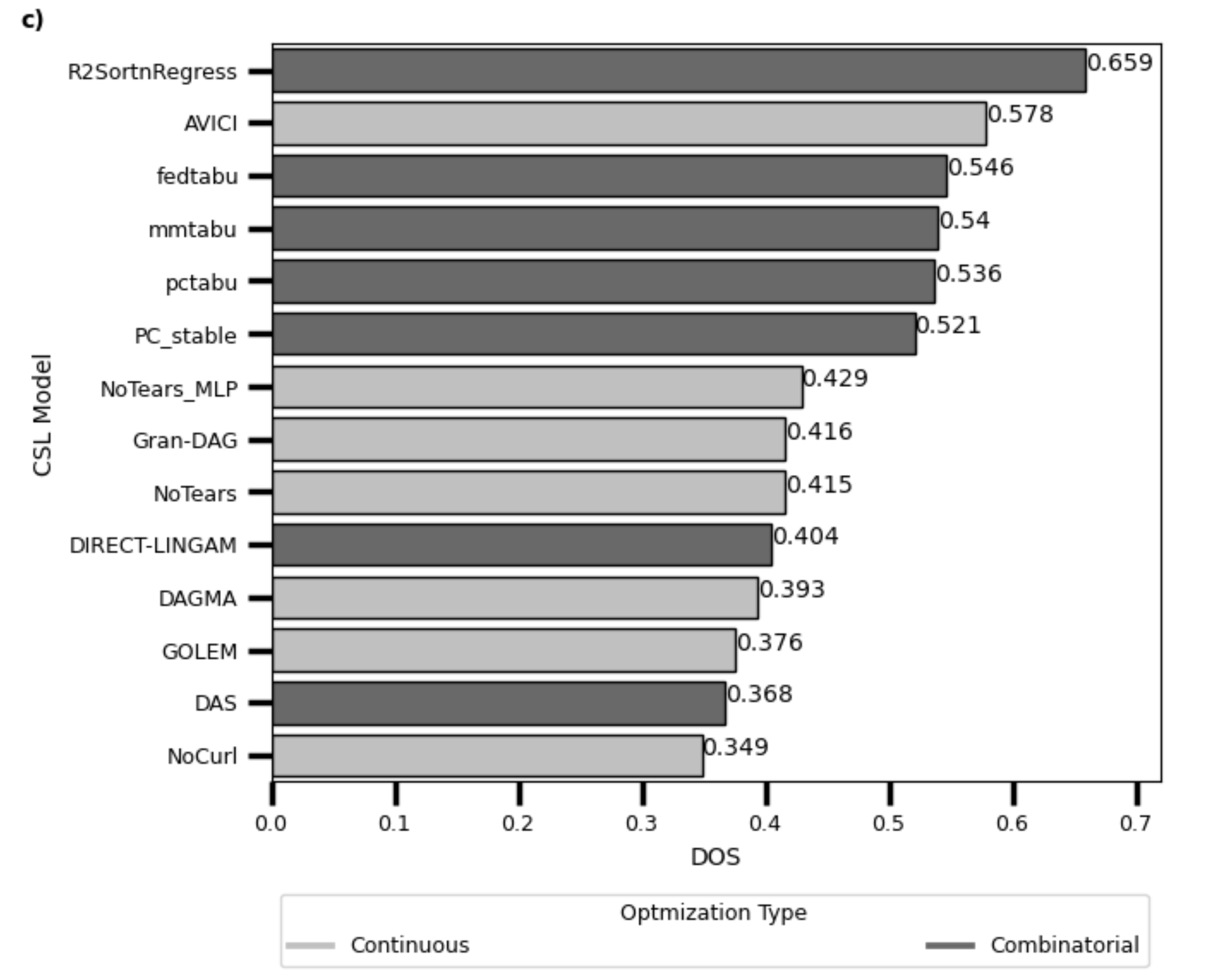}
\caption{a) Ranking of causal discovery models based on each one-dimensional evaluation criterion as well as based on the six-dimensional \ac{DOS} metric, b) spearman correlation of \ac{DOS} with each one-dimensional metric, c) barplot visualizing the magnitude of \ac{DOS} per causal discovery method.}
\label{fig:Ranking}
\end{figure}
Figure \ref{fig:Ranking}a) provides an overview of the ranking of all 14 \ac{CSL} models w.r.t. each one-dimensional evaluation metric as well as w.r.t. the integrated \ac{DOS} criterion. The model rankings are computed from test runs on re-scaled data only, to account for the severity of the varsortability problem in the field of causal discovery. In Appendix \ref{sec:Appendix6}, we highlight differences in the ranking of structure learning models resulting from changes in the data scale. Figure \ref{fig:Ranking}a) shows that on five out of six one-dimensional evaluation metrics, the $R^2$-based sorting method outperforms all other \ac{CSL} techniques. We suspect that the reason for the poor performance of $R^2$-SortnRegress in terms of FPR is related to the edge pruning step of the algorithm, which utilizes the coefficients estimated by linear regression models. The big differences in the rankings of NoTears and Gran-\ac{DAG} based on the one-dimensional metrics highlight the advantage of utilizing the integrated metric \ac{DOS} for the performance evaluation. The bar plots in Figure \ref{fig:Ranking}c) show that amortized variational inference is among the most promising continuous optimization causal discovery techniques. Nonetheless, utilizing a pre-trained transformer-based model for structure learning results in approximately 10\%\ decrease in terms of \ac{DOS} on average in comparison to the results obtained from the combinatorial optimization approach $R^2$-SortnRegress. Furthermore, hybrid bayesian networks outperform most of the continuous optimization methods included in this research except AVICI. This highlights the contribution of our benchmark study. As shown in Section \ref{subsec:SFECR}, hybrid bayesian networks are often omitted from the set of baseline techniques, when a novel causal discovery approach is introduced. Likewise, the recently published study of \citet{BNS_AssumptionViolations}, which examines the performance of \ac{CSL} techniques in misspecified scenarios, does not include any of the five best performing models in Figure \ref{fig:Ranking}c). The latter shows that six out of seven continuous optimization models achieved \ac{DOS} values, which are lower than 0.5 on average. Since \ac{DOS} $\in [0,1]$, a performance of less than 0.5 is indicative of the poor quality of the graphical structures, that most gradient-based techniques are expected to estimate in real-life scenarios.\\

The bar plots in Figure \ref{fig:Ranking}b) show that overall the ranking produced by \ac{DOS} is highly correlated with the rankings based on n\ac{COD} and n\ac{SID}. We suspect that the rank-based correlation results from assigning equal weights to each one-dimensional evaluation criterion for the computation of the integrated metric. Thus, manipulating these weights would result in a slightly different correlation with \ac{DOS}. Both metrics n\ac{COD} and n\ac{SID} play a pivotal role in the evaluation of the estimated causal \ac{DAG}s. n\ac{COD} quantifies the mistakes in the computed causal order, which represents the backbone of the estimated graph, and n\ac{SID} quantifies the capacity of the estimated \ac{DAG}s to be utilized in the context of causal inference. The fact that n\ac{COD} and n\ac{SID} are often not reported by studies presenting novel or benchmarking existing methods, as shown in Sections \ref{subsec:SFECR} and \ref{subsec:RelatedBenchmarks}, underlines the contribution of designing a novel, multi-dimensional performance indicator, which incorporates, among others, both n\ac{COD} and n\ac{SID}.\\

\section{Conclusion}
In this paper, we shed light on the performance of 14 causal discovery techniques for i.i.d. passively observed data, which we simulate with previously unexplored nonlinear, non-identifiable causal transformations. The results obtained from a total of 215.040 model runs are evaluated with a novel, interpretable, multi-dimensional performance assessment framework. The first component of our framework, i.e., \ac{DOS}, quantifies the structural similarities of the estimated \ac{DAG}s with the ground truth as well as the capacity of the discovered graphs for causal inference. The second component, i.e., \ac{EBM}, facilitates the sensitivity analysis w.r.t. different values of each of the seven experimental factors of our simulation framework as well as w.r.t. pairwise interaction effects. The latter were particularly useful in discovering settings, in which the performance of \ac{CSL} techniques showed notable variation w.r.t. the interaction between data scale and other experimental factors. This finding highlights both the severity of the problem of varsortability in causal discovery and the necessity to evaluate \ac{CSL} models on re-scaled data. In this regard, ranking the 14 approaches based on the results obtained from standardized data showed that hybrid bayesian networks, which are rarely included in the set of benchmark models in studies introducing methodological innovations, outperform most recently introduced continuous optimization techniques except the pre-trained transformer-based AVICI. The current state-of-the-art approach for structure learning from i.i.d. observational data, i.e., $R^2$-SortnRegress, which achieves an average \ac{DOS} of slightly more than 65\%\ , shows that the most promising computation of the backbone of the whole graph is based on the accumulation of the explainable variance along the causal order.\\

Concerning future research directions, we highlight that the simulation framework in our study is limited to causal transformations using a specific nonlinear formulation, i.e., ReLU-based parent-child transformations. Therefore, it would be useful to explore the performance of structure learning techniques on observational data with mixed nonlinear causal patterns, to account for the potentially more complex environment in real-life settings. Future research could also focus on providing similar insights about the performance of bayesian causal discovery models in terms of \ac{DOS} on a large scale. Last but not least, an extension of our \ac{DOS}-based evaluation framework for graphs extracted from time series data would deliver a significant contribution in terms of interpretability and comprehensiveness of the performance assessment of structure learning approaches designed to handle sequential data.   
\section*{Acknowledgments}
Stefan Lessmann acknowledges financial support through the project “AI4EFin AI for Energy Finance”, contract number CF162/15.11.2022, financed under Romania’s National Recovery and Resilience Plan, Apel nr. PNRR-III-C9-2022-I8.\\

\section*{Data and Code Availability}
The simulated datasets, the evaluation of the estimated adjacency matrices and a dataframe containing a summary of all results is available at \url{https://drive.google.com/drive/u/4/folders/1UOlMKeqokCwFhF1AkyeMdAKLtE46JfXa}. The code for reproducing our experiments with instructions how to utilize our interpretable, multi-dimensional evaluation framework is available at \url{https://github.com/gvelev123/Causal_Discovery_iid_Data}.

\newpage
\bibliography{sample}

\begin{thebibliography}{53}
\providecommand{\natexlab}[1]{#1}
\providecommand{\url}[1]{\texttt{#1}}
\expandafter\ifx\csname urlstyle\endcsname\relax
  \providecommand{\doi}[1]{doi: #1}\else
  \providecommand{\doi}{doi: \begingroup \urlstyle{rm}\Url}\fi

\bibitem[Barabasi and Albert(1999)]{Barabasi}
A.-L. Barabasi and R.~Albert.
\newblock {Emergence of scaling in random networks}.
\newblock \emph{Science}, 286, 1999.

\bibitem[Bello et~al.(2022)Bello, Aragam, and Ravikumar]{DAGMA}
K.~Bello, B.~Aragam, and P.~Ravikumar.
\newblock {DAGMA: Learning DAGs via M-matrices and a Log-Determinant Acyclicity
  Characterization}.
\newblock In \emph{{Neural Information Processing Systems}}, 2022.

\bibitem[Brouillard et~al.(2020)Brouillard, Lachapelle, Lacoste,
  Lacoste-Julien, and Drouin]{ENCO_SEM}
P.~Brouillard, S.~Lachapelle, A.~Lacoste, S.~Lacoste-Julien, and A.~Drouin.
\newblock {Differentiable Causal Discovery from Interventional Data}.
\newblock In \emph{{Neural Information Processing Systems}}, 2020.

\bibitem[Bühlmann et~al.(2014)Bühlmann, Peters, and Ernest]{CAM}
P.~Bühlmann, J.~Peters, and J.~Ernest.
\newblock {CAM: Causal additive models, high-dimensional order search and
  penalized regression.}
\newblock In \emph{{Annals of Statistics}}, 2014.

\bibitem[Chernozhukov et~al.(2013)Chernozhukov, Fernández‐Val, and
  Melly]{CFDistributions}
V.~Chernozhukov, I.~Fernández‐Val, and B.~Melly.
\newblock {Inference on counterfactual distributions}.
\newblock \emph{Econometrica}, 81, 2013.

\bibitem[Colombo and Maathuis(2014)]{PC_Stable}
D.~Colombo and M.~H. Maathuis.
\newblock {Order-independent constraint-based causal structure learning}.
\newblock \emph{Machine Learning Research}, 15, 2014.

\bibitem[Constantinou et~al.(2023)Constantinou, Kitson, Liu, Chobtham,
  Amirkhizi, Nanavati, Mbuvha, and Petrungaro]{BNS_CovidCSL}
A.~Constantinou, NK. Kitson, Y.~Liu, K.~Chobtham, AH. Amirkhizi, PA. Nanavati,
  R.~Mbuvha, and B.~Petrungaro.
\newblock {Open problems in causal structure learning: A case study of COVID-19
  in the UK.}
\newblock \emph{Expert Systems with Applications}, 324, 2023.

\bibitem[Donge and Jiwane(2023)]{EBM}
L.~Donge, K.and~Yadav and N.~Jiwane.
\newblock {InterpretML: A Unified Framework for Machine Learning
  Interpretability}.
\newblock \emph{Advanced Research in Computer and Communication Engineering},
  12, 2023.

\bibitem[Göbler et~al.(2024)Göbler, Windisch, Drton, Pychynski, Roth, and
  Sonntag]{BNS_CausalAssembly}
K.~Göbler, T.~Windisch, M.~Drton, T.~Pychynski, M.~Roth, and S.~Sonntag.
\newblock {CausalAssembly: Generating Realistic Production Data for
  Benchmarking Causal Discovery.}
\newblock In \emph{{ Causal Learning and Reasoning}}, 2024.

\bibitem[Hasan et~al.(2023)Hasan, Hossain, and Gani]{BNS_IIDTimeSeries}
U.~Hasan, E.~Hossain, and M.~O. Gani.
\newblock {A survey on causal discovery methods for iid and time series data}.
\newblock \emph{Transactions on Machine Learning Research}, 09, 2023.

\bibitem[Hoyer et~al.(2008)Hoyer, Janzing, Mooij, Peters, and Schölkopf]{NPM}
P.~Hoyer, D.~Janzing, J.M. Mooij, J.~Peters, and B.~Schölkopf.
\newblock {Nonlinear causal discovery with additive noise models}.
\newblock In \emph{{Neural Information Processing Systems}}, 2008.

\bibitem[Huang et~al.(2021)Huang, Kleindessner, Munishkin, Varshney, Guo, and
  Wang]{BNS_DataDriven}
Y.~Huang, M.~Kleindessner, A.~Munishkin, D.~Varshney, P.~Guo, and J.~Wang.
\newblock {Benchmarking of data-driven causality discovery approaches in the
  interactions of arctic sea ice and atmosphere}.
\newblock \emph{Frontiers in big Data}, 4, 2021.

\bibitem[Huegle et~al.(2021)Huegle, Hagedorn, Böhme, Pörschke, Umland, and
  Schlosser]{MANM_CS}
J.~Huegle, C.~Hagedorn, L.~Böhme, M.~Pörschke, J.~Umland, and R.~Schlosser.
\newblock {MANM-CS: Data Generation for Benchmarking Causal Structure Learning
  from Mixed Discrete-Continuous and Nonlinear Data}.
\newblock In \emph{{Neural Information Processing Systems}}, 2021.

\bibitem[Kalainathan et~al.(2022)Kalainathan, Goudet, Guyon, Lopez-Paz, and
  Sebag]{SAM}
D.~Kalainathan, O.~Goudet, I.~Guyon, D.~Lopez-Paz, and M.~Sebag.
\newblock {Structural agnostic modeling: Adversarial learning of causal
  graphs.}
\newblock \emph{Machine Learning Research}, 23, 2022.

\bibitem[Keropyan et~al.(2023)Keropyan, Strieder, and Drton]{PNL_Rankbased}
G.~Keropyan, D.~Strieder, and M.~Drton.
\newblock {Rank-Based Causal Discovery for Post-Nonlinear Models}.
\newblock In \emph{{Proceedings of International Conference on Artificial
  Intelligence and Statistics}}, 2023.

\bibitem[Koller and Friedman(2009)]{BookIntQueries}
D.~Koller and N.~Friedman.
\newblock \emph{{Probabilistic graphical models: principles and techniques}}.
\newblock MIT press, 2009.

\bibitem[Lachapelle et~al.(2020)Lachapelle, Brouillard, Deleu, and
  Lacoste-Julien]{GranDAG}
S.~Lachapelle, P.~Brouillard, T.~Deleu, and S.~Lacoste-Julien.
\newblock {Gradient-based neural dag learning}.
\newblock In \emph{{International Conference on Learning Representations}},
  2020.

\bibitem[Li et~al.(2019)Li, Cabeli, Sella, and Isambert]{PC_Consistent}
H.~Li, V.~Cabeli, N.~Sella, and H.~Isambert.
\newblock {Constraint-based Causal Structure Learning with Consistent
  Separating Sets}.
\newblock In \emph{{Neural Information Processing Systems}}, 2019.

\bibitem[Lippe et~al.(2022)Lippe, Cohen, and Gavves]{ENCO}
P.~Lippe, T.~Cohen, and E.~Gavves.
\newblock {Efficient neural causal discovery without acyclicity constraints}.
\newblock In \emph{{International Conference on Learning Representations}},
  2022.

\bibitem[Lorch et~al.(2021)Lorch, Rothfuss, Schölkopf, and Krause]{DIBS}
L.~Lorch, J.~Rothfuss, B.~Schölkopf, and A.~Krause.
\newblock {DiBS: Differentiable Bayesian Structure Learning}.
\newblock In \emph{{Neural Information Processing Systems}}, 2021.

\bibitem[Lorch et~al.(2022)Lorch, Sussex, Rothfuss, Krause, and
  Schölkopf]{AVICI}
L.~Lorch, S.~Sussex, J.~Rothfuss, A.~Krause, and B.~Schölkopf.
\newblock {Amortized Inference for Causal Structure Learning}.
\newblock In \emph{{Neural Information Processing Systems}}, 2022.

\bibitem[Menegozzo et~al.(2022)Menegozzo, Dall’Alba, and Fiorini]{BNS_CIPCAD}
G.~Menegozzo, D.~Dall’Alba, and P.~Fiorini.
\newblock {Cipcad-bench: Continuous industrial process datasets for
  benchmarking causal discovery methods.}
\newblock In \emph{{IEEE 18th International Conference on Automation Science
  and Engineering}}, 2022.

\bibitem[Montagna et~al.(2023{\natexlab{a}})Montagna, Mastakouri, Eulig,
  Noceti, Rosasco, Janzing, Aragam, and Locatello]{BNS_AssumptionViolations}
F.~Montagna, A.~Mastakouri, E.~Eulig, N.~Noceti, L.~Rosasco, D.~Janzing,
  B.~Aragam, and F.~Locatello.
\newblock {Assumption violations in causal discovery and the robustness of
  score matching}.
\newblock In \emph{{Neural Information Processing Systems}},
  2023{\natexlab{a}}.

\bibitem[Montagna et~al.(2023{\natexlab{b}})Montagna, Noceti, Rosasco, Zhang,
  and Locatello]{DAS}
F.~Montagna, N.~Noceti, L.~Rosasco, K.~Zhang, and F.~Locatello.
\newblock {Scalable causal discovery with score matching.}
\newblock In \emph{{Causal Learning and Reasoning}}, 2023{\natexlab{b}}.

\bibitem[Montagna et~al.(2023{\natexlab{c}})Montagna, Noceti, Rosasco, Zhang,
  and Locatello]{NoGAM}
F.~Montagna, N.~Noceti, L.~Rosasco, K.~Zhang, and F.~Locatello.
\newblock {Causal discovery with score matching on additive models with
  arbitrary noise.}
\newblock In \emph{{Causal Learning and Reasoning}}, 2023{\natexlab{c}}.

\bibitem[Ng et~al.(2020)Ng, Ghassami, and Zhang]{GOLEM}
I.~Ng, A.~Ghassami, and K.~Zhang.
\newblock {On the Role of Sparsity and DAG Constraints for Learning Linear
  DAGs}.
\newblock In \emph{{Neural Information Processing Systems}}, 2020.

\bibitem[Nogueira et~al.(2022)Nogueira, Pugnana, Ruggieri, Pedreschi, and
  Gama]{BNS_Wiley}
A.~R. Nogueira, A.~Pugnana, S.~Ruggieri, D.~Pedreschi, and J.~Gama.
\newblock {Methods and tools for causal discovery and causal inference}.
\newblock \emph{Data mining and knowledge discovery}, 12, 2022.

\bibitem[Peters and Bühlmann(2014)]{GaussianANMNonEqualVar}
J.~Peters and P.~Bühlmann.
\newblock {Identifiability of Gaussian structural equation models with equal
  error variances}.
\newblock \emph{{Biometrika}}, 101, 2014.

\bibitem[Peters and Bühlmann(2015)]{SID}
J.~Peters and P.~Bühlmann.
\newblock {Structural intervention distance for evaluating causal graphs.}
\newblock \emph{Neural computation}, 27, 2015.

\bibitem[Peters et~al.(2014)Peters, Mooij, Janzing, and
  Schölkopf]{CSLadditive}
J.~Peters, J.~M. Mooij, D.~Janzing, and B.~Schölkopf.
\newblock {Causal Discovery with Continuous Additive Noise Models}.
\newblock \emph{Machine Learning Research}, 15, 2014.

\bibitem[Peters et~al.(2017)Peters, Janzing, and Schölkopf]{CSLbook}
J.~Peters, D.~Janzing, and B.~Schölkopf.
\newblock \emph{{Elements of causal inference: foundations and learning
  algorithms}}.
\newblock MIT press, 2017.

\bibitem[Reisach et~al.(2021)Reisach, Seiler, and Weichwald]{Be_Ware}
A.~Reisach, C.~Seiler, and S.~Weichwald.
\newblock {Beware of the simulated dag! causal discovery benchmarks may be easy
  to game}.
\newblock In \emph{{Neural Information Processing Systems}}, 2021.

\bibitem[Reisach et~al.(2023)Reisach, Tami, Seiler, Chambaz, and
  Weichwald]{RSortability}
A.~Reisach, M.~Tami, C.~Seiler, A.~Chambaz, and S.~Weichwald.
\newblock {A Scale-Invariant Sorting Criterion to Find a Causal Order in
  Additive Noise Models}.
\newblock In \emph{{Neural Information Processing Systems}}, 2023.

\bibitem[Rolland et~al.(2022)Rolland, Cevher, Kleindessner, Russell, Janzing,
  Schölkopf, and Locatello]{SCORE}
P.~Rolland, V.~Cevher, M.~Kleindessner, C.~Russell, D.~Janzing, B.~Schölkopf,
  and F.~Locatello.
\newblock {Score matching enables causal discovery of nonlinear additive noise
  models.}
\newblock In \emph{{International Conference on Machine Learning}}, 2022.

\bibitem[Sanchez et~al.(2023)Sanchez, Liu, O'Neil, and Tsaftaris]{DiffCSL}
P.~Sanchez, X.~Liu, A.~Q. O'Neil, and S.~A. Tsaftaris.
\newblock {Diffusion Models for Causal Discovery via Topological Ordering}.
\newblock In \emph{{International Conference on Learning Representations}},
  2023.

\bibitem[Shimizu et~al.(2006{\natexlab{a}})Shimizu, Hoyer, Hyvärinen,
  Kerminen, and Jordan]{ICA_LINGAM}
S.~Shimizu, P.~O. Hoyer, A.~Hyvärinen, A.~Kerminen, and M.~Jordan.
\newblock {A linear non-Gaussian acyclic model for causal discovery.}
\newblock \emph{Machine Learning Research}, 7, 2006{\natexlab{a}}.

\bibitem[Shimizu et~al.(2006{\natexlab{b}})Shimizu, Hoyer, Hyvärinen,
  Kerminen, and Jordan]{NonGaussianANM}
S.~Shimizu, P.~O. Hoyer, A.~Hyvärinen, A.~Kerminen, and M.~Jordan.
\newblock {A Linear Non-Gaussian Acyclic Model for Causal Discovery}.
\newblock \emph{Machine Learning Research}, 7, 2006{\natexlab{b}}.

\bibitem[Shimizu et~al.(2011)Shimizu, Takanori, Yasuhiro, Aapo, Yoshinobu,
  Takashi, Patrik~O., Kenneth, and Patrik]{DIRECT_LINGAM}
S.~Shimizu, I.~Takanori, S.~Yasuhiro, H.~Aapo, K.~Yoshinobu, W.~Takashi,
  H.~Patrik~O., B.~Kenneth, and H.~Patrik.
\newblock {DirectLiNGAM: A direct method for learning a linear non-Gaussian
  structural equation model.}
\newblock \emph{Machine Learning Research}, 12, 2011.

\bibitem[Smith(2019)]{GAE}
Jane Smith.
\newblock {A Graph Autoencoder Approach to Causal Structure Learning}.
\newblock In \emph{{Neural Information Processing Systems}}, 2019.

\bibitem[Spirtes et~al.(2000)Spirtes, Glymour, and Scheines]{PC_original}
P.~Spirtes, C.~Glymour, and R.~Scheines.
\newblock {Causation, Prediction, and Search}.
\newblock In \emph{{The MIT Press}}, 2000.

\bibitem[Tsagris(2021)]{PCHC}
M.~Tsagris.
\newblock {A new scalable Bayesian network learning algorithm with applications
  to economics.}
\newblock \emph{Computational Economics}, 57, 2021.

\bibitem[Tsagris(2022)]{FEDHC}
M~Tsagris.
\newblock {The FEDHC Bayesian network learning algorithm.}
\newblock \emph{Mathematics}, 10, 2022.

\bibitem[Tsamardinos et~al.(2006)Tsamardinos, Brown, and Aliferis]{MMHC}
I.~Tsamardinos, L.~E. Brown, and C.~F. Aliferis.
\newblock {The max-min hill-climbing Bayesian network structure learning
  algorithm.}
\newblock \emph{Machine learning}, 65, 2006.

\bibitem[Uemura et~al.(2022)Uemura, Takagi, Takayuki, Yoshida, and
  Shimizu]{PNL_Second_Multivariate}
K.~Uemura, T.~Takagi, K.~Takayuki, H.~Yoshida, and S.~Shimizu.
\newblock {A multivariate causal discovery based on post-nonlinear model}.
\newblock In \emph{{Proceedings of Machine Learning Research}}, 2022.

\bibitem[Wang et~al.(2024)Wang, Huang, Wang, Liao, Li, and Liu]{BNS_SurveyFCM}
L.~Wang, S.~Huang, S.~Wang, J.~Liao, T.~Li, and L.~Liu.
\newblock {A survey of causal discovery based on functional causal model}.
\newblock \emph{Engineering Applications of Artificial Intelligence}, 133,
  2024.

\bibitem[Yu et~al.(2021{\natexlab{a}})Yu, Gao, Yin, and Ji]{DAG_GNN}
Y.~Yu, T.~Gao, N.~Yin, and Q.~Ji.
\newblock {DAG-GNN: DAG Structure Learning with Graph Neural Networks}.
\newblock In \emph{{International Conference on Machine Learning}},
  2021{\natexlab{a}}.

\bibitem[Yu et~al.(2021{\natexlab{b}})Yu, Gao, Yin, and Ji]{NoCurl}
Y.~Yu, T.~Gao, N.~Yin, and Q.~Ji.
\newblock {DAGs with No Curl: An Efficient DAG Structure Learning Approach}.
\newblock In \emph{{International Conference on Machine Learning}},
  2021{\natexlab{b}}.

\bibitem[Zhang and Hyvarinen(2009)]{PNL_Indentify}
K.~Zhang and A.~Hyvarinen.
\newblock {On the Identifiability of the Post-Nonlinear Causal Model}.
\newblock In \emph{{Annual Conference on Learning Theory}}, 2009.

\bibitem[Zhang and Hyvärinen(2010)]{PNL_Original}
K.~Zhang and Hyvärinen.
\newblock {Distinguishing causes from effects using nonlinear acyclic causal
  models}.
\newblock In \emph{{Workshop and Conference Proceedings of Machine Learning
  Research}}, 2010.

\bibitem[Zheng et~al.(2018)Zheng, Aragam, Ravikumar, and Xing]{NoTears_Linear}
X.~Zheng, B.~Aragam, P.~K. Ravikumar, and E.~P. Xing.
\newblock {DAGs with NO TEARS: Continuous Optimization for Structure Learning}.
\newblock In \emph{{Neural Information Processing Systems}}, 2018.

\bibitem[Zheng et~al.(2020)Zheng, Dan, Aragam, Ravikumar, and
  Xing]{NoTears_nonlinear}
X.~Zheng, C.~Dan, B.~Aragam, P.~Ravikumar, and E.~Xing.
\newblock {Learning sparse nonparametric DAGs}.
\newblock In \emph{{International Conference on Artificial Intelligence and
  Statistics}}, 2020.

\bibitem[Zhu et~al.(2020)Zhu, Ng, and Chen]{RL}
S.~Zhu, I.~Ng, and Z.~Chen.
\newblock {Causal discovery with reinforcement learning}.
\newblock In \emph{{International Conference on Learning Representations}},
  2020.

\bibitem[Zlaugotne et~al.(2020)Zlaugotne, Zihare, Balode, Kalnbalkite,
  Khabdullin, and Blumberga]{MCDM}
B.~Zlaugotne, L.~Zihare, L.~Balode, A.~Kalnbalkite, A.~Khabdullin, and
  D.~Blumberga.
\newblock {Multi-criteria decision analysis methods comparison.}
\newblock \emph{Environmental and Climate Technologies}, 24, 2020.

\end{thebibliography}

\newpage

\appendix
\appendixpage
\begin{appendices}
\section{\ac{CI} Relationships}
\label{sec:Appendix1}
\begin{figure}[h]
\centering
\includegraphics[scale=0.55]{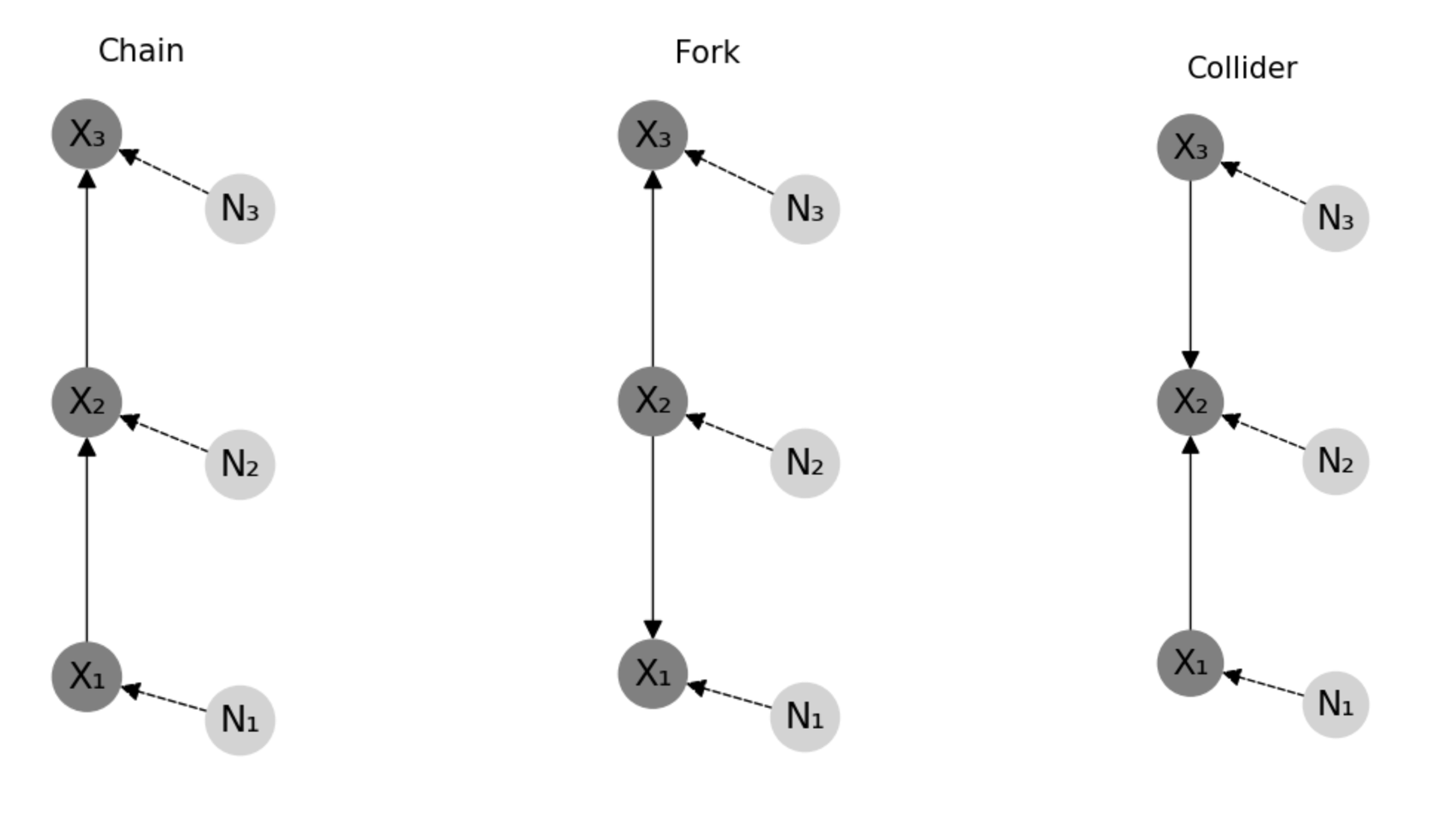}
\caption{Basic causal structures of \ac{DAG}s}
\label{fig:BaiscCausalStructures}
\end{figure}

In this section, we exemplify the concept of \ac{CI} relationships in three basic causal structures, i.e., chains, forks, and colliders, which are visualized in Figure \ref{fig:BaiscCausalStructures}. We also clarify the connection between \ac{CI} relations and Pearl's d-separation criterion \citep{PC_Stable,CSLbook}.
In each of the three graphical structures, the child nodes are caused by their parents and by the i.i.d. noise terms $N_i$. The chain and the the fork on the left side and in the middle of Figure \ref{fig:BaiscCausalStructures}, respectivelly, both visualize causal structures, in which the nodes $X_1$ and $X_3$ are said to be dependent. While in chains this is due to changes in $X_1$ causing variations in $X_2$, which in turn have a direct impact on $X_3$, in forks the dependence between $X_1$ and $X_3$ results from the two vertices changing together, when variations in $X_2$ occur. However, when conditioned on the middle node $X_2$ of a chain or a fork, then $X_3$ becomes independent of $X_1$. In a chain, this results from $X_1$ not providing any new information about $X_3$, if the changes taking place in $X_2$ are already known. Thus, when holding $X_1$ constant in a chain, then variations in $X_3$ could only occur due to changes in $N_2$ and $N_3$. In a fork the \ac{CI} rule $X_3 \upvDash X_1$ | $X_2$ can be explained with changes in $X_1$ and $X_3$ occurring only due to variations in $N_2$ and $N_3$, respectively, when holding the common cause $X_2$ constant. Furthermore, for both chains and forks, if $X_2$ is part of the separating set $S$, then the path between $X_1$ and $X_3$ is said to be blocked by the middle vertex $X_2$, and $X_1$ and $X_3$ nodes become d-separated by $X_2$.\\

In contrast to chains and forks, when the middle node $X_2$ represents a collision vertex, as visualized on the right side of Figure \ref{fig:BaiscCausalStructures}, then the nodes $X_1$ and $X_3$ are said to be independent of each other. This is due to changes in $N_1$ and $N_3$ causing independent variations in $X_1$ and $X_3$, respectively. Thus, only a collider vertex acting as the common effect between a pair of nodes can block the path between these nodes, when no conditioning on a third set of vertices is performed. A further difference compared to chains and forks is that the variables $X_1$ and $X_3$ become dependent on each other when conditioning on the collision vertex. This can be explained by the fact that the values of one of the parent nodes, e.g., $X_3$ can be inferred, only if both the values of the other parent vertex $X_1$ and the values of the common effect $X_2$ are known, e.g., by holding them constant. In addition to the collider, the pair of nodes with a single path between them also become conditionally dependent on any of the descendants of the collision vertex. In the context of d-separation, the path between $X_1$ and $X_3$ can regarded as blocked by $S$ in the presence of a collider, when conditioning on $S$, only if the collision node $X_2$ and the descendants of $X_2$ are not in $S$.\\

\section{\ac{ER} and \ac{SF} Graph Models}
\label{sec:Appendix2}
\ac{ER} and \ac{SF} graph models are the most common choice of graph structures used in the field of \ac{CSL} for the simulation of synthetic data. The notation \ac{ER}$k$ and \ac{SF}$k$ is used to indicate how densely connected the sampled graphs are for a given $d$:
\begin{equation} 
e=k\times d,
\label{eq:graphcapacity}
\end{equation}
where $k$ is a scalar value set by research studies to be in $[1,4]$, and $e$ is the expected number of edges. The simulation of the causal structure using \ac{ER} graphs starts with the set of all disconnected nodes, from which cause-effect pairs are sampled with the same probability:
\begin{equation}
p=\frac{2\times e}{d^2 \minus d}
\end{equation}
Therefore, \ac{ER} graphs are characterized by a homogeneous node degree distribution, as most vertices have comparatively similar number of undirected connections. In this respect, \citet{DAGMA} suggest that the orientation of the edges of an ER graph could follow the ordering produced by a random permutation of the vertices to generate a \ac{DAG}. Contrary to the \ac{ER} method, the \ac{SF} model produces directed graphs, which have high heterogeneity. The generation of \ac{SF} graphs follows the Barabási-Albert model, according to which nodes are added one at a time based on preferential attachment \citep{Barabasi}. The latter implies that new vertices prefer to get linked to already densely connected nodes:
\begin{equation}
p (q_i)=\frac{q_i}{\sum_j q_j},
\end{equation}
where $p(q_i)$ is the probability for connecting existing node $i$ with a new node, and $q_i$ is the degree distribution of existing node $i$. In this way, the \ac{SF} model encourages the development of the so-called $hubs$, which represent a small portion of highly connected nodes in the entire graph. The formation of cycles is prevented by orienting the edges from the existing nodes to the new nodes during the sampling procedure. The sequential growth of \ac{SF} graphs combined with the preferential attachment process makes them more similar to real-world networks, e.g. social networks, than \ac{ER} graphs.\\

\section{Benchmark Studies in the field of Causal Discovery for i.i.d. observational Data}
\label{sec:AppendixBNOverview}
In this section, we provide a tabular overview of benchmarking studies published in the field of \ac{CSL}.\\

The main contributions of \citep{BNS_CausalAssembly} and \citep{BNS_CIPCAD} are semi-synthetic data-generating frameworks, which use domain expertise from the production sector to simulate realistic causal transformations. 
\begin{table}[H]
\centering
\resizebox{\textwidth}{!}{
    \begin{tabular}{|c|c|c|c|c|c|c|}
    \hline
    \multicolumn{1}{|c|}{} & \multicolumn{3}{|c|}{\textbf{Performance Evaluation}} & \multirow{2}{*}{\ip{2.7cm}{\textbf{\\Nonidentifiable\\ nonlinear\\ \ac{SCM}} } } & \multirow{2}{*}{\ip{2.5cm}{\textbf{\\Interaction\\ Effects\\ Modelling }}}& \multirow{2}{*}{\ip{2.5cm}{\textbf{\\Continuous\\ Optimization\\ Benchmarks }}}\\ 
    
    \cline{2-4}
     \textbf{Reference} & \textbf{Interpretable} &  &  &  &  &  \\
     & \textbf{Multi-Dimensional} & \textbf{\ac{SID}}  &\textbf{\ac{COD}} &  &  &  \\
      & \textbf{Criterion} &  & &  &  &  \\
    \hline
    \citep{BNS_DataDriven} & - & - & - & - & - & 2 \\ 
    \hline
    \citep{Be_Ware} & - & \cm & - & - & - & 2 \\ 
    \hline
    \citep{BNS_Wiley} & - & \cm & - & - & - & - \\ 
    \hline
    \citep{BNS_CIPCAD}& - & - & - & - & - & 5 \\ 
    \hline
    \citep{BNS_IIDTimeSeries}& - & - & - & - & - & 5 \\ 
    \hline
    \ip{3.7cm}{\citep{BNS_AssumptionViolations}} & - & - & \cm & - & - & 1 \\ 
    \hline
    \citep{BNS_SurveyFCM} & - & - & - & - & - & - \\ 
    \hline
    \citep{BNS_CausalAssembly} & - & \cm & - & - & - & 2 \\ 
    \hline
    \citep{BNS_CovidCSL} & - & - & - & - & - & 2 \\
    \hline
    \textbf{Our Study} & \cm & \cm & \cm & \cm & \cm & 7 \\
    \hline
    \end{tabular}}
\caption{Overview of studies, which benchmark the performance of a set of causal discovery models}
\label{tab:relatedBenchmarks}
\end{table}
Thus, the studies provide insights into the expected performance of structure learning models in settings similar to real-life production scenarios. However, the simulation frameworks do not explicitly incorporate non-identifiable components.\\

In the context of assumption violations, \citet{BNS_AssumptionViolations} explore the performance of causal discovery models in several misspecified scenarios, e.g., confounded data generating model, autoregressive model, data generating mechanism, which is not faithful to the ground-truth graph model, etc. In the nonlinear setting, the study simulates \ac{PNL} datasets. The underlying structural equations for the \ac{PNL} model are kept identifiable, which highlights one of the differences between our benchmark study and the research of \citet{BNS_AssumptionViolations}. In contrast to our evaluation framework, the performance assessment of structure learning techniques in misspecified settings does not incorporate \ac{SID}, despite how essential it is to quantify to what extent the estimated graphs have the capacity for causal inference under assumption violations. The setting explored in \citep{BNS_CovidCSL} bears similarities to \citep{BNS_AssumptionViolations}, as the study compares the performance of \ac{CSL} models for i.i.d. data on COVID time series attributes. The true causal graph of the real-life dataset is constructed from human knowledge. While \citet{BNS_CovidCSL} use the estimated graphs to analyze the effect of hypothetical interventions on specific variables of interest, the evaluation framework of the study does not incorporate the \ac{SID} to the assumed ground truth knowledge graph.
Only $\frac{1}{3}$ of the studies in Table \ref{tab:relatedBenchmarks} report \ac{SID}. Another commonly overlooked, and yet important one-dimensional evaluation criterion is \ac{COD}. Additionally, none of the studies in Table \ref{tab:relatedBenchmarks} attempts to formulate an interpretable, multi-dimensional performance evaluation framework, which evaluates the quality of the estimated graphs from different perspectives.\\

\section{One-dimensional Evaluation Criteria for Causal Discovery}
\label{sec:Appendix3}
In this section, we provide details about the computation of one-dimensional evaluation metrics frequently used in the field of structure learning, which we also include in our empirical research. We also highlight modifications, which we make in the formulation of some of the presented performance indicators, to improve their interpretability.\\

\ac{SHD}, which represents the most commonly selected performance indicator in causal discovery, is defined in the following way:\\
\begin{equation}
SHD=E \plus M \plus R,\\
\label{eq:SHD}
\end{equation}
where $E$ is the extra edges in the estimated causal \ac{DAG}, which are not present in the ground truth skeleton, $M$ is the edges missing from the true skeleton, and $R$ is related to the causal links present in the true \ac{DAG} with reversed direction. Thus, $SHD$ measures the total number of edges, that have to be deleted, added, and reversed to transform the estimated causal \ac{DAG} in the true graph. The magnitude of this evaluation criterion depends on the connectivity of the true \ac{DAG}, i.e., $E(G)$, and the number of estimated edges, i.e., $E(\Tilde{G})$, since $\ac{SHD} \in [0, E(G) \plus E(\Tilde{G})]$. For this reason, in our empirical research, we make use of $nSHD \in [0,1]$, which is computed in the following way:\\
\begin{equation}
nSHD=\frac{SHD}{E(G) \plus E(\Tilde{G})} ,\\
\label{eq:nSHD}
\end{equation}
The main advantage of normalizing $SHD$ is that metrics $\in [0,1]$ are easier to interpret than performance indicators, which do not have an upper bound.\\

Similarly to $nSHD$, $F1$ $score$ measures the overall quality of the estimated graph. We suspect that the reason, why $F1$ $score$ is reported by some studies published in the field of causal discovery, is related to the sparsity of the ground truth simulated \ac{DAG}s. $F1$ $score$ is suitable for the evaluation of imbalanced class distributions. Thus, applying $F1$ $score$ to the true and estimated flattened adjacency matrices, accounts for the fact that there are much more non-existing edges than the actual connections present in the simulated graphs. Since $F1$ $score$ $\in [0, 1]$, we do not make any adjustments in the formula of this metric.\\

In the context of assessment criteria $\in [0,1]$, $\ac{TPR}$ and $\ac{FPR}$ in causal discovery are defined in the following way:\\ 
\begin{equation}
TPR=\frac{TP}{T}
\label{eq:TPR}
\end{equation}
\begin{equation}
FPR=\frac{R \plus FP}{F},\\
\label{eq:FPR}
\end{equation}
where $TP$ is the set of estimated causal connections, which are also present in the true \ac{DAG}, $T$ is associated with the number of all edges in the ground truth graph, $FP$ is related to all estimated edges, which are not present in the true undirected skeleton, and $F$ is the set of the so-called non-edges in the true \ac{DAG}. While the computation of $TPR$ in causal discovery is essentially the same as the computation of this metric in a classical machine learning context, this is not the case for $FPR$, as among the two metrics in Equations \ref{eq:TPR} and \ref{eq:FPR} only $TPR \in [0,1]$. This is due to the denominator in $FPR$, i.e., $F$, which is defined as follows \citep{NoTears_Linear}:\\
\begin{equation}
F=0.5\cdot d\cdot (d \minus 1) \minus T,\\
\label{eq:F}
\end{equation}
where the scalar 0.5 pushes the whole metric to exceed the upper bound of 1.0, which $FPR$ as an evaluation criterion usually has. For this reason, in our evaluation framework, we refrain from multiplying the denominator with the scalar 0.5 and keep the remaining computation of $FPR$ as defined in Equations \ref{eq:FPR} and \ref{eq:F}.\\

The next metric, which we include in our research, quantifies the mistakes in the computation of the causal order of the estimated graph. A recently emerged trend in the field of causal discovery is concerned with the estimation of the topological order of the directed graph in the first step, which is followed by an edge-pruning phase, to reduce the number of connections implied by the causal order, e.g., \citep{SCORE,DAS,RSortability}. In this regard, \citet{SCORE} defined the causal order divergence ($COD$) as follows:\\
\begin{equation}
COD=\sum^{d}_{i=1} \sum_{j \colon \pi_{i} > \pi_{j}} A_{i,j},\\
\label{eq:COD}
\end{equation}
where $\pi$ refers to the estimated causal order, and $A$ is the ground truth binary adjacency matrix. Equation \ref{eq:COD} indicates, that looping through each of the variables in the estimated topological order and taking the sum of edges in the true graph, which are related to succeeding nodes in the computed causal order amounts to calculating the number of connections, which could not possibly be inferred in the second step of the causal discovery process due to erroneous estimation of the causal order. Similarly to $SHD$, $COD$ has an open upper bound, since $COD \in [0,E(G)]$. Thus, we formulate the normalized $COD$, as follows:\\
\begin{equation}
nCOD=\frac{COD}{E(G)} ,\\
\label{eq:nCOD}
\end{equation}

Last but not least, we include $\ac{SID}$ in the set of one-dimensional evaluation criteria, which are taken into account in our empirical research, due to the relevance of this metric in the context of causal inference. \citet{SID} define $\ac{SID}$ as follows:\\
\begin{equation}
\begin{split}
SID=\#\ \{\ (i,j), i\neq j | \textrm{\textit{ falsely estimated intervention }}\\
\textrm{ \textit{distributions by} $\Tilde{G}$ \textit{w.r.t.} $G$} \}\ ,\\
\end{split}
\label{eq:SID}
\end{equation}
For details on the computation of the falsely inferred interventional distributions, we refer the reader to the pseudo-codes in Appendix F in the study \citep{SID}. Similarly to $SHD$ and $COD$, $SID$ has an open upper bound, as $SID \in [0, d\cdot (d \minus 1)]$. Thus, to make the metric interpretable, i.e., reduce its range of possible values to [0,1], we normalize $SID$ as follows:\\
\begin{equation}
nSID=\frac{SID}{d\cdot (d \minus 1)} ,\\
\label{eq:nSID}
\end{equation}
\\

\section{Multi-dimensional Performance Indicator for Causal Discovery: \ac{DOS}}
\label{sec:Appendix7}
In this section, we provide details about the computation of \ac{DOS}. We split the calculation of our interpretable, multi-dimensional criterion overall into four steps:
\begin{itemize}
    \item \textbf{1. Step:} define the best and the worst scenario, i.e., $s_{\plus}$ and $s_{\minus}$, for the list of metrics following a specific order, e.g., $[TPR, FPR, nSHD, F score, nCOD, nSID]$, see Table \ref{tab:bestworstcase}.
    \item \textbf{2. Step:} compute the result for each test run using all six one-dimensional evaluation criteria. The normalization of the performance indicators, which $\notin [0,1]$, follows the computations presented in Appendix \ref{sec:Appendix3}.
    \item \textbf{3. Step:} compute the Euclidean distance between the results obtained from each run to the previously defined best and worst case scenario, to obtain $dist_{\plus}$ and $dist_{\minus}$.
    \item \textbf{4. Step:} compute the distance to the optimal solution per test run, i.e., $\ac{DOS}=\frac{dist_{\minus}}{dist_{\minus} \plus dist_{\plus}}$. Therefore, the smaller the distance to the best case scenario $dist_{\plus}$ is, and the higher the distance to the worst case scenario $dist_{\minus}$ is, then the higher $\ac{DOS}$ would be.
\end{itemize}

{\renewcommand{\arraystretch}{1.3}%
\begin{table}[H]
		\centering
		\begin{tabular}{|c|c|c|}
			\hline
			\textbf{Metric}  & \textbf{Best Scenario $(s_{\plus})$} &  \textbf{Worst Scenario $(s_{\minus})$} \\
                \hline
			$TPR$ & 100.0 \%\ & 0.0 \%\ \\
			\hline
                $FPR$ & 0.0 \%\ & 100.0 \%\ \\
			\hline
                $nSHD$ & 0.0 \%\ & 100.0 \%\ \\
    		\hline
                $F$ $score$ & 100.0 \%\ & 0.0 \%\ \\
			\hline
                $nCOD$ & 0.0 \%\ & 100.0 \%\ \\
			\hline
                $nSID$ & 0.0 \%\ & 100.0 \%\ \\
			\hline
\end{tabular}
\caption{Best and worst case scenario per one-dimensional metric}
\label{tab:bestworstcase}
\end{table}}

\section{Selection of \ac{EBM} Threshold for significant Variation in \ac{DOS}}
\label{sec:Appendix8}
In this section, we present details about the choice of the \ac{EBM} threshold indicating significant variation in \ac{DOS}. \\

\begin{figure}[h]
\centering
\includegraphics[scale=1.60]{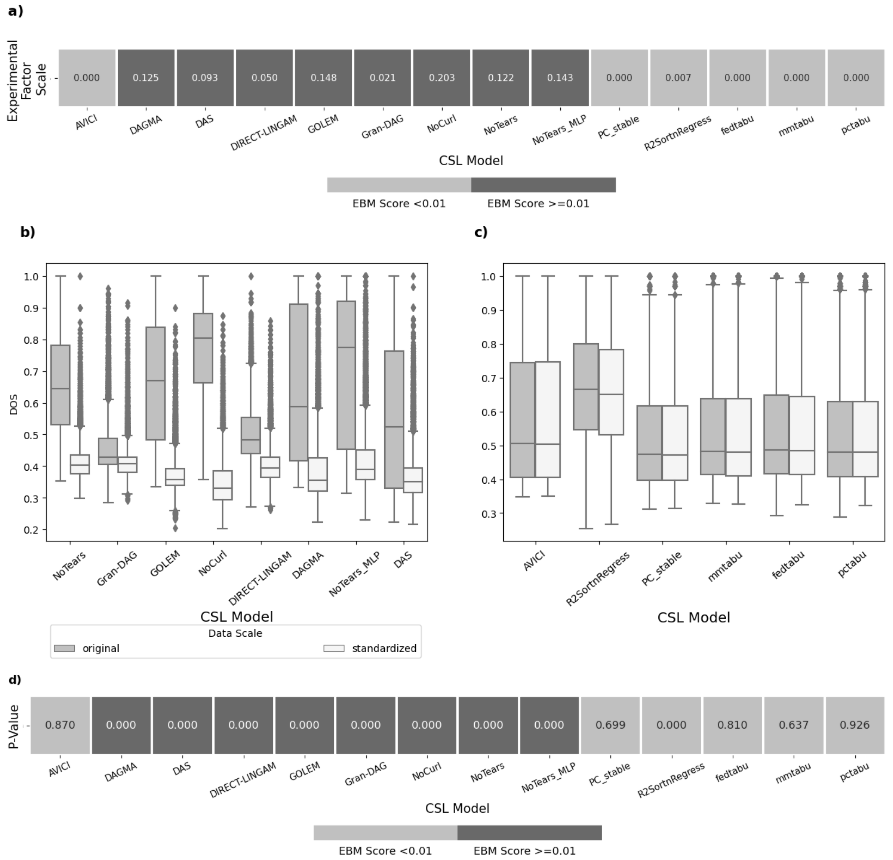}
\caption{a) \ac{EBM} importance scores for the experimental factor data scale, b) and c) conditional boxplots visualizing the variation in \ac{DOS} for \ac{EBM} scores higher or equal to and lower than 0.01, respectively, d) p-values from univariate Linear Regressions with \ac{DOS} as the target variable and data scale as the predictor attribute.}
\label{fig:EBMThresholdJustification}
\end{figure}

We select the experimental factor related to the data scale to examine the variation in \ac{DOS}. The reason for choosing the data scale is its direct connection to the problem of varsortability in causal discovery. Figure \ref{fig:EBMThresholdJustification}a) visualizes a heatmap of the \ac{EBM} scores estimated for the data scale. The coloring of the scores in the heatmap is according to the threshold 0.01, as in our empirical research we observed that overall \ac{EBM} scores lower than 0.01 are associated with comparable \ac{DOS} distributions. This can be seen in the conditional boxplots in Figures \ref{fig:EBMThresholdJustification}b) and \ref{fig:EBMThresholdJustification}c). The models in these plots are filtered based on the \ac{EBM} threshold 0.01.
The fact that the magnitude of the \ac{EBM} scores aligns with the variation in the distribution of \ac{DOS} in Fugures \ref{fig:EBMThresholdJustification}b) and \ref{fig:EBMThresholdJustification}c) highlights the suitability of utilizing the tree-based model as the second component of our evaluation framework. For instance, Gran-\ac{DAG}, Direct-LiNGAM and DAS have comparatively lower \ac{EBM} scores w.r.t data scale than the other models in Figure \ref{fig:EBMThresholdJustification}a). The differences in the interquartile range of \ac{DOS} of these models in Figure \ref{fig:EBMThresholdJustification}b) are also smaller than of the remaining models sensitive to data scale.\\

Figure \ref{fig:EBMThresholdJustification}d) visualizes the resulting p-values from univariate linear regressions of \ac{DOS} on the data scale. The latter is transformed into a binary predictor for the model fitting. The lower the p-values of the hypothesis testing for independence between the dependent and the predictor variable are, the more likely it is that changes to the scale of attributes have a significant impact on the variation in \ac{DOS}. The coloring of the p-values in Figure \ref{fig:EBMThresholdJustification}d), which is based on the \ac{EBM} scores in Figure \ref{fig:EBMThresholdJustification}a), highlights the unsuitability of utilizing such hypothesis testing tools in certain cases, e.g., $R^2$-SortnRegress. While the p-value of the causal order-based models indicates significant variation w.r.t. data scale, the conditional boxplots in Figure \ref{fig:EBMThresholdJustification}c) visualize the opposite pattern. In addition, $R^2$-SortnRegress' main contribution is scale invariance. We suspect the reason for the false signals in certain cases is related to the nonlinear computation of \ac{DOS}. This further highlights the necessity to utilize the inherently interpretable and nonlinear \ac{EBM} for the assessment of significant variation in \ac{DOS} w.r.t. various characteristics of the simulated datasets. While Figure \ref{fig:EBMThresholdJustification} highlights the variation in \ac{DOS} w.r.t. to data scale only, we could observe a similar dependency between the changes in the integrated metric and the remaining experimental factors in our simulation framework.\\

\section{Example for synthetic Data Generating Process}
\label{sec:Appendix4}
In this section, we exemplify the data-generating process to provide clarity about the connection between the different steps of the simulation framework used in our empirical research.\\

We focus  on the simulation of causal dependencies using the \ac{SF} graph model. First, the graph structure is generated, and the parameters related to the causal links in the weighted 
graph adjacency matrix $W$ are sampled. Afterward, the topological ordering of the nodes is computed. This results in a causal list, in which predecessors appear before their children \citep{DiffCSL}. Therefore, the topological order of the vertices can be computed only if there are no cycles in the graph. 

\begin{figure}[h]
\centering
\includegraphics[scale=0.55]{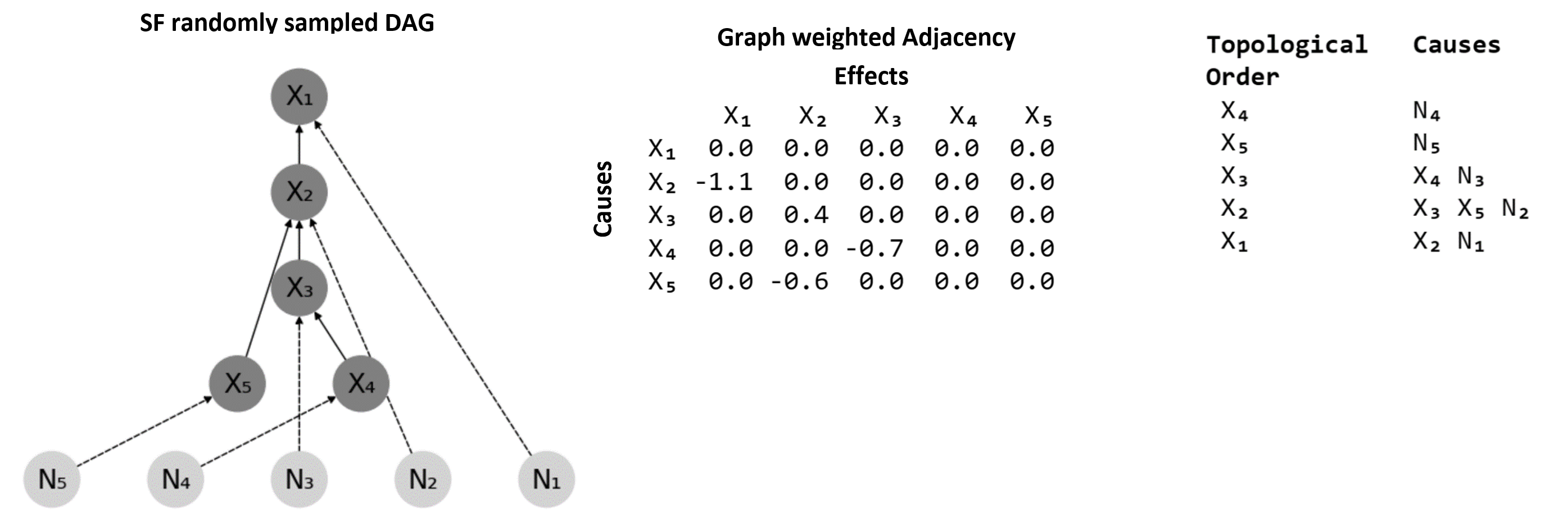}
\caption{Components of the data generation process as implemented by research studies in \ac{CSL}}
\label{fig:SimProcess}
\end{figure}

The structural equations of SCMs are sequentially applied to each node, as the simulation of the variables follows the order implied by the topological sorting. Figure \ref{fig:SimProcess} exemplifies the simulation process for a causal system with five variables.
The arrows in the visualized \ac{SF} \ac{DAG} represent the causal links. The solid edges connect a pair of nodes, which are observed in the sampled dataset. By contrast, the dashed arrows link the exogenous noise terms with the vertices. The roots of the \ac{DAG} are $X_5$ and $X_4$ with zero in-degree, and the leaf of the \ac{DAG} is $X_1$ with zero out-degree. The root nodes represent the exogenous vertices simulated as random noise, and the remaining nodes are the endogenous attributes. Given the small number of variables in this causal system, only node $X_2$ could be regarded as the $hub$ resulting from the preferential attachment process, as this vertex has the highest number of connections. The weighted adjacency matrix in the middle of Figure \ref{fig:SimProcess} is a sparse square matrix representation of the \ac{DAG}. The rows and the columns of the adjacency matrix are associated with the parent and child nodes in the \ac{DAG}, respectively. If we assume that, among the five nodes, $X_2$ and $X_3$ result from $ReLU$-based causal transformations, then the topological order of the \ac{DAG} on the right side of Figure \ref{fig:SimProcess} would indicate the following simulation using both linear and nonlinear \ac{SCM}:
\begin{equation}
\begin{aligned}
\qquad\qquad\qquad\qquad\qquad\qquad\qquad 
&X_4=N_4,\\
\qquad\qquad\qquad\qquad\qquad\qquad\qquad 
&X_5=N_5,\\
\qquad\qquad\qquad\qquad\qquad\qquad\qquad 
& X{_3}=ReLU(\minus 0.7\times X_4) \plus N_3,\\
\qquad\qquad\qquad\qquad\qquad\qquad\qquad 
&X_2=ReLU(0.3\times X_3 \minus 0.6\times X_5) \plus N_2,\\
\qquad\qquad\qquad\qquad\qquad\qquad\qquad 
&X_1=\minus 1.1\times X_2 \plus N_1
\end{aligned}
\end{equation}


The nodes $X_3$, $X_2$ and $X_1$ can be sampled using the \ac{ANM}s only after attributes $X_4$ and $X_5$ are generated.\\

\section{Impact of Beta Upper Limit on \ac{DOS}}
\label{sec:Appendix5}
In this section, we provide details about the counter-intuitive performance of some causal discovery techniques to estimate less accurate causal \ac{DAG}s with increasing beta upper limit.\\

\begin{figure}[h]
\centering
\includegraphics[scale=0.65]{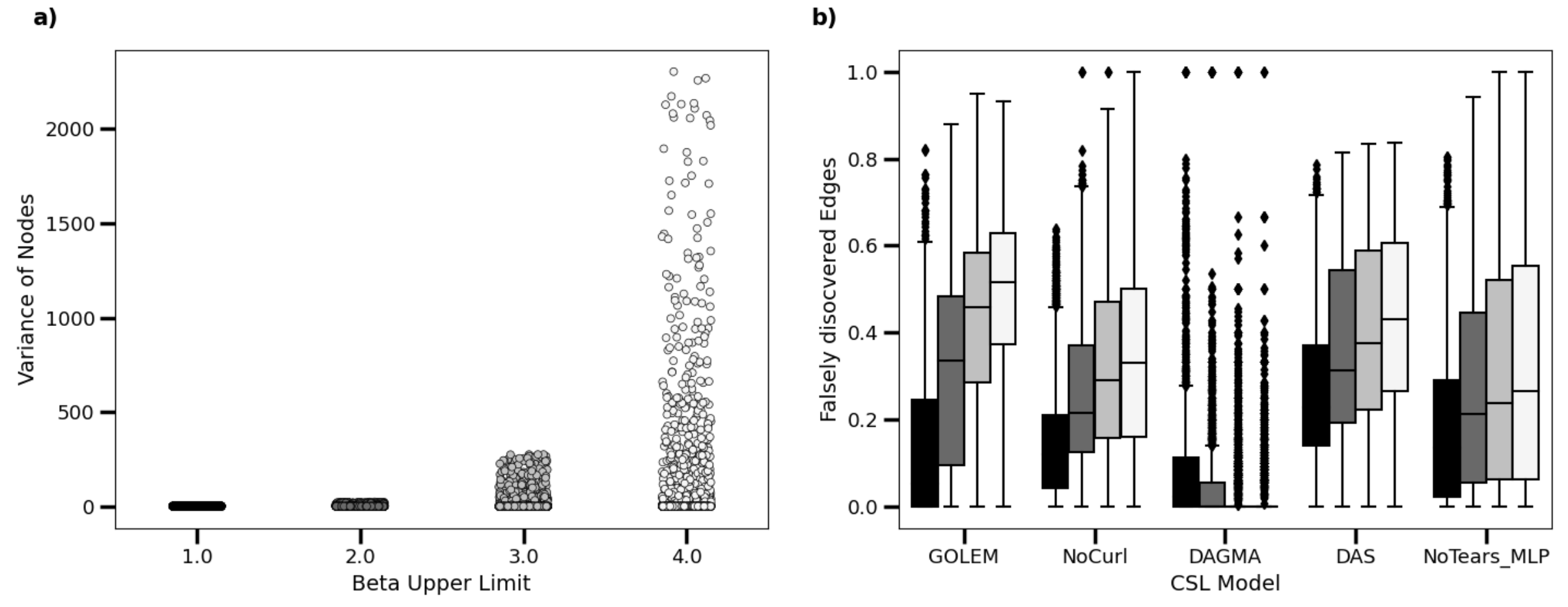}
\caption{a) The variance of passively observed nodes with an increasing beta upper limit, b) the number of falsely estimated edges, which are not existent in the true graph.}
\label{fig:additionalBetaAnalysis1}
\end{figure}

The strip plot in Figure \ref{fig:additionalBetaAnalysis1}a) shows that increasing values for the experimental factor beta upper limit overall result in an increase in the variance of the nodes in the simulated data.  
This is because during the data generation process increasing the beta upper limit implies that higher coefficients (in absolute magnitude) of the parent attributes participate in the dot product computation of the child variables. Therefore, the magnitude of the coefficients in the ground-truth adjacency matrices has a direct impact on the scale of the passively observed attributes. We suspect this is the reason why some of the models, which are sensitive to changes in the scale of the data, are also sensitive to the interaction effect of the beta upper limit with scale. Among the \ac{CSL} models in Figure \ref{fig:additionalBetaAnalysis1}b), only DAS belongs to the family of combinatorial optimization techniques. While GOLEM, NoCurl, DAGMA, and NoTears-\ac{MLP} produce a weighted adjacency matrix, the weights in which are pruned with the pre-defined threshold of 0.3, DAS applies CAM-pruning on the binary adjacency matrix estimated in the hypothesis testing phase. Figure \ref{fig:additionalBetaAnalysis1}b) shows that increasing the magnitude of the beta coefficients in absolute terms leads to a higher percentage of falsely discovered edges in most of the cases. The connection between the ground truth betas and the number of falsely predicted links is not straightforward. Since increasing the magnitude of the betas results in a higher variance of the simulated attributes, we suspect, that the increased variation in the variables of the observational datasets results in a higher spread of the estimated coefficients in the continuous adjacency matrices produced by GOLEM, NoCurl and NoTears-\ac{MLP}. The higher the variation of the estimated beta weights is, the more likely it would be that more of the predicted betas in absolute magnitude exceed the pre-defined pruning threshold of 0.3, which is applied at the end of the causal discovery process. This in turn would result in estimating a higher number of edges, which are not present in the true graph. \\

In comparison to GOLEM, NoCurl, and NoTears-\ac{MLP}, the post-processing step of DAS takes as inputs the binary adjacency matrix resulting from hypothesis testing for potential parent and child pairs. The cause-effect direction between the parent and the child vertices is implied by the causal order computed using the gradient hessian estimator. We suspect that increasing the beta upper limit leads to an increased number of false discoveries, which remain unpruned in the hypothesis testing step. This in turn leads to less of the falsely identified causal links getting eliminated in the final CAM-pruning step.\\

In contrast to GOLEM, NoCurl, DAS, and NoTears-\ac{MLP}, Figure \ref{fig:additionalBetaAnalysis1}b) shows that in the case of DAGMA, the opposite impact of increasing beta values can be observed on the percentage of falsely discovered edges. Similarly to GOLEM, NoCurl, and Notears-\ac{MLP}, DAGMA belongs to the family of continuous optimization models, which make use of acyclicity constraint during the \ac{CSL} training process and implement the post-processing step with the same pruning threshold. Furthermore, DAGMA and NoTears-\ac{MLP} are both Sigmoid-based \ac{MLP} models. While DAGMA optimizes log-determinant-based acyclicity constraint, NoTears-\ac{MLP} quantifies the \ac{DAG}-ness of the estimated graphs with a trace-based acyclicity function. Besides this difference, NoTears-\ac{MLP} produces the estimated adjacency matrices from the difference of \ac{MLP} trainable weights with a positive and a negative bound, i.e., the sum of two weights matrices, while DAGMA uses a single set of trainable \ac{MLP} weights associated with the functional relationship between parent and child vertices. Therefore, especially at the beginning of the causal discovery process, the beta coefficients estimated by NoTears-\ac{MLP} are expected to have higher absolute magnitude than DAGMA's coefficients. To better understand the impact of these differences on the quality of the causal \ac{DAG}s estimated by DAGMA and NoTears-\ac{MLP}, we simulated a small set of additional 40 datasets associated with beta upper limit $\in [1,4,6,8]$ and 10 different seeds per beta level. The other characteristics of the datasets are kept constant: the datasets are sampled from \ac{SF} graphs with 20 nodes and 0.3 connectivity. The datasets are simulated with 10\%\ linear and 90\%\ ReLU causal transformations applied to the original scale of the attributes.\\

\begin{figure}[h]
\centering
\includegraphics[scale=1.60]{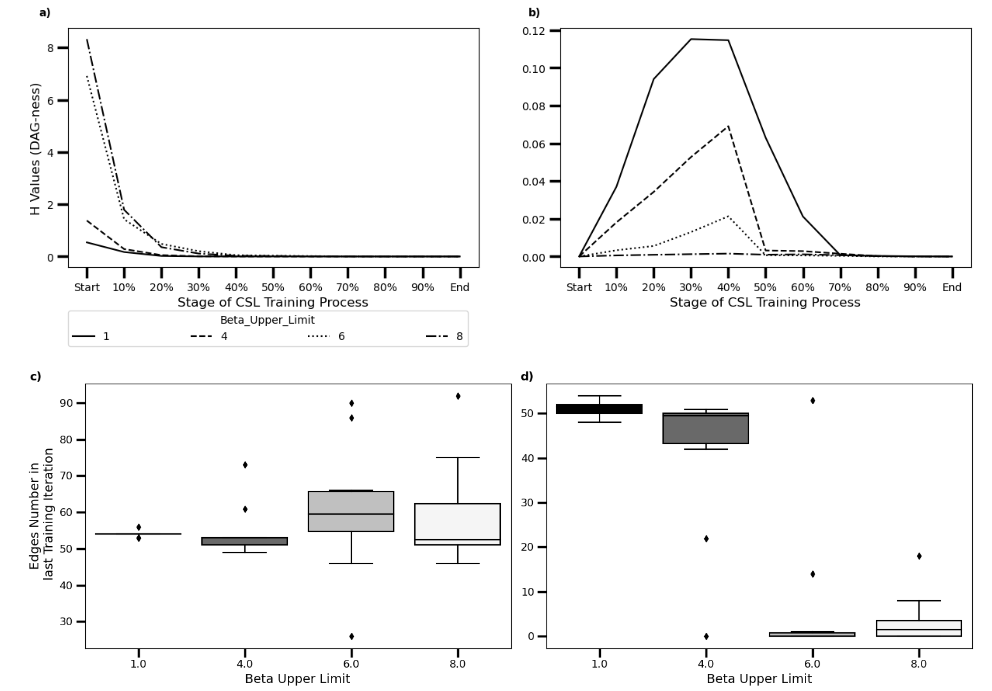}
\caption{a) and b) The value of the acyclicity function, i.e., the \ac{DAG}-ness, of the estimated graphs at different stages of the structure learning process of NoTears-\ac{MLP} and DAGMA, respectively, for beta upper limit $\in [1,4,6,8]$, c) and d) number of edges inferred in the last training iteration of NoTears-\ac{MLP} and DAGMA, respectively, with increasing beta upper limit.}
\label{fig:additionalBetaAnalysis2}
\end{figure}

Figures \ref{fig:additionalBetaAnalysis2}a) and \ref{fig:additionalBetaAnalysis2}b) visualize the change in the \ac{DAG}-ness of the estimated graphs in different stages of the structure learning process. 
While NoTears-\ac{MLP} gradually reduces the value of the acyclicity function during the training process, in the case of DAGMA for most beta levels the value of the acyclicity constraint first increases and then decreases towards the end of the training process. This can be explained by the fact that the acyclicity constraint is applied to the weighted adjacency matrix produced by these models. Since NoTears-\ac{MLP} makes use of the sum of two sets of trainable weights to quantify the causal links between the observed variables, at the beginning of the causal discovery process the magnitude of the acyclicity is higher due to many of the initialized weights still having values different from zero. By contrast, at the beginning of the training process of DAGMA, the value of the acyclicity constraint is very low as DAGMA uses only a single set of trainable weights to generate the causal adjacency matrix. Since the trainable weights in \ac{MLP}s are usually initialized to small values, the value of the resulting acyclicity constraint is also very small. As the training process progresses, DAGMA approximates the functional relationship between the parent and the child vertices, which leads to increasing the magnitude of some weights associated with potential parent-child pairs. At the end of the training process, both NoTears-\ac{MLP} and DAGMA manage to minimize the acyclicity constraint to a value very close to zero. The conditional boxplots in Figures \ref{fig:additionalBetaAnalysis2}c) and \ref{fig:additionalBetaAnalysis2}d) show that with increasing beta upper limit Notears-\ac{MLP}'s pruning phase produces a higher number of causal links, whereas DAGMA tends to estimate a lower number of causal edges. This explains why in general with increasing beta upper limit the number of false discoveries among the predicted edges increases for NoTears-\ac{MLP}, and decreases for DAGMA, as shown in Figure \ref{fig:additionalBetaAnalysis1}b).\\

In addition to the decrease in the number of (false positive) edges with increasing beta coefficients in the case of DAGMA, higher values of beta upper limit lead to a lower change in the \ac{DAG}-ness of the estimated graphs throughout the causal discovery process, as shown in Figure \ref{fig:additionalBetaAnalysis2}b). This implies, that overall the model is barely capable of extracting causal patterns from the data for beta values higher than or equal to eight. We suspect this is due to the fact, that without re-scaling of the passively observed attributes, many of the values in the original data end up in the saturation regions of Sigmoid. The higher the number of saturated values is, the more of the trainable weights would receive no gradient updates. In this regard, we suspect the reason why this tendency cannot be observed for NoTears-\ac{MLP}, is that the approach uses the sum of trainable variables from two weight matrices, which alters the magnitude of the computed gradients w.r.t. acyclicity function. 

\section{Ranking of \ac{CSL} Techniques on re-scaled vs. original Data}
\label{sec:Appendix6}
In this section, we present details about the difference in the \ac{DOS} values on average, when changes to the data scale are performed.\\
\begin{figure}[h]
\centering
\includegraphics[scale=0.65]{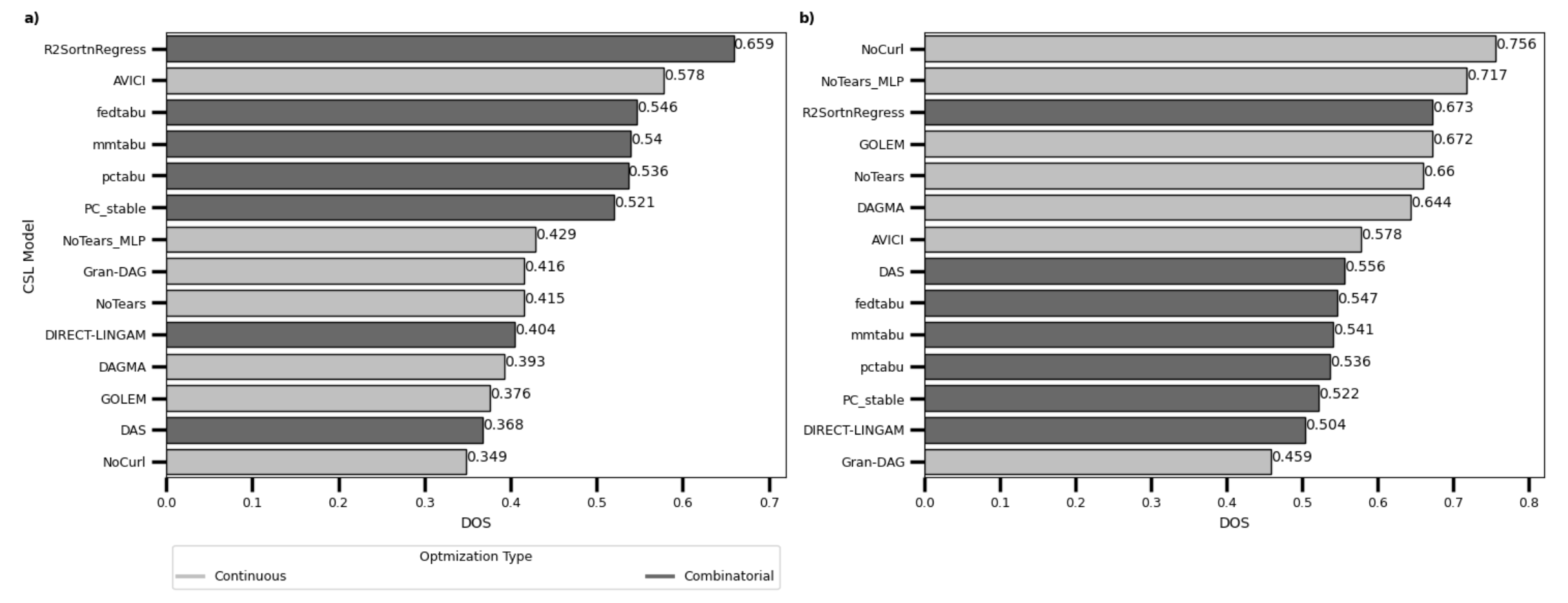}
\caption{Performance of causal discovery models on average in descending order w.r.t. \ac{DOS} a) when using standardized data, b) when using original data.}
\label{fig:RankingRescaledOrignal}
\end{figure}
\indent\\

Figure \ref{fig:RankingRescaledOrignal} shows that the results obtained from test runs on the re-scaled data are fundamentally different compared to the results on the original observational synthetic data. In this regard, $R^2$-SortnRegress represents one of the few exceptions as it takes in both cases one of the first three spots in terms of \ac{DOS} on average. By contrast, the ranking of most continuous optimization techniques changes completely, as on the original data the majority of the gradient-based approaches outperform combinatorial optimization \ac{CSL} models. However, in Section \ref{sec:results} we showed that the data scale has the highest number of interaction effects with other experimental factors included in our simulation framework. In real-life scenarios, a certain attribute could be available in two measuring units, e.g., kilogram and gram. In such settings, the difference in the average \ac{DOS} values produced by the models in Figure \ref{fig:RankingRescaledOrignal} indicates that using the same attributes in different units could potentially result in inferring completely different graphical structures from the passively observed data. For this reason, we believe Figure \ref{fig:RankingRescaledOrignal}a) provides a more realistic overview of the expected performance of the 14 \ac{CSL} techniques included in this empirical research than Figure \ref{fig:RankingRescaledOrignal}b).
\end{appendices}

\end{document}